\documentclass[10pt,journal,compsoc]{IEEEtran}
 
\usepackage{amsmath}
\usepackage{times}
\usepackage{graphicx}
\usepackage{subfigure}
\usepackage[outdir=./figures/]{epstopdf}
\usepackage{color}
\usepackage{multirow}

\usepackage{rotating}
\usepackage{bbm}
\usepackage{latexsym}
\usepackage{amsmath}
\usepackage{latexsym}
\usepackage{wrapfig}
\usepackage{subfigure}
\usepackage{multirow}
\usepackage{comment}
\usepackage{amsmath}
\usepackage{latexsym}
\usepackage{lipsum}

\usepackage{multirow}
\usepackage{comment}
\usepackage{color}
\usepackage{amssymb}
\usepackage{tikz}
\usepackage{bm}
\usepackage{times}
\usepackage[ruled,linesnumbered]{algorithm2e}
\usepackage[utf8]{inputenc}
\usepackage[english]{babel}
\usepackage{lineno}
\usepackage[utf8]{inputenc}
\usepackage[english]{babel}

\usepackage[utf8]{inputenc}
\usepackage[english]{babel}
 
\usepackage{amsthm}
 \usepackage{url}
\usepackage{ marvosym }

\usepackage{lipsum}
\usepackage[bottom]{footmisc}
\usepackage[utf8]{inputenc}
\usepackage[english]{babel}
\usepackage{booktabs}
\usepackage{makecell}
\usepackage[section]{placeins}
 
\newtheorem{theorem}{Theorem}

\newtheorem{lemma}{Lemma}
\usepackage{amsthm,amssymb}
\usepackage{braket}

\usepackage{amsthm}
\usepackage{algorithm2e,setspace}
\usepackage{caption}

\newcommand{\ie}{\emph{i.e.}}
\newcommand{\eg}{\emph{e.g.}}
\newcommand*{\dif}{\mathop{}\!\mathrm{d}}

\def\cao{\textcolor{black}}
\usepackage[colorlinks,linkcolor=blue,citecolor=black]{hyperref}

\usepackage{thmtools}
\declaretheoremstyle[headfont=\normalfont]{normalhead}
\newtheorem{assumption}{Assumption}

\newtheorem{proposition}{Proposition}
\newtheorem{definition}{Definition}

\DeclareMathOperator*{\argmin}{arg\,min}

\ifCLASSOPTIONcompsoc
  \usepackage[nocompress]{cite}
  
  \usepackage{ragged2e}
\else
  \usepackage{cite}
\fi
\ifCLASSINFOpdf
\else
\fi

\begin{document}
%
\title{A Survey of Learning on Small Data: Generalization, Optimization, and Challenge
   %
   }
%
%

\author{Xiaofeng~Cao,  
        Weixin~Bu, Shengjun~Huang,   
         Minling Zhang, \IEEEmembership{Senior Member IEEE} 
        \\
         Ivor W. Tsang, \IEEEmembership{Fellow IEEE},
         Yew Soon Ong, \IEEEmembership{Fellow IEEE},
         and  James T. Kwok, \IEEEmembership{Fellow IEEE} 
\IEEEcompsocitemizethanks{\IEEEcompsocthanksitem \emph{X. Cao  and W. Bu    are with the School of Artificial Intelligence, Jilin University, Changchun, 130012, China.  E-mail: xiaofeng.cao.uts@gmail.com, buwx21@mails.jlu.edu.cn.}} \protect 

\IEEEcompsocitemizethanks{\IEEEcompsocthanksitem \emph{S. Huang  is with the MIIT Key Laboratory of Pattern Analysis and Machine Intelligence, College of Computer Science and Technology, Nanjing University of Aeronautics and Astronautics, Nanjing, 211106, China. 
E-mail: huangsj@nuaa.edu.cn.     }}\protect


\IEEEcompsocitemizethanks{\IEEEcompsocthanksitem \emph{M. Zhang  is with the School of Computer Science and Engineering,
Southeast University, Nanjing 210096, China, and also with the Key Laboratory of Computer Network and Information Integration (Southeast University), Ministry of Education, China. Email:zhangml@seu.edu.cn.     }}\protect

\IEEEcompsocitemizethanks{\IEEEcompsocthanksitem \emph{I. W. Tsang is with the  
Australian Artificial Intelligence Institute,  University of Technology Sydney, NSW 2008, Australia, and the Centre for Frontier AI Research (CFAR), Agency for Science, Technology and Research (A$^*$STAR). 
E-mail: ivor.tsang@uts.edu.au.   }}\protect

\IEEEcompsocitemizethanks{\IEEEcompsocthanksitem \emph{Yew Soon Ong is with the  
School of Computer Science and Engineering, Nanyang Technological University, Nanyang Avenue 639798, Singapore, and  the Chief Artificial Intelligence Scientist of A$^*$STAR. 
E-mail: asysong@ntu.edu.sg.   }}\protect

\IEEEcompsocitemizethanks{\IEEEcompsocthanksitem \emph{J. T. Kwok is with the Department of Computer Science
and Engineering, The Hong Kong University of Science and Technology,
Hong Kong, China. Email: jamesk@cse.ust.hk.}}\protect

\IEEEcompsocitemizethanks{\IEEEcompsocthanksitem \emph{Part of this work was finished when Dr. Cao was a research assistant at AAII of the University of Technology  Sydney.  This work was supported in part by National Natural Science Foundation of China (Grant Number: 62206108), in part by Maritime AI Research Programme (SMI-2022-MTP-06) and AI Singapore OTTC Grant (AISG2-TC-2022-006), and in part by the Research Grants Council of the Hong Kong Special Administrative Region (Grant 16200021). }}\protect 
}
%
%

\markboth{Journal of \LaTeX\ Class Files,~Vol.~14, No.~8, August~2015}%
{Shell \MakeLowercase{\textit{et al.}}: Bare Demo of IEEEtran.cls for Computer Society Journals}
%


\IEEEtitleabstractindextext{%
\begin{abstract}\justifying  
Learning on big data brings success for artificial intelligence (AI), but the annotation and training costs are expensive. In future, learning on small data that approximates the generalization ability of big data is one of the ultimate purposes of AI, which requires machines to recognize objectives and scenarios relying on small data as humans. A series of learning topics is going on this way such as active learning and few-shot learning. However, there are few theoretical guarantees for their generalization performance. Moreover, most of their settings are passive, that is, the label distribution is explicitly controlled by finite training resources from known distributions. This survey follows the agnostic active sampling theory under a PAC (Probably Approximately Correct) framework to analyze the generalization error and label complexity of learning on small data in model-agnostic supervised and unsupervised fashion. Considering multiple learning communities could produce small data representation and related topics have been well surveyed, we thus subjoin novel geometric representation perspectives for small data: the Euclidean and non-Euclidean (hyperbolic) mean, where the optimization solutions including the Euclidean gradients, non-Euclidean gradients, and Stein gradient are presented and discussed. Later, multiple learning communities that may be improved by learning on small data  are summarized, which yield data-efficient representations, such as transfer learning, contrastive learning, graph representation learning. Meanwhile, we find that the meta-learning may provide effective parameter update policies for learning on small data. Then, we explore multiple challenging scenarios for small data, such as the weak supervision and multi-label. Finally, multiple data applications that may benefit from efficient small data representation are surveyed.
\end{abstract}

\begin{IEEEkeywords}
Big data, artificial intelligence, small data, active learning, PAC, theoretical guarantee, model-agnostic, hyperbolic.
\end{IEEEkeywords}}

\maketitle

\IEEEdisplaynontitleabstractindextext

\IEEEpeerreviewmaketitle

\IEEEraisesectionheading{\section{Introduction}\label{sec:introduction}}

\IEEEPARstart{``T}{hat's} a cat sleeping in the bed, the boy is patting the elephant, those are people that   are going on an airplane, that's a big airplane...''.  ``This is a three-year child describing the pictures she sees'' -- said by Fei-Fei Li.  She presented a  famous lecture of  "how we are teaching computers to understand pictures''   in the Technology Entertainment Design (TED) 2015 \footnote{\url{https://www.ted.com/talks/fei_fei_li_how_we_re_teaching_computers_to_understand_pictures?language=en}}.
In the real world, humans can recognize objectives and scenarios only relying on one picture based on their prior knowledge. However, machines may need more. In the past few decades, artificial intelligence  (AI)  \cite{russell2002artificial} \cite{nilsson2014principles} technology helped machines to be more intelligent like humans by learning on big data \cite{o2013artificial} \cite{labrinidis2012challenges}. By modeling  the neuron propagation of the human brain, a series of  expressive AI systems are built,  \textit{e.g.}, Deep Blue \cite{campbell2002deep}, AlphaGo \cite{wang2016does}.

Of course, the talent of AI is not innate. Training on big data helps Al to recognize different objectives and scenarios. To process big data,  a set of techniques, \textit{e.g.},  MapReduce \cite{dean2008mapreduce}, Hadoop \cite{white2012hadoop}, were implemented to access large-scale data,  extracting useful information for AI decisions.   Specifically, MapReduce is  distributed across multiple heterogeneous clusters and Hadoop processes data  through   cloud providers.
However, training and annotating large-scale data are quite expensive, although  we adopt those big data processing techniques.

 ``AI is not just for the big guys anymore\footnote{\url{https://www.forbes.com/sites/forbestechcouncil/2020/05/19/the-small-data-revolution-ai-isnt-just-for-the-big-guys-anymore/?sh=4b8d35c82bbb}}.'' A novel perspective thinks  the small data revolution is ongoing and  training on small data with a desired performance is one of the ultimate purpose of AI. Technically, human experts expect  to relieve the need on big data  and find  a new breakthrough for AI systems, especially for the configuration of the deep neural networks \cite{segler2018planning}. Related works including limited labels \cite{chen2019scene}     \cite{iosifidis2017large}, fewer labels\cite{luvcic2019high} \cite{ji2019learning} \cite{xu2020weakly}, less data \cite{moore1993prioritized} \cite{baraniuk2011more}, \emph{etc.}, were already realized by those low-resource deep learning researchers. Formally, few-shot learning \cite{sung2018learning}, which  is referred to as low-resource learning,  is a  unified topic which studies small data with limited information.  Based on  Wang \emph{et al.}'s survey \cite{wang2020generalizing}, an explicit scenario of few-shot learning  is feature generation \cite{xian2019f}, that is, generating artificial features by the given   limited or insufficient information.  
 Another scenario with implicit supervision information   is  more challenging, which relies  on retraining the learning model \cite{xian2019f}  \cite{feng2019partial}  with  those highly-informative examples, such as private data. Theoretically, most of the few-shot learning  scenarios are passive, that is, the label   distribution is explicitly controlled by   finite training resources from known   distributions, such as massive  training samples drawn from low-representative distribution regions, and 
task-dependent non-optimal model configurations. Thence, active learning \cite{settles2009active}   attracts our eyes where its label  acquisitions are controlled by a learning algorithm or humans.

Different from few-shot learning, the annotation scenario of   active learning  is not so limited.  One active learning algorithm can stop its iterative sampling at anytime due to desired algorithm performance or  exhausted annotation budget. There are two categories for active learning: active sampling theory over hypothesis class \cite{hanneke2014theory} and active sampling algorithm over realized scenarios \cite{settles2012active}, where the theory studies present the label complexity and convergence guarantees for those algorithmic paradigms.  The typical theoretical analyses are derived from a PAC ((Probably Approximately Correct)) \cite{haussler1990probably} style, which aims at an agnostic setting such as \cite{dasgupta2008general}. To control the active sampling, there is one kind of  error disagreement coefficient  that searches a target data, which can maximize the hypothesis update the most, where those updates are required to be positive and helpful. Therefore, the active sampling  also is a hypothesis-pruning \cite{cao2020shattering} process, which tries to find the optimal hypothesis from a given hypothesis class, where the hypotheses are maintained from a version space \cite{hirsh1994generalizing} \cite{mitchell1977version} over the decision boundaries \cite{lee1997decision} of classes. Geometrically, the version space of an enclosed class is usually embedded in a tube structure \cite{ben2008relating}   \cite{li2020finding} which has a homeomorphic topology as with the assumed spherical class. 
 
 \subsection{Motivations and Contributions}
Learning on small data is essential for advancing AI.  As a  preemptive topic, few-shot learning presented exploration for limited data training. However, the setting of few-shot learning is a passive scenario, which stipulates   insufficient label information by the task itself. Meanwhile, 
  there are few theoretical guarantees for its generalization performance in task-independent settings. This motivates us to give  theoretical analysis for learning on small data with model-agnostic generalization. 
 \cao{Considering that active learning theory provides an effective i.i.d. protocol on sampling},  we follow its PAC framework to present a set of   error and label complexity bounds for learning on small data. To summarize those algorithmic paradigms, we then categorize the small data learning models  into: the Euclidean  and hyperbolic  (non-Euclidean)  representation  including their  deep  learning scenarios.  Concretely, contributions of this survey are summarized as follows.
 
 \begin{itemize}
 \item  \cao{We present a formal definition for learning on small data that presents efficient generalization approximation to   big data representation.} The definition is a model-agnostic setting that derives  a more general concept from a  generalization perspective. 
 
 \item From a PAC perspective, we are the first to present theoretical guarantees for learning on small data via active sampling theory. The  generalization error and label complexity bounds of learning on small data are presented under a model-agnostic fashion.
 
 \item From a geometric perspective, we divide the    small data representation models  into two categories: the Euclidean  and hyperbolic representation, where their optimization solvers \cao{including the Euclidean gradients, non-Euclidean gradients, and Stein gradient are   analyzed and discussed.}
 
 \item \cao{We investigate multiple potential   learning communities that may be improved to be more data-efficient  from    small data representation,   such as transfer learning, contrastive learning, graph representation learning, and also find that meta-learning may provide effective parameter optimization policies for learning on small data.}
 
 \item \cao{We explore multiple scenarios that may bring challenge for learning on small data, \textit{e.g.},    weak  supervision,  multi-label, and imbalanced distribution. The data applications that may benefit from efficient small data representation but still have challenge are also surveyed, including large image data, large language oracle,  and science data. }
 
 \end{itemize}
 
  \begin{figure} 
    \centering
    \includegraphics[scale=0.43]{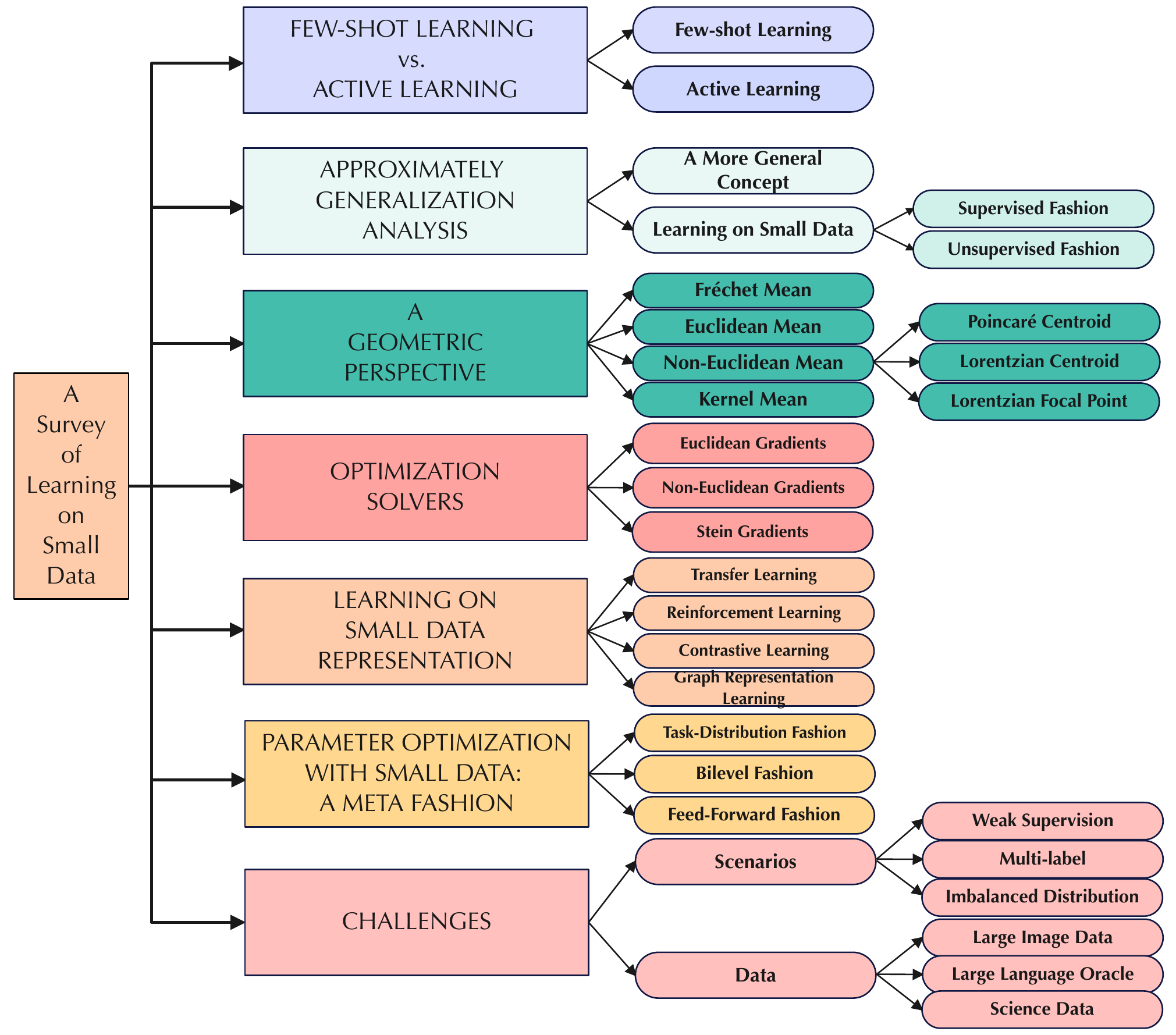}
    \caption{The organization of the survey.} 
    \label{fig:Organization}
\end{figure}
\subsection{PAC Framework of This Survey}
In 1984,  Leslie Valiant proposed a computational learning concept-Probably Approximately Correct (PAC) \cite{valiant1984theory}, which presents  mathematical analysis for machine learning under fixed  distribution and parameter assumptions.  Theoretically,  the learner is required to select a generalization hypothesis (also called conceptual function) from a candidate hypothesis class by observing those received data and labels. The goal is to converge the hypothesis into approximately correct generalization, which properly describes the probability distribution of the unseen samples. One key content of PAC leaning is to derive the computational complexity  bounds such as sample complexity \cite{hanneke2016optimal}, 
generalization errors, Vapnik–Chervonenki(VC) dimension \cite{blumer1989learnability}.

In computational learning theory,   active learning tries to prune the candidate infinite concept class into the optimal  hypothesis, which keeps consistent properties as with its labeled examples \cite{hanneke2014theory}. The main difference to typical PAC learning is that the active learning controls the hypothesis-pruning by receiving fewer training data. Therefore, active learning that finds hypotheses consistent with a small set of labeled examples,  could be deemed as a standard hypothesis-pruning to   PAC learning. With this framework, we present this survey.

 \subsection{Organization of The Survey}

  The outline  of this survey is presented  as follows. 
 \begin{itemize}
     \item \cao{ Section~2  characterizes the few-shot vs. active  learning from a hypothesis-pruning perspective, where the hypothesis update policy of active  learning presents a fundamental guidance for i.i.d. sampling of learning on small data.}
     \item \cao{Section~3 presents a formal definition for learning on small data and presents its PAC analysis including the label complexity and generalization error bounds.   }
     
       \item \cao{From a geometric perspective, Section~4 introduces the Euclidean and non-Euclidean paradigms  for learning on small data. }
       
       \item \cao{To optimize the geometric representation,  Section~5 presents related optimization solvers including  the Euclidean gradients, non-Euclidean
gradients, and Stein gradient.}
       
       \item \cao{Section~6 presents  multiple potential learning communities
that may be improved to be more data-efficient from small
data representation, such as transfer learning, contrastive
learning, and   graph representation learning.}

    \item \cao{ Section~7 presents the meta-learning that  may provide effective parameter optimization policies for learning on small data.}
    
    \item \cao{Section~8 investigates multiple   scenario and data  challenges  for
      small data representation, \textit{e.g.}, weak 
supervision,  multi-label, imbalanced distribution, large image data, large language oracle. }
      
     \item \cao{Section~9 finally concludes this survey. }
 \end{itemize}
 With the above outline, we present the organization of the survey in Figure~\ref{fig:Organization}.
 
\section{Few-shot Learning vs. Active Learning}
Few-shot learning could be considered as   a preemptive topic of learning on small data with passive scenario. Differently, active learning also presents solutions for small data, but with an active sampling scenario. 

\subsection{Few-shot Learning}
Finding the optimal hypothesis consistent with the full training set is a standard  theoretical description of machine learning to PAC. The convergence process is to perform hypothesis-pruning in the candidate hypothesis class/space. Therefore, the number of the hypotheses of a fixed geometric region  determines  the volume of the  hypothesis space, which affects the speed and cost of hypothesis-pruning. 

Given a full training set $\mathcal{X}$ with $n$ samples, let $\mathcal{H}$  be the hypothesis class, the VC dimension bound of $\mathcal{H}$ can be used to describe the complexity of the  convergence difficulty of the hypothesis-pruning 
for  a given   learning algorithm.
We thus follow the agnostic active learning \cite{dasgupta2008general} to define $N(\mathcal{H}, n, k, \mathcal{A})$ as a class of function which controls the convergence of  the learning algorithm  $\mathcal{A}$  in hypothesis class $\mathcal{H}$,  associating from  $n$ training samples with $k$  classes.  Note that the following definitions describe each learning process from a pruning perspective in the model-agnostic hypothesis space. 
 Related background of few-shot learning and active learning have been well surveyed in \cite{wang2020generalizing} and \cite{settles2009active}, \textit{etc}. We thus don not repeat more.

\begin{definition}
 Machine learning. From a hypothesis-pruning perspective, given any machine learning algorithm $\mathcal{A}$,   its candidate hypothesis class  is characterized  by $\mathcal{H}$, which satisfies   1) a VC dimension bound of  $\mathcal{O}(2^n)$, and 2) a safety uniform bound  of pruning into a non-null hypothesis is  $N(\mathcal{H}, n, k, \mathcal{A})\geq \mathcal{O}(\frac{k-1}{k}n)$,  and 3)  the  VC dimension bound of a non-null hypothesis subspace is   $\mathcal{O}\big(2^n-2^{\frac{(k-1)n}{k}}\big)$.
\end{definition}
Note that the uniform bound aims at an expected complexity and a non-null hypothesis requires the training examples to cover all label categories. Given any class has at least $\eta$ data, 
 a safety guarantee requires   $N(\mathcal{H}, n, k, \mathcal{A})\geq \mathcal{O}(n-\eta)$. For a uniform estimation on $\eta\approx \mathcal{O}(\frac{n}{k})$ over all possibilities of $\eta=1,2,3,...,\frac{n}{k}$, a safety uniform bound  of pruning into a non-null hypothesis is  $\mathcal{O}(\frac{k-1}{k}n)$.

Assume that  $\eta \ll \frac{n}{k}$, the typical machine learning scenario becomes a few-shot learning process.
\begin{definition}\label{few-shot-learning}
Few-shot learning. From a hypothesis-pruning perspective, given any few-shot learning algorithm   
$\mathcal{A}$,   its candidate hypothesis class  is characterized  by $\mathcal{H}$, which satisfies   1) a VC dimension bound of  $\mathcal{O}(2^n)$,  2) a safety uniform bound  of pruning into a non-null hypothesis is  $N(\mathcal{H}, n, k, \mathcal{A})\geq \mathcal{O}(n-\eta)$, and 3) a  VC dimension bound of shrinking into a non-null hypothesis subspace is  $\mathcal{O}(2^n-2^{n-\eta})$, where $2^n-2^{n-\eta} \gg \frac{k-1}{k}n$.
\end{definition}

 Here, `safety' means that the pruning process can properly converge.  From Definition~\ref{few-shot-learning}, few-shot learning can be deemed as a special case of typical machine learning with limited supervision information. One important characteristics of it is its tighter volume of the non-null hypothesis space since $\mathcal{O}(2^n-2^{n-\eta}) \gg \mathcal{O}(\frac{k-1}{k}n)$. Therefore, compared with typical machine learning, any few-shot learning algorithm  will result in looser safety bound to prune into a non-hypothesis. 

 In realizable settings,  one typical few-shot learning   scenario is    feature generation \cite{xian2019f} via model retraining. In this scenario, the learning algorithm generates  handwritten
features like humans by pre-training the model with prior knowledge, where the  retaining will not stop, until a desired performance is achieved.
However, retraining the learning model is adopted in rare cases, which cannot properly generate highly-trust features. For example,  one special case of few-shot learning is one-shot learning \cite{vinyals2016matching} which relies on only one data in some classes; the other  more extreme  case is zero-shot learning \cite{romera2015embarrassingly} where some classes don not include any data or labels.  Their detailed definitions are presented as follows.
 \begin{definition}
One-shot learning. From a hypothesis-pruning perspective, given any one-shot learning algorithm   
$\mathcal{A}$,   its candidate hypothesis class  is characterized  by $\mathcal{H}$, which satisfies   1) a VC dimension bound of  $\mathcal{O}(2^n)$,  2) a safety uniform bound  of pruning into a non-null hypothesis is  $N(\mathcal{H}, n, k, \mathcal{A})\geq \mathcal{O}(n)$, and 3) the  VC dimension bound of shrinking into a non-null hypothesis subspace is  $\mathcal{O}(2^n)$.
 \end{definition}
 
Different to the `safety' bound, `inapplicable' means that the pruning can not converge into the desired hypothesis.
 
 \begin{definition}
Zero-shot learning. From a hypothesis-pruning perspective, given any zero-shot learning algorithm   
$\mathcal{A}$,   its candidate hypothesis class  is characterized  by $\mathcal{H}$, which satisfies   1) a VC dimension bound of  $\mathcal{O}(2^n)$,  2)  an inapplicable safety uniform bound  of pruning into a non-null hypothesis, and 3) an inapplicable  VC dimension bound of shrinking into a non-null hypothesis subspace.
 \end{definition}
  
 Usually, the few-shot learning is also related to weakly-supervised learning \cite{liu2019prototype} \cite{li2019towards}, which includes incomplete, inaccurate, noisy, and outlier information, \emph{etc.} From this perspective, few-shot learning can be considered as one special setting of weakly-supervised learning with incomplete label information.
Imbalanced learning \cite{he2009learning}, transfer learning \cite{yu2020transmatch}, meta-learning \cite{jamal2019task}, \emph{etc.}, also have inherent connections with few-shot learning. However, there are no theoretical analyses for the convergence of the optimal hypothesis.
 
\subsection{Active Learning} 
Active learning prunes the candidate hypothesis class into a desired convergence. The pruning process usually is to zoom out \cite{cortes2019active} the hypothesis space by querying those highly-informative updates. Therefore, the assumption of active learning requires any update from the hypothesis-pruning should be non-null. Here we present the definition on active learning.

 \begin{definition}
Active  learning. From a hypothesis-pruning perspective, given any active  learning algorithm   
$\mathcal{A}$ with a querying budget $\mathcal{Q}$,   its candidate hypothesis class  is characterized  by $\mathcal{H}$, which satisfies   1) a VC dimension bound of  $\mathcal{O}(2^n)$,  2) a safety uniform bound  of pruning into a non-null hypothesis is   $\mathcal{O}(1)$, and 3) the VC dimension bound of a non-null hypothesis subspace is   $\mathcal{O}(2^\mathcal{Q})$.
 \end{definition}

Note that active learning requires the updates of hypotheses are positive and non-null. Any subsequent hypotheses can converge into a safety state, which derives a safety uniform bound of $\mathcal{O}(1)$. 
Different from  few-shot learning, the scenario  of active learning is controlled by humans, which always keep a non-null update on hypotheses. It is thus that its VC dimension bound is tighter than the typical machine learning and few-shot learning. To find feasible hypothesis updates, active learning always uses an error disagreement coefficient  \cite{dasgupta2011two} to control the hypothesis-pruning.

\textbf{Error disagreement.} Given a finite hypothesis class $\mathcal{H}$, active learning iteratively updates  the current hypothesis $h_\mathcal{Q}\in \mathcal{H}$  at $t$-time into the optimal hypothesis $h^*\in \mathcal{H}$. Let an active learning algorithm $\mathcal{A}$ perform $\mathcal{Q}$ rounds of querying from $\mathcal{X}$,  assume that $\ell(\cdot,\cdot)$ denotes the loss of mapping $\mathcal{X}$ into $\mathcal{Y}$ with multi-class setting, we define the total loss of the  $\mathcal{Q}$ rounds of querying as $R(h_\mathcal{Q})=\sum_{i=1}^{\mathcal{Q}}\frac{q_t}{p_i}\ell(h(x_t), y_t)$, where $y_t$ denotes the label of $x_t$, $q_t$ satisfies the Bernoulli distribution of $q_t\in \{0,1\}$, and $\frac{1}{p_i}$ denotes the weight of sampling $x_t$. On this setting, the sampling process then adopts an error disagreement to control the hypothesis updates:
 \begin{equation}\label{eq:hypothesis-updates}
   \theta_{\rm \mathcal{A}}= \mathbb{E}_{x_t\in \mathcal{D}}   \mathop{{\rm sup}}\limits_{h\in B(h^*,r)} \left \{  \frac{ \ell(h(x_t),\mathcal{Y})-\ell(h^*(x_t),\mathcal{Y})   }{r}       \right\},
  \end{equation}
 where $\mathcal{D}$ denotes the margin distribution over  $\mathcal{X}$, drawing the candidate hypotheses. To reduce the complexity of the pruning process, one can shrink  $\mathcal{D}$  from the marginal distribution of $\mathcal{X}$, which derives the most of hypotheses, such as   \cite{dasgupta2008general}, \cite{cohn1994improving}.

 Correspondingly,     $\ell(h(x), h'(x))$   denotes  the hypothesis disagreement  of $h$ and $h'$, which can be specified as the best-in-class error  on $\mathcal{Y}$
\begin{equation}\label{eq:loss}
\begin{split}
\ell(h(x),h'(x))= \Big|\mathop{ {\rm max}}\limits_{y \in \mathcal{Y}} \ell(h(x),y)-\ell(h'(x),y)\Big|,
\end{split}
\end{equation}
where   $y\in \mathcal{Y}$, and $\ell(h(x),h'(x))$ also can be simply written as $\ell(h,h')$.  Once the hypothesis update w.r.t. error  after adding $x_t$ is larger than $\theta_{\rm \mathcal{A}}$, the active learning algorithm $\mathcal{A}$  solicits $x_t$ as a significant update. Besides Eq.~(\ref{eq:loss}), $\ell(h(x),h'(x))$ also can be specified as all-in-class error \cite{cortes2020adaptive},  error entropy \cite{roy2001toward}, \emph{etc.}

\section{Approximately Generalization Analysis}
 From a hypothesis-pruning perspective, we firstly present a more general concept for learning on small data. Then, we present the generalization analysis of the convergence of the optimal hypothesis on error and label complexity bounds under a model-agnostic supervised and unsupervised fashion, respectively.
\subsection{A More General Concept}
For agnostic sampling, any hypothesis $h\in \mathcal{H}$ can achieve a  generalization error of $err(h)$. With a probability at least $1-\delta$, after $\mathcal{Q}$ times of sampling,  if $err(h)$ converges into its optimal error, based on Theorem~1 of \cite{dasgupta2008general}, there exits  an upper bound of $err(h)+c \Big(\frac{1}{\mathcal{Q}}\big(d {\rm log}\mathcal{Q}+ {\rm log} \frac{1}{\delta}\big)+\sqrt{ \frac{err(h)}{\mathcal{Q}}{\mathcal{Q}(d {\rm log} \mathcal{Q}+ {\rm log}(\frac{1}{\delta}) }} \Big)$. By relaxing the constant $c$ and $err(h)$ ($err(h)<1$), 
the label complexity of any learning algorithm satisfies an upper bound of
\begin{equation}\label{eq:upper-bound}
N(\mathcal{H}, n, \mathcal{Q}, \mathcal{A}) \leq \mathcal{O}\Big(\frac{1}{\mathcal{Q}}\big(d {\rm log}\mathcal{Q}+ \rm{log} \frac{1}{\delta}\big) \Big).
\end{equation}
Eq.~(\ref{eq:upper-bound}) presents a coarse-grained observation on the upper bound of the label complexity. 
We next  introduce the error disagreement coefficient $\theta_\mathcal{A}$ to prune the hypothesis class.  
If the learning algorithm controls the hypothesis updates by Eq.~(\ref{eq:hypothesis-updates}), based on Theorem~2 of \cite{dasgupta2008general},  the expected label cost for the convergence of $err(h)$ is at most $1+c\theta_\mathcal{A}\Big(\big(d {\rm log} \mathcal{Q} + {\rm log} \frac{1}{\delta}\big) {\rm log} \mathcal{Q}\Big)$. By relaxing the constant $c$, we have 
\begin{equation}\label{eq:N}
N(\mathcal{H}, n, \mathcal{Q}, \mathcal{A}) \leq \mathcal{O}\Big( \theta_\mathcal{A}\big(d {\rm log} \mathcal{Q} + {\rm log} \frac{1}{\delta}\big) {\rm log} \mathcal{Q}\Big).
\end{equation}
With the inequalities of  Eqs.~(\ref{eq:upper-bound}) and (\ref{eq:N}), we present a more general concept for small data.
\begin{definition}Learning on small data.
With standard empirical risk minimization, learning small data from $\mathcal{D}$ over $\mathcal{Q}$ times of sampling satisfies an incremental update on the optimal hypothesis with an error of  $err(h^*)$,
\begin{equation}
\begin{split}
& \argmin_{\mathcal{Q}} err(h_\mathcal{Q})\!\leq \!   \Bigg( err(h^*)+ \mathcal{O}\Big(\sqrt{err(h^*)  \Omega}  +\Omega  \Big) \Bigg)\\
 &{\rm  s.t.} \ \  \Omega=  \frac{d   {\rm log} \mathcal{Q}+{\rm log}\frac{1}{\delta}}{ \mathcal{Q}}, \\
 \end{split}
\end{equation}
\cao{where $h_\mathcal{Q}$ denotes the updated hypothesis at the $\mathcal{Q}$-time of sampling, yielding  efficient    generalization error  approximation  to the optimal $h^*$, that is, $h_\mathcal{Q}$ holds nearly consistent generalization ability  to $h^*$ in the hypothesis class $\mathcal{H}$.}
\end{definition}

 \cao{Definition~6  present a formal definition for learning on small data which requires efficient generalization error approximation to that of  the optimal hypothesis relying on the original  big (full) data representation. The definition is a model-agnostic setting via  $\mathcal{Q}$ times of i.i.d. sampling,   deriving an efficient approximation for the $\mathcal{Q}$-time hypothesis $h_\mathcal{Q}$ to the optimal $h^*$ in the hypothesis space $\mathcal{H}$.} 

\subsection{Learning on Small Data}
With the standard definition on small data, we next study that how to learn  small data via empirical risk minimization (ERM), which can be generalized into different loss functions in real-world models. Our main theorem of the label complexity in regard to ERM is then presented in Theorems~\ref{Supervised-Fashion} and \ref{Unsupervised-Fashion}.

Before presenting Theorem~\ref{Supervised-Fashion}, we need a technical lemma about the  importance-weighted empirical risk minimization on $\ell(h_\mathcal{Q},h^*)$. The involved techniques  refer to the Corollary~4.2  of J. Langford \emph{et al.}'s work  
in \cite{langford2005tutorial},  and the   Theorem~1 of    C.~Sahyoun \emph{et al.}'s work
\cite{DBLP:conf/icml/BeygelzimerDL09}. 
 
\begin{lemma}
Let $R(h)$ be the   expected loss (also called learning risk) that stipulates   $R(h)=\mathbb{E}_{x\sim \mathcal{D}}[\ell(h(x),y)]$, and $R(h^*)$ be its  minimizer. On this setting, $\ell(h_\mathcal{Q},h^*)$ then can be bounded by  $\ell(h_\mathcal{Q},h^*) \leq R(h_\mathcal{Q})-R(h^*) $ that stipulates  $\mathcal{H}_{\mathcal{Q}}:=\{h\in \mathcal{H}_{\mathcal{Q}-1}: R(h_\mathcal{Q})\leq R(h^*)+2\Delta_{\mathcal{Q}-1} \}$, where $\Delta_{\mathcal{Q}-1}$ adopts a form \cite{cortes2020region} of 
\[ \frac{1}{\mathcal{Q}-1}  \Bigg[ \sqrt{ \Big[\sum_{s=1}^{\mathcal{Q}-1} p_s\Big]  \rm{log}\Big[\frac{(\mathcal{Q}-1)|\mathcal{H}|}{\delta}\Big]   }  +  \rm{log}\Big[\frac{(\mathcal{Q}-1)|\mathcal{H}|}{\delta}\Big]  \Bigg ], \]
where $|\mathcal{H}|$ denotes the number of hypothesis in  $\mathcal{H}$, and $\delta$ denotes a   probability threshold requiring $\delta>0$.  
Since $\sum_{s=1}^{\mathcal{Q}-1} p_s \leq \mathcal{Q}-1$, $\Delta_{\mathcal{Q}-1}$ can then be bounded by 
 \[\Delta_{\mathcal{Q}-1}=\sqrt{\Big(\frac{2}{\mathcal{Q}-1}\Big){\rm log}\Big(2\mathcal{Q}(\mathcal{Q}-1)\Big)  \frac{|\mathcal{H}|^2  }{\delta})},\] 
which denotes the loss disagreement bound to approximate a desired target hypothesis such that $R(h_\mathcal{Q})-R(h^*)\leq 2\Delta_{\mathcal{Q}-1}$.
\end{lemma}

There are two fashions to learn small data including supervised and unsupervised learning. We next present their different generalization analyses. 
\subsubsection{Supervised Fashion}  
We follow the setting of Lemma~1 to present the learning risk and label complexity for learning on small data under $\mathcal{Q}$ rounds of importance sampling.

\begin{theorem}\label{Supervised-Fashion}
Given $\mathcal{Q}$ rounds of querying by employing the active learning algorithm $\mathcal{A}$,  with a probability at least  $1-\delta$, for all $\delta>0$, for any $\mathcal{Q}>0$,  the error disagreement of   $R(h_\mathcal{Q})$ and   $R(h^*)$ of learning on small data is bounded by  
\begin{equation*}
\begin{split}
&R(h_\mathcal{Q})-R(h^*)\\
&\leq  \max_{\mathcal{Q}} \Bigg\{\frac{2}{\mathcal{Q}}  \Bigg[\sqrt{\sum_{t=1}^{\mathcal{Q}}p_t}+6\sqrt{{\rm log}\Big[\frac{2(3+\mathcal{Q})\mathcal{Q}^2}{\delta}\Big] } \Bigg]\\
  &\qquad \qquad\qquad\qquad\times \sqrt{{\rm log}\Big[\frac{16\mathcal{Q}^2|\mathcal{H}_i|^2 {\rm log}\mathcal{Q}}{\delta}\Big]}\Bigg\},
\end{split}
\vspace{-5pt}
\end{equation*} 
Then, with a probability at least  $1-2\delta$, for all $\delta>0$, the label complexity of learning on small data can be bounded by
\begin{equation*} 
\begin{split}
&N(\mathcal{H}, n, \mathcal{Q}, \mathcal{A}) \leq   \max_{ \mathcal{Q}} K_\ell\Bigg\{ \Big[\sum_{j=1}^{ \mathcal{Q}} \theta_{\mathcal{A}} R_j^* \mathcal{Q}  p_j\Big]\\ 
 &  \!+\!\sum_{j=1}^{ \mathcal{Q}} O\Bigg(\sqrt{R_j^*\mathcal{Q}  p_j{\rm log}\Big[\frac{ \mathcal{Q}|\mathcal{H}_i| \mathcal{Q}}{\delta} \Big]}\Bigg)   \!  + \! O\Bigg( \mathcal{Q} {\rm log}^3\Big(\frac{\tau|\mathcal{H}_i| \mathcal{Q}}{\delta}\Big)\Bigg)\Bigg\}.
\end{split}
\end{equation*} 
where $K_\ell$ is the slope asymmetry  over the   loss $\ell$,   $K_\ell= \mathop{{\rm sup}}\limits_{x_t', x_t\in \mathcal{D}} \left |\frac{{\rm max} \ \ell(h(x_t), \mathcal{Y})-\ell(h(x_t'), \mathcal{Y}) }    {{\rm min} \ \ell(h(x_t), \mathcal{Y})- \ell(h(x_t'), \mathcal{Y}) } \right|$, $R_j^*$ denotes the best-in-class risk at $j$-time querying, and $|\mathcal{H}|$ denotes the element number of  $\mathcal{H}$. 
\end{theorem}
The proofs of Theorem~1 and 2 of \cite{cortes2020region} can be adopted to prove the two inequalities of Theorem~\ref{Supervised-Fashion}, respectively.
\subsubsection{Unsupervised Fashion} 
By employing unsupervised learning, the learning risk and label complexity of Theorem~\ref{Supervised-Fashion} are degenerated into a polynomial expression \cite{CaoMHE2022}.

Given the input dataset $\mathcal{X}$ with $n$ samples, it is divided into $k$ clusters: $\{\mathcal{B}_1, \mathcal{B}_2,..., \mathcal{B}_k\}$, where  $\mathcal{B}_i$ has $N_i$ samples. Learning  small data   performs IWAL for any $\mathcal{B}_i$. Specifically, it employs a new error disagreement $\theta_{\rm LSD}$ to control the hypothesis updates: 
\begin{equation}
 \theta_{\rm LSD}= \mathbb{E}_{x_t\in \mathcal{B}_i}   \mathop{{\rm sup}}\limits_{h\in B(h^*,r)} \left \{  \frac{ \ell(h(x_t),\mathcal{Y})-\ell(h^*(x_t),\mathcal{Y})   }{r}       \right\}.
\end{equation}
\begin{theorem}\label{Unsupervised-Fashion}
Given $T$ rounds of querying by employing the active learning algorithm $\mathcal{A}$, let $\mathcal{Q}$ be the number of ground-truth queries. If  learning small data performs $\mathcal{A}$  for any $\mathcal{B}_i$, each cluster  will have $\tau=T/k$ rounds of querying. Then, with a probability at least  $1-\delta$, for all $\delta>0$, for any $\mathcal{Q}>0$,  the error disagreement of   $R(h_\tau)$ and   $R(h^*)$ of learning on small data  is bounded by $k$ times of  polynomial 
\begin{equation*}
\begin{split}
& R(h_\tau)-R(h^*)   \\
&\leq k \times \max_{\mathcal{H}_i, i=1,2,...,k} \Bigg\{\frac{2}{\tau}  \Bigg[\sqrt{\sum_{t=1}^{\tau}p_t}+6\sqrt{{\rm log}\Big[\frac{2(3+\tau)\tau^2}{\delta}\Big] } \Bigg]\\
& \qquad\qquad\qquad\qquad \qquad \times \sqrt{{\rm log}\Big[\frac{16\tau^2|\mathcal{H}_i|^2 {\rm log}\tau}{\delta}\Big]}\Bigg\},
\end{split}
\vspace{-5pt}
\end{equation*} 
Then, with a probability at least  $1-2\delta$, for all $\delta>0$, the label complexity of learning on small data can be bounded by
\begin{equation*} 
\begin{split}
&N(\mathcal{H}, n, \mathcal{Q}, \mathcal{A})  \\
 &\leq 8k \times \max_{\mathcal{H}_i, i=1,2,...,k} K_\ell\Bigg\{ \Big[\sum_{j=1}^{N_i} \theta_{\rm LSD} R_j^*\tau p_j\Big]\\ 
 &   +\sum_{j=1}^{N_i} O\Bigg(\sqrt{R_j^*\tau p_j{\rm log}\Big[\frac{\tau|\mathcal{H}_i|N_i}{\delta} \Big]}\Bigg)   \!+\! O\Bigg(N_i {\rm log}^3\Big(\frac{\tau|\mathcal{H}_i|N_i}{\delta}\Big)\Bigg)\Bigg\}.
\end{split}
\end{equation*} 
where $K_\ell$ is the slope asymmetry over the limited loss $\ell$ on $\mathcal{B}_i$, \textit{i.e.}, 
$\ell_{\mathcal{B}_i}$,   $K_\ell= \mathop{{\rm sup}}\limits_{x_t', x_t\in \mathcal{B}_i} \left |\frac{{\rm max} \ \ell_{\mathcal{B}_i}(h(x_t), \mathcal{Y})-\ell_{\mathcal{B}_i}(h(x_t'), \mathcal{Y}) }    {{\rm min} \ \ell_{\mathcal{B}_i}(h(x_t), \mathcal{Y})- \ell_{\mathcal{B}_i}(h(x_t'), \mathcal{Y}) } \right|$, $R_j^*$ denotes the best-in-class risk at $j$-time querying, and $|\mathcal{H}|$ denotes the element number of  $\mathcal{H}$. More details and proofs are presented in  Supplementary Material.
\end{theorem}
\cao{In summary, learning on small data could be derived from hypothesis-pruning under the PAC framework via i.i.d.  sampling. Therefore, the typical supervised or unsupervised strategies that can capture effective small data representation  could be adopted, \textit{e.g.}, including active learning, few-shot learning,   deep clustering.}
\section{A Geometric Perspective}
Considering that the aforementioned learning topics have been well surveyed such as  \cite{wang2020generalizing} and \cite{settles2009active}, we thus present another new perspective for small data representation: a structured geometric perspective. In this setting, learning on small data could be performed in the   Euclidean and non-Euclidean (hyperbolic) space. To learn effective geometric representation, we investigate the properties of the Euclidean and hyperbolic mean with respect to the unified expression of Fréchet mean \cite{lou2020differentiating}.
\begin{figure*}[htbp] \label{figure1}
  \centering
  \subfigure[Poincaré disk model and geodesics]{\includegraphics[width=0.238\textwidth]{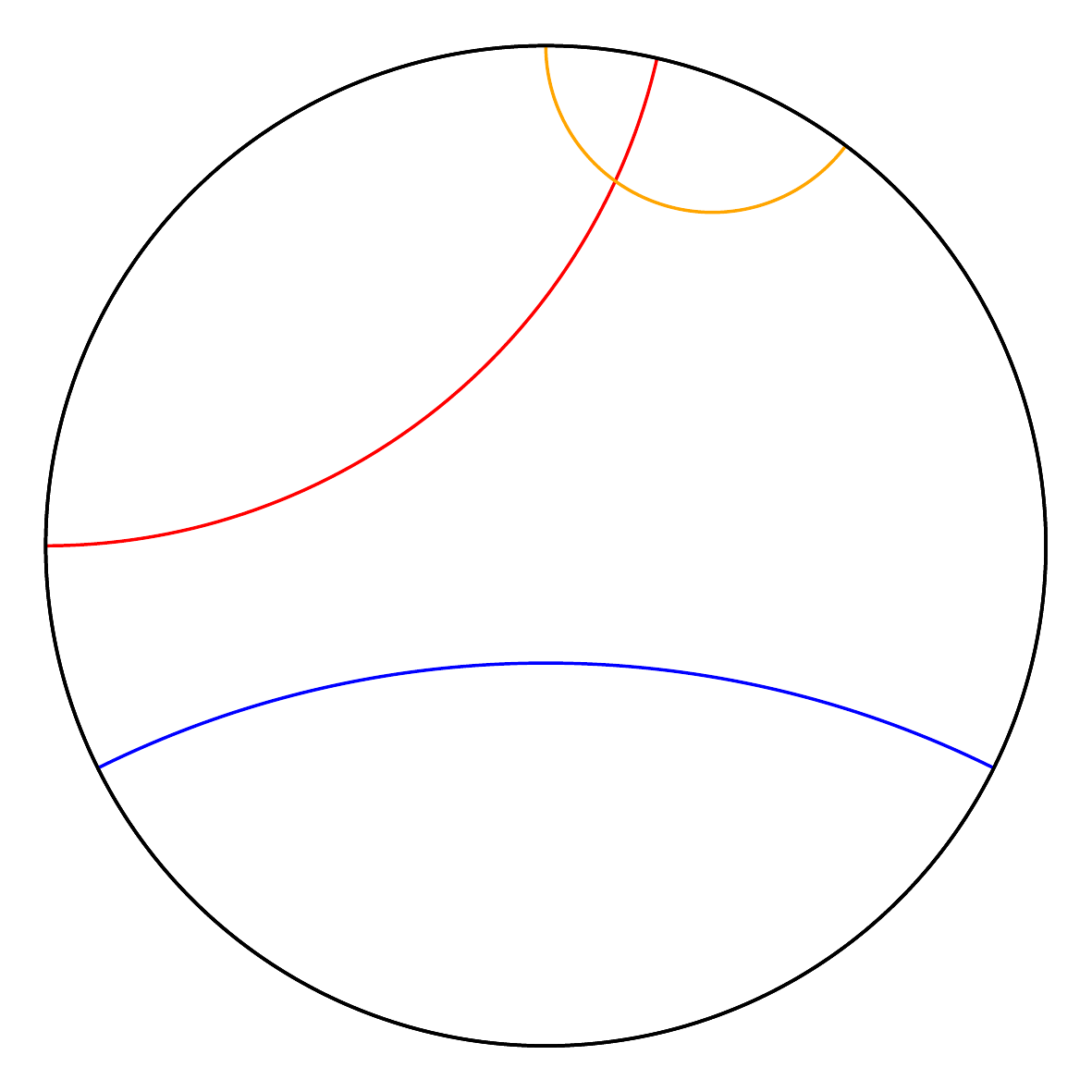}
    \label{fig:poincare}
  }\hspace{6mm}
  \subfigure[Poincaré ball model]{
    \includegraphics[width=0.25\textwidth]{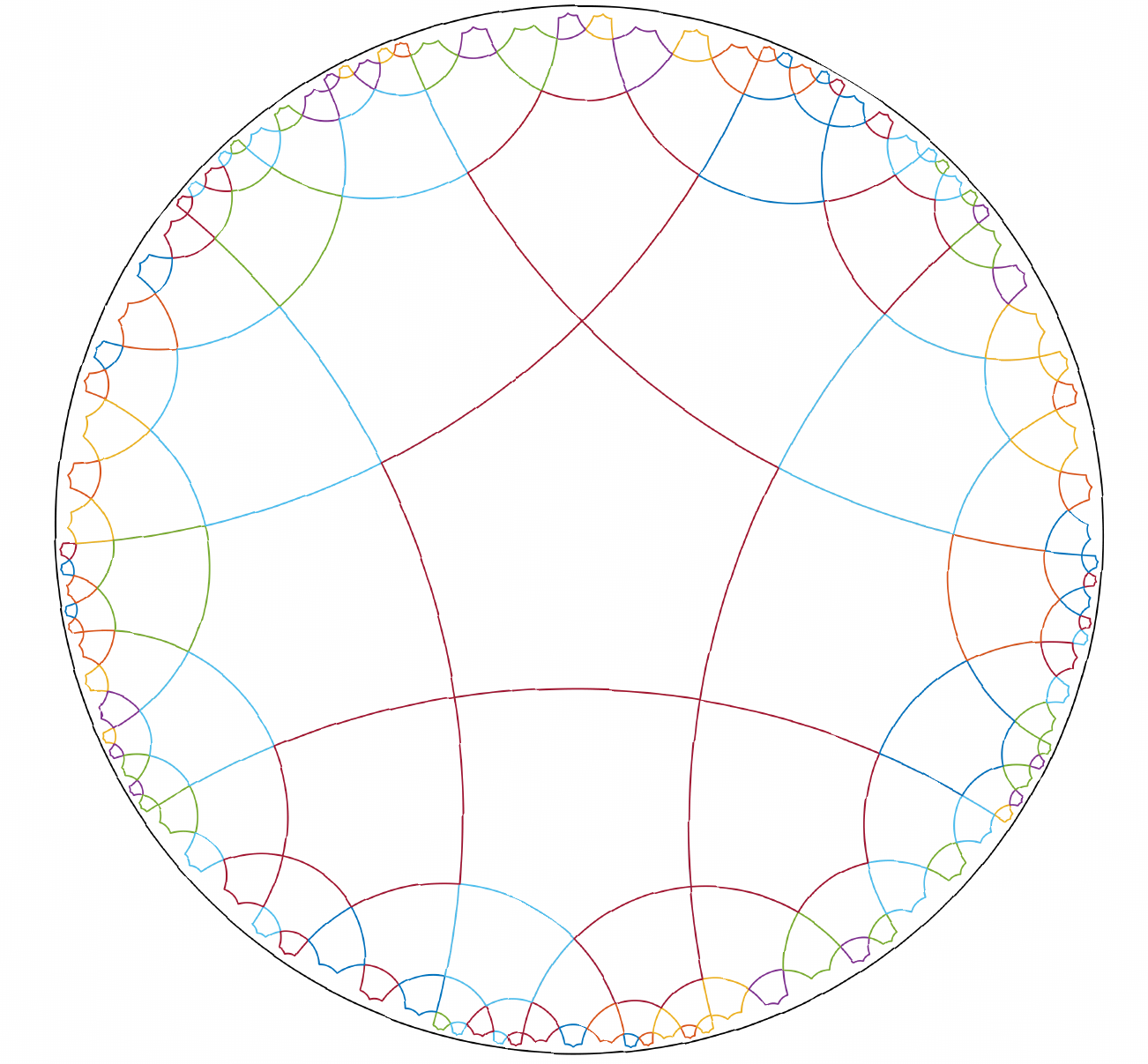}
      \label{fig:poincare}
  }\hspace{3mm}
  \subfigure[Lorentz model]{
    \includegraphics[width=0.34\textwidth]{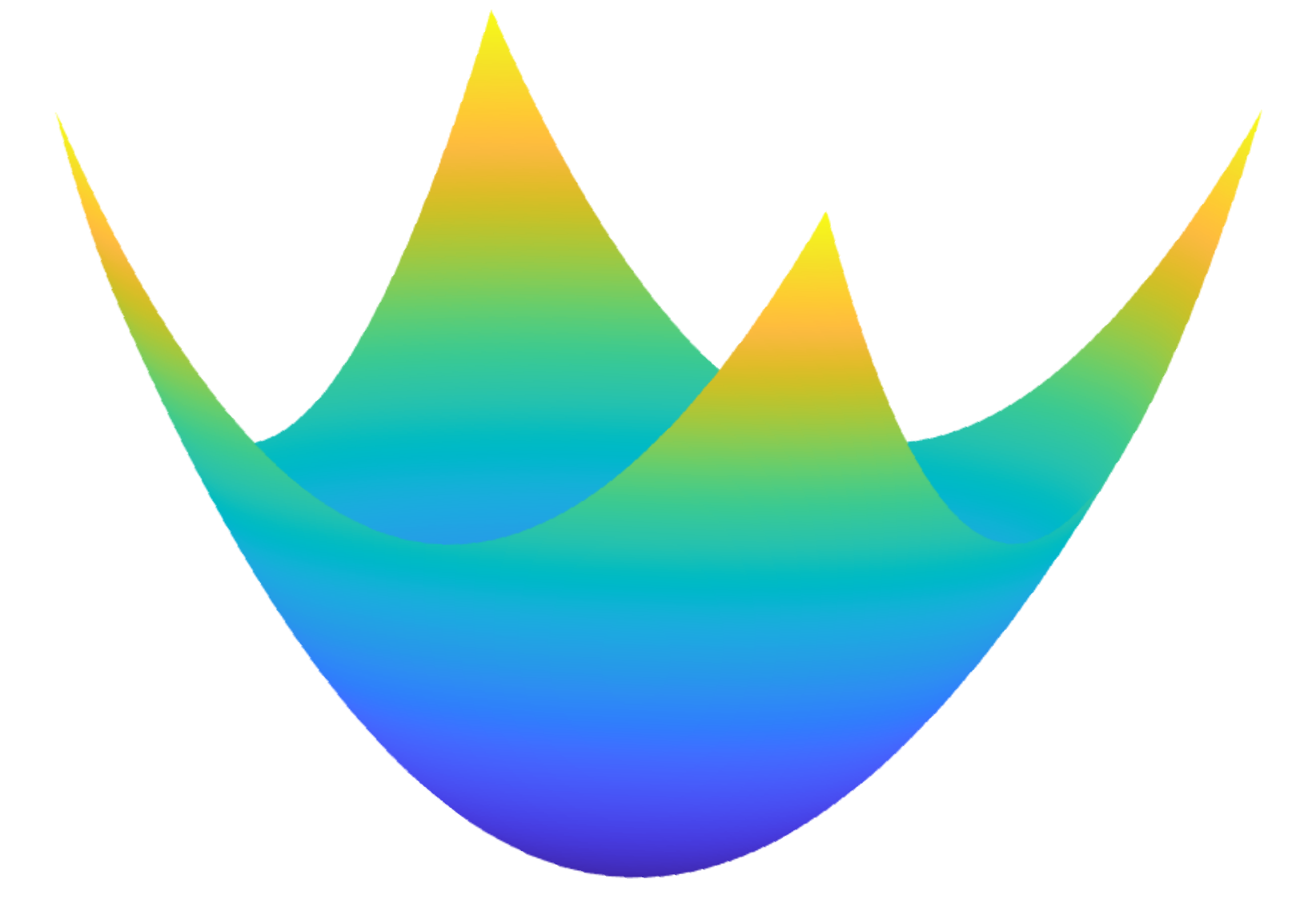}
      \label{fig:lorentz}
  }
  \caption{Illustration of Poincaré model and Lorentz model. (a) Poincaré disk model (2D Poincaré ball model) and geodesics. Poincaré disk model is an open unit circle and all of its geodesics are perpendicular to the boundary. (b) Poincaré ball model is an open unit ball. The closer two points inside the model are to the boundary, the more the distance between them grows infinitely. (c) Lorentz model. It is defined as the upper sheet of a two-sheeted n-dimensional hyperboloid.}
  \label{fig:two-models}
\end{figure*}

\subsection{Fréchet Mean}
To understand a collection of observations sampled from a statistical population, the \textit{mean} of observations is adopted as one powerful statistics to summarize the observations from an underlying distribution. What does \textit{mean} mean? It may vary under different data distributions and depend on the goal of the statistics.  To characterize the representation of real-world data, typical candidates for mean may be the arithmetic mean and the median, but in some cases, the geometric or harmonic mean may be preferable. When the data exists in a set without a vector structure, such as a manifold or a metric space, a different concept of \textit{mean} is required, \ie, Fréchet mean \cite{lee2006riemannian}.\\
\textbf{Fréchet mean on probability measures.}  We explore a general mean that can be defined with less structure but encompasses the common notions of \textit{mean} – the \textit{Fréchet mean}. The Fréchet mean is an important entailment (implication) in geometric representations, that embeds a ``centroid”  to indicate its local features (neighbors) on a metric space. For a distance space $(\mathcal{X},d_{\mathcal{X}})$, let $\mathbb{P}$ be a probability measure on $\mathcal{X}$ with $\int d_{\mathcal{X}}^2(x,y)\dif \mathbb{P}(x) < \infty $ for all $y\in\mathcal{X}$, the Fréchet mean is to operate the argmin optimization \cite{lou2020differentiating} of
\begin{equation}
\mu_{\mathcal{X}} = \argmin_{\mu \in \mathcal{X} } \int d_{\mathcal{X}}^2(x,\mu) \dif \mathbb{P}(x).
    \label{FM_pro}
\end{equation}
The Fréchet means defined by the probability measures is more generalized and can be derived to more common objects.\\
\textbf{Advantages of Fréchet mean.} The Fréchet mean w.r.t. Eq.~(\ref{FM_pro}) has two significant advantages  \cite{lou2020differentiating}. 1) It provides a common construction for many well-known notions of \textit{mean} in machine learning and thus implies many interesting properties of data. 2) It provides the notion of \textit{mean} in spaces with less structure than Euclidean spaces, \eg, metric spaces or Riemannian manifolds, thus widening the possibility of adopting machine learning methods in these spaces.\\
\textbf{Fréchet mean on Riemannian manifold.} We next observe the Fréchet mean on manifold.
For an arbitrary Riemannian manifold $\mathcal{M}$ with metric $g_x(\cdot,\cdot)$ that projects the tangent space $\mathcal{T}_x\mathcal{M}\times \mathcal{T}_x\mathcal{M}\rightarrow \mathbb{R}^n$ where $\|v\|_g=\sqrt{g_x(v,v)}$, let $\gamma(t):[a,b]\to \mathcal{M}$ be a geodesics  that stipulates distance  $d_{\mathcal{M}}(\cdot,\cdot)$ as the integral of the first order of the geodesics. For all $x,y\in\mathcal{M}$, the distance $d_{\mathcal{M}}(x,y):={\rm inf} \int_a^b \|\gamma '(t)\|_g\dif t$ where $\gamma(t)$ denotes any geodesics such that $\gamma(a)=x,\gamma(b)=y$. Given a point set $\mathcal{X}=\{x_1,x_2,...,x_m\}\subseteq\mathcal{M}$ and set the probability density of each point $x_i\in\mathcal{X}$ as $\frac{1}{m}$, the \textit{weighted Fréchet mean} \cite{lou2020differentiating} is to operate the argmin optimization  of
\begin{equation}
\begin{split}
\mu_\mathcal{M} = \argmin_{\mu \in \mathcal{M} } \sum_{i=1}^{m} \omega_i   d^2_\mathcal{M}(x_i,\mu),
\end{split}
\label{argmin}
\end{equation}
where  $\omega_i$ denotes the weight of $x_i$, and the constraint of $\mu \in \mathcal{M}$ stipulates that $\mu$ may converge in $\mathcal{M}$ with infinite candidates. Given  $d_\mathcal{M}(a,b):=\|a-b\|_2$ defined in Euclidean geometry and $\omega_i=1/m$ for all $i$, the weighted Fréchet mean is then simplified into the Euclidean mean, which results in a fast computational time. This specified setting  achieved promising results in the $k$-means clustering, maximum mean discrepancy optimization \cite{gretton2012kernel}, \emph{etc.} 

\subsection{Euclidean Mean}
The Euclidean mean is the most widely adopted \textit{mean} in machine learning. Meanwhile, the Euclidean mean is of great importance to perform aggregation operations such as attention \cite{zhou2020end}, 
batch normalization \cite{bjorck2018understanding}. 
Let $\mathcal{R}^n$ denote the Euclidean manifold with zero curvature, and the corresponding Eculidean metric of which is defined as $g^{E}=\mathrm{diag}([1,1, \ldots, 1])$. For $\textbf{x}, \textbf{y} \in\mathcal{R}^n$, the Euclidean distance is given as:
\begin{equation}
    d_{\mathcal{R}}(\textbf{x}, \textbf{y})=\| \textbf{x}-\textbf{y} \|_2.
\end{equation}
Then $(\mathcal{R}^n,d_{\mathcal{R}})$ is a complete distance space. 
We next present a formal description of Euclidean mean \cite{aloise2009np} based on the weighted Fréchet mean w.r.t. Eq.~(\ref{argmin}).

\begin{proposition} \label{proposition-em}
Given a set of points $\mathcal{X}=\{x_1,x_2,....,x_m\}\subseteq\mathcal{R}^n$, the Euclidean mean $\mu_{\mathcal{R}}$ minimizes the following problem:
\begin{equation}\label{equ:euclidean mean}
\begin{split}
\min_{\mu\in \mathcal{R}^{n} } \sum_{i=1}^{m} \omega_i         d^2_\mathcal{R}(x_i,\mu),
\end{split}
\end{equation}
where $\omega_i\geq0$ denotes the weight coefficient of $x_i$. 
\end{proposition}
The completeness of $(\mathcal{R}^{n},d_{\mathcal{R}})$ guarantees that Eq.~(\ref{equ:euclidean mean}) has a closed gradient, so a unique solution exists. With Proposition~\ref{proposition-em}, the Euclidean mean $\mu_{\mathcal{R}}$ is unique with the closed-form solution of
\begin{equation}
    \mu_{\mathcal{R}}=\frac{1}{m}\sum_{i=1}^m x_i.
\end{equation}


\subsection{Non-Euclidean  Mean}
Recent studies demonstrate that hyperbolic geometry has stronger expressive capacity than the Euclidean geometry to model hierarchical features \cite{chami2019hyperbolic}, \cite{liu2019hyperbolic}. Meanwhile, the Euclidean mean extends naturally to the Fréchet mean on hyperbolic geometry. Next, we discuss the Fréchet mean on the Poincaré  and  Lorentzian model with respect to Riemannian manifold. The illustrations of Poincaré  and Lorentz model are presented in Figure~\ref{fig:two-models}.
\subsubsection{Poincaré Centroid}
The Poincaré ball model $\mathcal{P}^{n}$ with constant negative curvature corresponds to the Riemannian manifold $(\mathcal{P}^n$, $\left.g_x^{\mathcal{P}}\right)$ \cite{peng2021hyperbolic}, where $\mathcal{P}^n=\left\{x \in \mathbb{R}^{n}:\|x\|<1\right\}$ denotes an open unit ball defined as the set of $n$-dimensional vectors whose Euclidean norm are smaller than $1$. The Poincaré metric  is defined as $g_{x}^{\mathcal{P}}=\lambda_{x}^{2} g^{E}$, where $\lambda_{x}=\frac{2}{1-\|x\|^{2}}$ denotes the conformal factor and $g^{E}$ denotes the Euclidean metric. For any $\textbf{x},\textbf{y}$ in $\mathcal{P}^n$, the Poincaré distance is given as \cite{nickel2017poincare}:
\begin{equation}
   d_{\mathcal{P}}(\textbf{x}, \textbf{y})=\cosh ^{-1}\left(1+2 \frac{\|\textbf{x}-\textbf{y}\|^{2}}{\left(1-\|\textbf{x}\|^{2}\right)\left(1-\|\textbf{y}\|^{2}\right)}\right). 
\end{equation}
Then $(\mathcal{P}^n,d_{\mathcal{P}})$ is a distance space. We thus have the following proposition for the Poincaré centroid \cite{nickel2017poincare} based on the Poincaré distance.

\begin{proposition}
Given a set of points $\mathcal{X}=\{x_1,x_2,....,x_m\}\subseteq\mathcal{P}^n$, the Poincaré centroid $\mu_{\mathcal{P}}$ minimizes the following problem:
\begin{equation}
    \min_{\mu\in \mathcal{P}^{n}} \sum_{i=1}^m \omega_i d^2_\mathcal{P}(x_i,\mu),
\end{equation}
where $\omega_i\geq0$ denotes the weight coefficient of $x_i$.
\end{proposition}
There is no closed-form solution for the Poincaré centroid $\mu_\mathcal{P}$, so Nickel \emph{et al.} \cite{nickel2017poincare} computes it via gradient descent.

 \subsubsection{Lorentzian Centroid}
\textbf{Lorentz Model.} The Lorentz model $\mathcal{L}^{n}$ \cite{peng2021hyperbolic} with constant curvature $-1 / K$ avoids numerical instabilities that arise from the fraction in the Poincaré metric,  
for $\textbf{x}, \textbf{y} \in \mathbb{R}^{n+1}$, the Lorentzian scalar product is formulated as \cite{law2019lorentzian}:
\begin{equation}
\left<\textbf{x}, \textbf{y} \right>_\mathcal{L} = -x_0y_0 + \sum_{i=1}^n x_iy_i \leq -K.
\end{equation}
This model of $n$-dimensional hyperbolic space corresponds to the Riemannian manifold ($\mathcal{L}$, $g^\mathcal{L}_\textbf{x}$), 
where $\mathcal{L} = \{\textbf{x} \in \mathbb{R}^{n+1} : \left<\textbf{x}, \textbf{y} \right>_\mathcal{L} = -K, x_0 > 0\}$ (\textit{i.e.}, the upper sheet of a two-sheeted n-dimensional hyperboloid)
and $g^\mathcal{L}_\textbf{x} = \mathrm{diag}([-1, 1, ..., 1])$ denotes the Lorentz metric. The squared Lorentz distance for $\textbf{x}, \textbf{y} \in \mathbb{R}^{n+1}$ which satisfies all the axioms of a distance other than the triangle inequality is defined as \cite{law2019lorentzian}:
\begin{equation}
\begin{split}
d_\mathcal{L}^2(\textbf{x}, \textbf{y}) = \|\textbf{x}-\textbf{y}\|^2_\mathcal{L} = -2K - 2\left<\textbf{x}, \textbf{y}\right>_\mathcal{L}.
\end{split}
\end{equation}
Proposition~\ref{Lorentzian-Centroid} presents Lorentz centroid that represents the aspherical distributions under a Lorentz model\cite{law2019lorentzian}. 

\begin{proposition}\label{Lorentzian-Centroid} Given a set of points $\mathcal{X}=\{x_1,x_2,....,x_m\}\subseteq\mathcal{L}^n$, the Lorentzian centroid $\mu_{\mathcal{L}}$ minimizes the following problem:
\begin{equation}
    \min_{\mu\in \mathcal{L}^{n}} \sum_{i=1}^m \omega_i d^2_\mathcal{L}(x_i,\mu),
\end{equation}
where $\omega_i\geq0$ denotes the weight coefficient of $x_i$.
\end{proposition}
The Lorentzian centroid $\mu_{\mathcal{L}}$ is unique with the closed-form solution of
\begin{equation}\label{eq:lorentzian-center}
\mu_{\mathcal{L}} = \sqrt{K} \frac{\sum_{i=1}^m \omega_i x_i}{\left|\|\sum_{i=1}^m \omega_i x_i\|_\mathcal{L}\right|},
\end{equation}
where $\left|\|\mathbf{a}\|_{\mathcal{L}}\right|=\sqrt{\left|\|\mathbf{a}\|_{\mathcal{L}}^{2}\right|}$ denotes the modulus of the imaginary Lorentzian norm of the positive time-like vector $\mathbf{a}$.
\subsubsection{Lorentzian Focal Point}
With \cite{9468944}, the Euclidean norm of Lorentz centroid $\mu_{\mathcal{L}}$ decreases, thus yielding an effective approximation to the focal point which is more representative than Lorentz centroid for the aspherical distributions. However, the approximation cannot only depend on $K$ due to uncertain parameter perturbations. We should also control the coefficient $\omega_i$ to approximate the Lorentzian focal point. Here $\omega_i$ w.r.t. Eq.~(\ref{eq:lorentzian-center}) can be written as \cite{9468944}:
\begin{equation}\label{eq:vi}
\begin{split}
\omega_i = \frac{d^2_\mathcal{L}(x_i, \mu)}{\sum_{i=1}^m d^2_\mathcal{L}(x_i, \mu)},
\end{split}
\end{equation}
Then we present the approximation of Lorentzian focal point in Proposition \ref{Lorentzian-Focal-Point} \cite{9468944}.
\begin{proposition}\label{Lorentzian-Focal-Point}
Given a set of points $\mathcal{X}=\{x_1,x_2,....,x_m\}\subseteq\mathcal{L}^n$, the Lorentzian focal point $\mu_{\mathcal{F}}$ minimizes the following problem:
\begin{equation}
   \min _{\mu \in \mathcal{L}^{n}} \sum_{i=1}^m \omega_i \left<x_i, \mu \right>_\mathcal{L}.
\end{equation}
Then the Lorentzian focal point $\mu_{\mathcal{F}}$ can be approximately given as:
\begin{equation}
\mu_{\mathcal{F}} = \sqrt{K} \frac{\sum_{i=1}^m \omega_i x_i}{|\|\sum_{i=1}^m \omega_i x_i\|_\mathcal{L}|},
\end{equation}
where $\omega_i \geq 0$ follows Eq.~(\ref{eq:vi}). 
\end{proposition}

\subsection{Kernel Mean} 
The kernel mean \cite{muandet2016kernel} could be generalized in Euclidean and hyperbolic geometry, which presents a kernel expression for the geometric mean with respect to  the probability measures of Fréchet mean. 

We first review some properties of reproducing kernel Hilbert space (RKHS). Let $\mathcal{H}$ denote a RKHS over $\mathcal{X}$, then every bounded linear functional is given by the inner product with a unique vector in $\mathcal{H}$ \cite{alt2016linear}. For any $x\in\mathcal{X}$, there exists a unique vector $k_x\in\mathcal{H}$ such that $f(x)=\Braket{f, k_x}$ for every $f \in \mathcal{H}$. The function $k_x=K(x,\cdot)$ is called the \textit{reproducing kernel} for point $x$, where $K(x_1,x_2):\mathcal{X}\times\mathcal{X}\to\mathbb{R}$ is positive definite. For any $k_x,k_y$ in $\mathcal{H}$, the Hilbert distance is given as \cite{gretton2012kernel}:
\begin{equation}
    d_{\mathcal{H}}(k_x,k_y)=\|  k_x-k_y \| = \sqrt{\Braket{k_x-k_y,k_x-k_y}}.
\end{equation}
Then $(\mathcal{H},d_{\mathcal{H}})$ is a complete distance space. Following Eq.~(\ref{FM_pro}), Proposition~\ref{pro:kernel mean} gives a formal description of kernel mean \cite{muandet2016kernel}.
\begin{proposition}\label{pro:kernel mean}
Given a separable RKHS $\mathcal{H}$ endowed with a measurable reproducing kernel $K(x_1,x_2):\mathcal{X}\times\mathcal{X}\to\mathbb{R}$ such that $\int d_{\mathcal{H}}^2(k_x,k_y)\dif \mathbb{P}(x) < \infty$ for all $y\in\mathcal{X}$, where $\mathbb{P}$ denotes a probability measure on $\mathcal{X}$. Then the kernel mean $\mu_{\mathcal{H}}$ minimizes the following problem:
\begin{equation}\label{equ:Kernel Fréchet mean}
        \min_{\mu \in \mathcal{H}} \int d_{\mathcal{H}}^2(k_x,\mu) \dif \mathbb{P}(x).
\end{equation}
\end{proposition}
Based on the completeness of the distance space $(\mathcal{H},d_{\mathcal{H}})$, the following theorem gives the solution of kernel mean, which is consistent with the classical kernel mean defined in \cite{muandet2016kernel}.
\begin{theorem}\label{thm:kernel mean}
The kernel mean $\mu_{\mathcal{H}}$ is unique with the closed-form solution:
\begin{equation}
    \mu_{\mathcal{H}}=\int K(x,\cdot) \dif \mathbb{P}(x),
\end{equation}
where $K(x,\cdot)$ indicates that the kernel has one argument fixed at $x$ and the second free.
\end{theorem}
Theorem~\ref{thm:kernel mean} includes kernel mean in Fréchet mean for the first time, thus maintaining formal uniformity with other standard means, \textit{e.g.} Euclidean mean. More details and proofs are presented in Supplementary Material.

\section{Optimization Solvers}
In order to explore the optimization solvers for the above geometric representations of Euclidean and non-Euclidean paradigms, we category these solvers into three gradient-based methods: Euclidean gradient for the optimization of Euclidean geometric paradigms, Riemannian gradient for optimization of hyperbolic geometric paradigms and Stein gradient for optimization of both Euclidean and hyperbolic geometric paradigms, the details of which are presented below. Note that we collect these optimization paradigms for  learning on   small data; meanwhile, they can also be adopted for big data issues.
\subsection{Euclidean Gradients}
Stochastic Gradient Descent (SGD) \cite{ruder2016overview} is an effective approach to find the local minima of a cost function, it can be adopted to optimize Euclidean centroids which are formulated as an argmin problem in Euclidean space. \\
\textbf{Stochastic gradient descent.} Given a minimization problem of $\min \limits_{x\in \mathbb{R}^{n} } J(x)$ in Euclidean space, at $t$-time, the parameters $x_t$ is updated as \cite{ruder2016overview}:
\begin{equation}
\begin{split}
x_{t + 1} = x_t - \eta \cdot \nabla_{x}J(x),
\end{split}  
\end{equation}
where $J(x)$ denotes the cost function parameterized by $x$, and $\eta$ denotes the learning rate.
\subsection{Non-Euclidean Gradients}
Manifold optimization \cite{boumal2020introduction} aims to seek solutions for various constrained optimization problems in Euclidean space by transforming these problems into unconstrained optimization problems on Riemannian manifolds. Correspondingly, Riemannian  Gradient Descent (RGD) \cite{liu2018riemannian}, \cite{chen2021decentralized} is introduced to perform iterative optimization. With the scheme, Riemannian optimization domain reaps rapid development. Not surprisely, hyperbolic geometry also adopt RGD to optimize different paradigms on Poincaré ball $\mathcal{P}^{n}$ and Lorentz model $\mathcal{L}^{n}$.\\
 \textbf{Riemannian gradient descent.} Given a minimization problem of $\min \limits_{x \in \mathcal{M}} J(x)$ on a Riemannian  manifold $\mathcal{M}$,   $x_t$ at $t$-time  is updated by the  exponential map  ${\rm exp}_x$  \cite{chen2021decentralized}: 
\begin{equation}
\begin{split}
 x_{t+1}={\rm exp}_{x_t}\Big(-\eta J'(x_t)\Big),
\end{split}
\end{equation}
 where $J'(x_t)$ denotes the Riemannian gradient on the tangent space $\mathcal{T}_x \mathcal{M}$ and $\eta$ denotes the learning rate. \\
\textbf{Exponential map on Poincaré model.} Given a Riemannian metric $g_{x}(\cdot,\cdot)$ that induces an inner product $\langle u,v\rangle:=g_x(u,v)$ on tangent space $\mathcal{T}_x\mathcal{M}$. For each point $x\in\mathcal{M}$ and vector $u\in \mathcal{T}_x\mathcal{M}$, there exists a unique geodesic $\gamma: [0,1]\rightarrow \mathcal{M}$ where $\gamma(0)=x,\gamma'=u$.
The exponential map ${\rm exp}_{x} : \mathcal{T}_x\mathcal{M} \rightarrow \mathcal{M}$ is defined as ${\rm exp}_{x}(u)=\gamma(1)$, where $d_\mathcal{P}(x, {\rm exp}_{x}(u))=\sqrt{g_\mathcal{M}(u,u)}$. With  \cite{lou2020differentiating},
  \begin{equation}
\begin{split}
{\rm exp}_{x}(u)=\frac{(1-2\langle x,z\rangle_2-\|z\|^2)x+(1+\|x\|^2)z}{1-2\langle x,z\rangle_2+\|x\|^2\|z\|^2}, 
\end{split}
\end{equation}
 where $z={\rm tanh}( \frac{\|u\|^2}{1+\|x\|^2})\frac{u}{\|u\|}$. \\
\textbf{Exponential map on Lorentz model.} With Proposition~3.2 of  \cite{chami2019hyperbolic}, ${\rm exp}_{x}(u)=\gamma(1)$ on a Lorentz model $\mathcal{L}^{n}$ is expressed as  
 \begin{equation}
\begin{split}
{\rm exp}_{x}(u)={ \rm cosh}(\|u\|_\mathcal{L})x+u\frac{{\rm sinh}(\|u\|_\mathcal{L})}{\|u\|_\mathcal{L}}.
\end{split}
\end{equation}
\subsection{Stein Gradient}
Bayesian inference \cite{box2011bayesian} is a statistical inference that invokes the Bayes theorem to approximate the probability distribution. Variational inference \cite{blei2017variational} approximates parameterized distribution through  probabilistic optimization that involves sampling tractable variables, such as Markov Chain Monte-Carlo (MCMC). However, the approximation errors of bayesian and variational inference on estimation over likelihoods or posterior parameter distribution are not easy to control, resulting in unstatistically significant results with calibration. To tight the approximation, Liu \emph{et al.} \cite{chwialkowski2016kernel} adopt the Stein operation which controls the bounds on the distance between two probability distributions in a given probability metric. With such proposal, Liu \emph{et al.} then propose the Stein Variational Gradient Descent (SVGD) algorithm \cite{liu2019stein} that minimizes the KL divergence \cite{hershey2007approximating} of two probability distributions $p$ and $q$ by utilizing Kernelized Stein Discrepancy (KSD) and smooth transforms, thus conducting iterative probability distribution approximation. 
\par In detail, MCMC estimates the denominator integral of posterior distribution by sampling, thus bringing the problem of computational inefficiencies. Let $p_0(x)$ be the prior, $\{D_k\}$ be a set of i.i.d. observations, and $\Omega=\{q(x)\}$ be the distribution set, variational inference adopts a novel idea to alleviate this by minimizing the KL Divergence between the target posterior distribution $p(x)$ and another distribution $q^*(x)$ so as to approximate $p(x)$:
\begin{equation}\label{eq:vi-kl}
\begin{split}
{{q^*(x)}} & = \argmin \limits_{q(x) \in \Omega} \{{\rm KL}(q(x) || p(x))\equiv \mathcal{W}\},
\end{split}
\end{equation}
where $\overline p(x) := p_0(x) \prod \limits_{i=1}^N p(D_i | x)$, ${Z= \int {\overline p(x)}\dif x}$ denotes the normalization constant which requires complex calculations, $\mathcal{W} = \mathbbm{E}_q [\log{q(x)}] - \mathbbm{E}_q [{\rm log} {\overline p(x)}] + \log Z$. Hence, to circumvent the computation of troublesome normalization constant $Z$ and seek for a general purpose bayesian inference algorithm, Liu \emph{et al.} adopt the Stein methods and propose the SVGD algorithm. More details are presented in Supplementary Material.
\par Given notions of Stein’s Identity (Eq.~(\ref{eq:SI})), Stein Discrepancy (Eq.~(\ref{eq:SD})) and Kernelized Stein Discrepancy (Eq.~(\ref{eq:KSD})) of the Stein methods, Liu \emph{et al.} rethink the goal of variational inference which is defined in Eq.~(\ref{eq:vi-kl}), they consider the distribution set $\Omega$ could be obtained by smooth transforms from a tractable reference distribution $q_0(x)$ where $\Omega$ denotes the set of distributions of random variables which takes the form $r = T(x)$ with density:
\begin{equation}\label{eq:density}
\begin{split}
q_{[T]}(r) = q(\mathcal{R}) \cdot |\det(\nabla_r \mathcal{R})|,
\end{split}
\end{equation}
where $T: \mathcal{X} \to \mathcal{X}$ denotes a smooth transform,  $\mathcal{R}=T^{-1}(r)$ denotes the inverse map of $T(r)$ and $\nabla_r \mathcal{R}$ denotes the Jacobian matrix of $\mathcal{R}$. With the density, there should exist some restrictions for $T$ to ensure the variational optimization in Eq.~(\ref{eq:vi-kl}) feasible. For instance, $T$ must be a one-to-one transform, its corresponding Jacobian matrix should not be computationally intractable. Also, with \cite{rezende2015variational}, it is hard to screen out the optimal parameters for $T$.
\par Therefore, to bypass the above restrictions and minimize the KL divergence in Eq.~(\ref{eq:vi-kl}), an incremental transform ${T(x) = x + \varepsilon \varphi(x)}$ is proposed, where $\varphi(x)$ denotes the smooth function controlling the perturbation direction and $\varepsilon$ denotes the perturbation magnitude. With the knowledge of Theorem \ref{Stein-KL} and Lemma \ref{Stein-varphi}, how can we approximate the target distribution $p$ from an initial reference distribution $q_0$ in finite transforms with  $T(x)$? Let $s$ denote the total distribution number, an iterative procedure which can obtain a path of distributions $\{q_t\}^s_{t=1}$ via Eq.~(\ref{eq:iterative-procedure}) is adopted to answer this question:
\begin{equation}\label{eq:iterative-procedure}
\begin{split}
& \quad \quad q_{t+1} = q_t[T^*_t],\\
& T^*_t(x) = x + \varepsilon_t \varphi^*_{q_t, p}(x),
\end{split}
\end{equation}
where $T^*_t$ denotes the transform direction at iteration $t$, which then decreases the KL Divergence with $\varepsilon_t {{\rm KSD}(q_t, p)}$ at {$t$-th} iteration. Then, the distribution $q_t$ finally converges into the target distribution $p$.
To perform above iterative procedure, Stein Variational Gradient Descent (SVGD) adopts the iterative update procedure for particles which is presented in Theorem \ref{Stein-iteration} to approximate $\varphi^*_{q, p}$ in Eq.~(\ref{eq:varphi}).
\begin{theorem}\label{Stein-iteration}
Let $p(x)$ denote the target distribution, $\{x^0_i\}^m_{i=1}$ denote the initial particles \cite{liu2019stein}. 
Also, at iteration $t$, let ${\vartheta} = \nabla_{x^t_j} {\log} p(x^t_j)$, ${\mu}(x^t_j, x)=\nabla_{x^t_j} k(x^t_j, x)$ denote a regular term, $\Phi$ denote $\varepsilon_t \hat{\varphi}^*(x)$, the particles set are updated iteratively with $T^*_t$ defined in Eq.~(\ref{eq:iterative-procedure}), in which the expectation under $q_t$ in $\varphi^*_{q_t, p}$
is approximated by the empirical mean of $\{x^t_i\}^m_{i=1}$:
\begin{equation}
\begin{split}
x^{t}_i + \Phi \rightarrow x^{t + 1}_i,
\end{split}
\end{equation}
where 
\begin{equation}
\begin{split}
\hat{\varphi}^*(x) = \frac{1}{m} \sum_{j=1}^{m} [k(x^t_j, x) {\vartheta} + {\mu(x^t_j, x)}].
\end{split}
\end{equation}
\end{theorem}
Regarding $\hat{\varphi}^*(x)$, the first term $k(x^t_j, x) {\vartheta}$ denotes the weighted sum of the gradients of all the points weighted by the kernel function, which follows a smooth gradient direction to drive the particles towards the probability areas of $p(x)$; The second term $\mu(x^t_j, x)$ denotes a regular term to prevent the collapse of points into local modes of $p(x)$, \textit{i.e.}, pushing $x$ away from $x^t_j$.
\section{Learning on Small Data Representation}
\begin{figure*}[ht]
    \centering
    \includegraphics[scale=0.5]{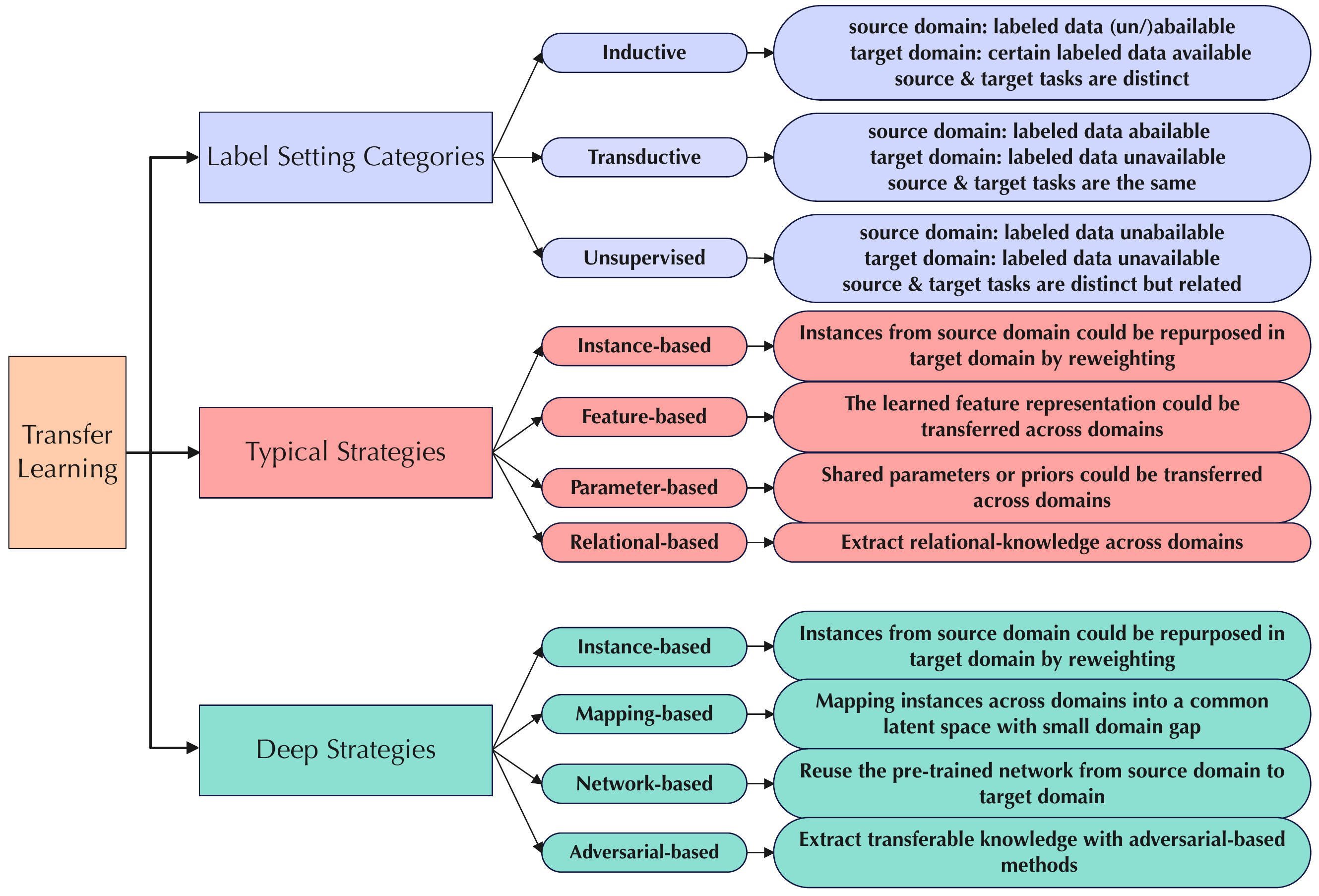}
    \caption{A simple summary of transfer learning.}
    \label{fig:transfer-learning}
\end{figure*}
With the increasing demand of learning on small data, we explore to facilitate the model learning on the efficient small data representation under multiple potential learning communities, which benefits the Transfer Learning, Reinforcement Learning, Contrastive Learning, and Graph Representation Learning for more robust and efficient training. In this section, we introduce these learning topics and explain the potentiality of learning on small data for them.
\subsection{Transfer Learning on Small data}
Most of machine learning topics are based on a common assumption: the training and testing data follow the same distribution. However, the assumption may not hold in many real-world scenarios. Transfer learning \cite{5288526} looses the constraint of the assumption, \textit{i.e.}, the training and testing data could be drawn from different distributions or domains. It aims to mining domain-invariant features and structures between domains to conduct effective data and knowledge transfer, and finally generalize to unseen domains. Specifically, transfer learning tries to improve the model's capability in target domain by leveraging the knowledge learned from source domain, \textit{e.g.}, transferring the knowledge of riding a bike to driving a car.  
\par With \cite{5288526}, one of the core issues of transfer learning is that: \emph{What cross-domain knowledge could be migrated to improve the performance of models in target domains?} Although existing solutions can answer this question efficiently in some specific scenarios, there exists few general data-driven ones. Yet, learning on small data provides an appreciable paradigm from the perspective of data-efficient representation for crossing-domain knowledge exploitation in transfer learning. Moreover, learning on small data may help harvest more efficient and robust models due to its powerful representation capability. Besides, there may exist noisy or perturbed data to be transferred from source domains, which may degenerate the performance of models in target domains, while learning on small data may help eliminate such untrustworthy data to be transferred, thus improving the robustness and generalizability of models. Specifically, we start from the definition of transfer learning \cite{5288526}:
\begin{definition}
Transfer Learning. Let $D_{s}$ denote a source domain with learning task $T_{s}$, $D_{t}$ denote a target domain and its corresponding learning task is denoted by $T_{t}$. Transfer learning intends to improve the learning performance of the target predictive function $f_{t}$ in $D_{t}$ with the knowledge learned from $D_{s}$ and $T_{s}$, where $D_{s} \neq D_{t}$, or $T_{s} \neq T_{t}$. 
\end{definition}
Based on different conditions between the source and target domains, we provide a simple summary of transfer learning from the perspective of label setting categories, typical strategies and deep strategies in Figure \ref{fig:transfer-learning}. Specifically, we summarize the popular strategies of transfer learning based on the type of data to be transferred. Furthermore, deep learning has  been widely explored in transfer learning to leverage the knowledge obtained from source domains for deep neural networks. Formally, deep transfer learning \cite{jiang2022transferability} is defined as follows.


\begin{definition}
Deep Transfer Learning. Let $T_{s}$ be a deep learning task in source domain $D_{s}$, $T_{t}$ be a 
 deep learning task in target domain $D_{t}$,  deep transfer learning is to improve the   performance of
 a deep neural network with a non-linear target  predictive function $f_{t}$  in $D_{t}$ by extracting knowledge from $D_{s}$ and $T_{s}$, where $D_{s} \neq D_{t}$, or $T_{s} \neq T_{t}$. 
\end{definition}
It is of great potential to introduce learning on small data to transfer learning \cite{raghu2019transfusion}. For instance, in typical transfer learning, the significant and informative small data obtained from source domain could be utilized to perform instance transfer by re-weighting \cite{xia2013instance}; By analogy, in deep transfer learning, we could also conduct instance transfer with the efficient small data representation by deep neural networks as well. Moreover, in different scenarios of feature-based transfer learning, the domain-invariant features may also be efficiently extracted with small data learning methods \cite{gong2013connecting}, \cite{li2018domain}, \cite{phung2021learning}. With this, the generalization ability of models could be enhanced accross domains. Similarly, in the context of deep transfer learning, extracting partial instances in the source domain of instance-based deep transfer learning, reusing partial network which is pre-trained in the source domain of network-based deep transfer learning and extracting transferable representations \cite{zhuang2015supervised} applicable to both the source domain and the target domain  in deep transfer learning can reap better performance with the power of learning on small data. It is noteworthy that transfer learning may not always produce a positive transfer, which is known as $\textit{negative transfer}$ \cite{jiang2022transferability}. Learning on small data could attempts to avoid the issue by recognizing and reject harmful knowledge during transfer, meanwhile, extract an efficient data representation for transfer learning. In conclusion, learning on small data could be adopted to various transfer learning scenarios, it still awaits in-depth research.

\subsection{Reinforcement Learning on Small data}
Reinforcement learning (RL) \cite{franccois2018introduction} is an AI paradigm which focus on addressing sequential decision-making problems such as games, robotics, autonomous driving. It emphasizes on maximizing the desired benefits by rewarding expected actions and/or punishing unexpected ones in an interactive environment. In RL, an $agent$ interacts with its $environment$: $agent$ could perceive the state of $environment$ and reward the feedbacks from $environment$, thus making sound decisions (\textit{i.e.}, take an action). 
\par With \cite{dulac2021challenges}, numerous research work in RL cannot be easily leveraged in real-world systems because the assumptions of these work are rarely satisfied in practice, resulting in many critical real-world challenges, and one of which is: learning on the real system from limited samples. Besides, it is noteworthy that existing RL strategies hold low sample efficiency, which results in large amount of interactions with $environment$ \cite{yu2018towards}. In additional, the policy learning is required to be data-efficient. In the case, the demand of providing an efficient proposal is becoming more urgent. learning on small data emerges as a feasible and data-efficient solution for this. Specifically, it may help extract valuable prior knowledge from previously collected interaction data, it enables us to pre-train and deploy $agents$ capable of learning efficiently. In addition, efficient data extracted by learning on small data contributes to the model obtaining better robustness and generalizability. Therefore, learning on small data could be considered as a novel data-driven paradigm for reinforcement learning. Let us take a breif look at the value-based, policy-based and actor-critic strategies \cite{arulkumaran2017deep} in RL. \\
\textbf{Value-based strategies.}
\begin{figure}[ht]
    \centering
    \includegraphics[scale=0.32]{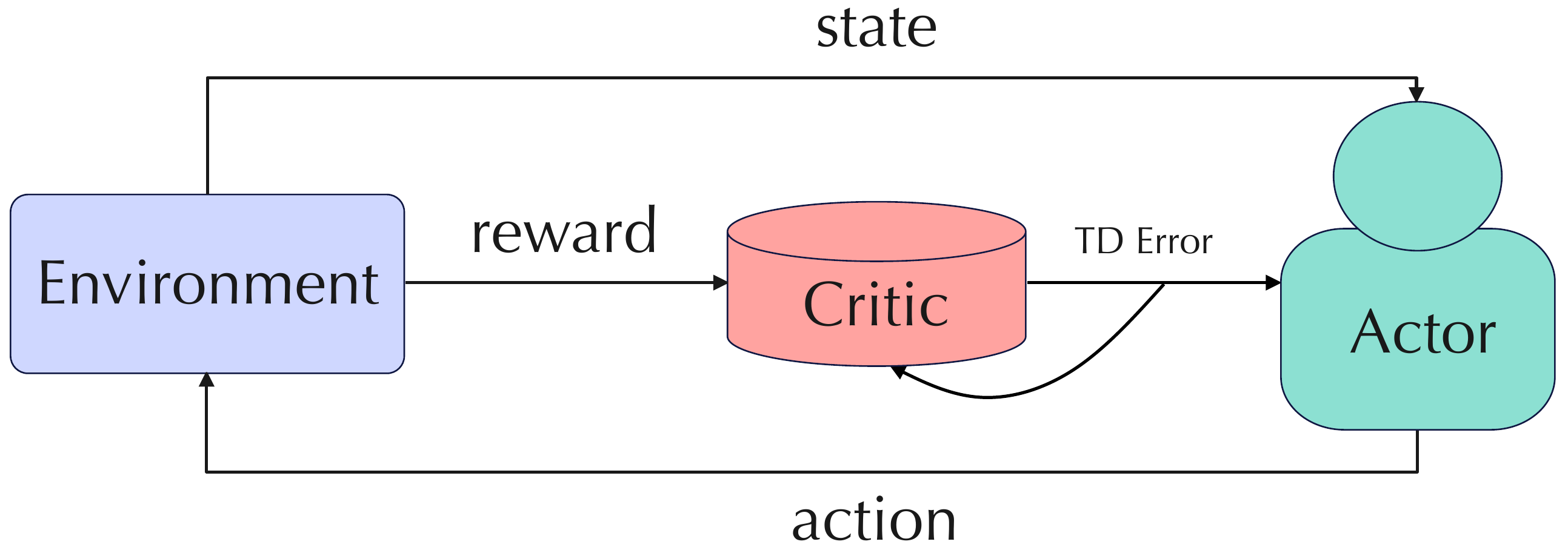}
    \caption{The illstration of actor-critic strategies in reinforcement learning. There are three objects in the scheme: Environment, Critic (value function) and Actor (policy). Environment interacts with Critic and Actor by providing rewards to Critic, states to both Critic and Actor. Then, Actor conducts an action refer to the provided state, meanwhile, Critic learns about and criticizes the action conducted by Actor. TD Error is introduced to evaluate the critique process and drives the updates of both Critic and Actor.}
    \label{fig:RL-AC}
\end{figure}
Value-based strategies \cite{zang2020metalight}, \cite{bell2021reinforcement} are introduced for estimating the expected under a specifed state. The state-value function $V_{\mathbf{\pi}}(\mathbf{s})$ which denotes the expected return given state ${\mathbf{s}}$ and policy $\mathbf{\pi}$ is formalized as:
\begin{equation}
\begin{split}
V_{\pi}(\mathbf{s}) = \mathbb{E}[\mathcal{R} |{\mathbf{s}}, \mathbf{\pi}].
\end{split}
\end{equation}
Since $V_{\pi}(\mathbf{s})$ is designed for evaluating policy $\mathbf{\pi}$, all polices can be evaluated to obtain the optimal policy ${\mathbf{\pi}}^{\ast}$ and the corresponding state-value function $V_{{\mathbf{\pi}}^{\ast}}(\mathbf{s})$ is defined as:
\begin{equation}
\begin{split}
V_{{\mathbf{\pi}}^{\ast}}(\mathbf{s}) = \max \limits_{\pi} V_{\pi}(\mathbf{s}), \quad {\forall} \mathbf{s} \in {\rm S},
\end{split}
\end{equation}
where ${\rm S}$ denotes the state set.
Hence, we can obtain ${\pi}^{\ast}$ by greedy search among all actions at $\mathbf{s_{t}}$ and taking the action $\mathbf{a}$ which maximizes the following objective:
\begin{equation}
\begin{split}
{\rm O} = {\mathbb{E}}_{{\mathbf{s}}_{t+1} \sim \mathcal{T}({\mathbf{s}}_{t+1} | {\mathbf{s}}_t, {\mathbf{a}})}[V_{{\mathbf{\pi}}^{\ast}}({\mathbf{s}}_{t+1})],
\end{split}
\end{equation}
where $\mathcal{T}({\mathbf{s}}_{t+1} | {\mathbf{s}}_t, {\mathbf{a}})$ is the so-called \textit{transition dynamics} which construct a mapping of state-action pair at time $t$ onto a set of states at time $t+1$.
However, with \cite{arulkumaran2017deep}, it is common sense that $\mathcal{T}$ is not available in RL settings. Therefore, the so-called \textit{Q-function}: $\mathcal{Q}_{\pi}({\mathbf{s}}, {\mathbf{a}})$ is introduced as the alternative to $V_{\pi}(\mathbf{s})$:
\begin{equation}
\begin{split}
\mathcal{Q}_{\pi}({\mathbf{s}}, {\mathbf{a}}) = \mathbb{E}[\mathcal{R}|{\mathbf{s}}, {\mathbf{a}}, {\mathbf{\pi}}],
\end{split}
\end{equation}
where the initial action $\mathbf{a}$ and the following policy $\mathbf{\pi}$ is pre-given. $\mathcal{Q}_{\pi}({\mathbf{s}}, {\mathbf{a}})$ denotes the expected return value of taking an action $\mathbf{a}$ in a state $\mathbf{s}$ following the policy ${\pi}$.
Thus, given $\mathcal{Q}_{\mathbf{\pi}}({\mathbf{s}}, {\mathbf{a}})$, the optimal policy ${\mathbf{\mathbf{\pi}}}^{\ast}$ can be obtained by the greedy search among all actions and the corresponding $V_{{\mathbf{\pi}}^{\ast}}(\mathbf{s})$ can be defined as:
\begin{equation}
\begin{split}
V_{{\mathbf{\pi}}^{\ast}}(\mathbf{s}) = \max_{\mathbf{a}}\mathcal{Q}_{\mathbf{\pi}}({\mathbf{s}}, {\mathbf{a}}).
\end{split}
\end{equation}
Therefore, it is vital to learn $\mathcal{Q}_{\mathbf{\pi}}$ in such a scheme. Let $\mathcal{H}$ denote $\mathcal{Q}_{\mathbf{\pi}}({\mathbf{s}}_{t+1}, \mathbf{\pi}({\mathbf{s}}_{t+1}))$, and the recursive form of $\mathcal{Q}_{\mathbf{\pi}}$ can be obtained by utilizing the notion of Markov property and Bellman equation \cite{fei2021exponential}:
\begin{equation}
\begin{split}
\mathcal{Q}_{\mathbf{\pi}}({\mathbf{s}}_t, {\mathbf{a}}_t) = \mathbb{E}_{{\mathbf{s}}_{t+1}}[{\rm r}_{t+1} + \gamma \mathcal{H}],
\end{split}
\end{equation}
where ${\rm r}_{t+1}$ denotes the immediate rewards at time ${t+1}$, ${\gamma} \in [0, 1]$ denotes the discount factor for moderating the weight of short-term rewards and long-term rewards. When ${\gamma}$ is close to 0, $Agent$ cares more about short-term rewards; On the contrary,  when ${\gamma}$ is close to 1, $Agent$ cares more about long-term rewards.\\
\textbf{Policy-based strategies.}
Different from value-based strategies, policy-based strategies \cite{kober2013reinforcement}, \cite{peters2010relative}, \cite{deisenroth2013survey}, \cite{nachum2017bridging}, \cite{curi2020efficient} do not depend on a value function, but greedily search for the optimal policy ${\mathbf{\pi}}^{\ast}$ in the policy space. Most policy search strategies perform optimization locally around all policies which are parameterized by a set of policy parameters ${\eta}_i$ respectively. The update of the policy parameters adopts a gradient-based way which follows the gradient of the expected return $E$ with a pre-defined step-size $\alpha$:
\begin{equation}
\begin{split}
{\eta}_{i+1} = {\eta}_{i} + \alpha \triangledown_{\eta} E.
\end{split}
\end{equation}
There exists various approaches to estimate the gradient $\triangledown_{\eta} E$. For instance, in finite difference gradients, the estimate of the gradient can be obtained by evaluating $G$ perturbed policy parameters. Given ${\Delta}E_g \approx E({\eta}_i + {\Delta}{\eta}_g) - E_{ref}$, where $g = [1,2...,G]$ denotes the perturbations, ${\Delta}E_g$ denotes the estimate of influence of $g$ on the expected return $E$ and $E_{ref}$ denotes a reference return (\textit{e.g.}, the return of the unperturbed parameters), let $L = \bigtriangleup {\Upsilon}^{\top} \bigtriangleup {\Upsilon}$, $\triangledown_{\eta} E$ can thus be estimated by
\begin{equation}
\nabla_{\eta}E \approx {L}^{-1} {\bigtriangleup {\Upsilon}^{\top}} {\bigtriangleup \widetilde{E}},
\end{equation}
where ${\bigtriangleup {\Upsilon}}$ denotes a matrix which contains all the samples of the perturbations ${\Delta}{\eta}_g$ and $\bigtriangleup \widetilde{E}$ denotes the matrix contains the corresponding ${\Delta}E_g$. \\
\textbf{Actor-critic strategies.} 
\begin{figure*}[ht]
    \centering
    \includegraphics[scale=0.6]{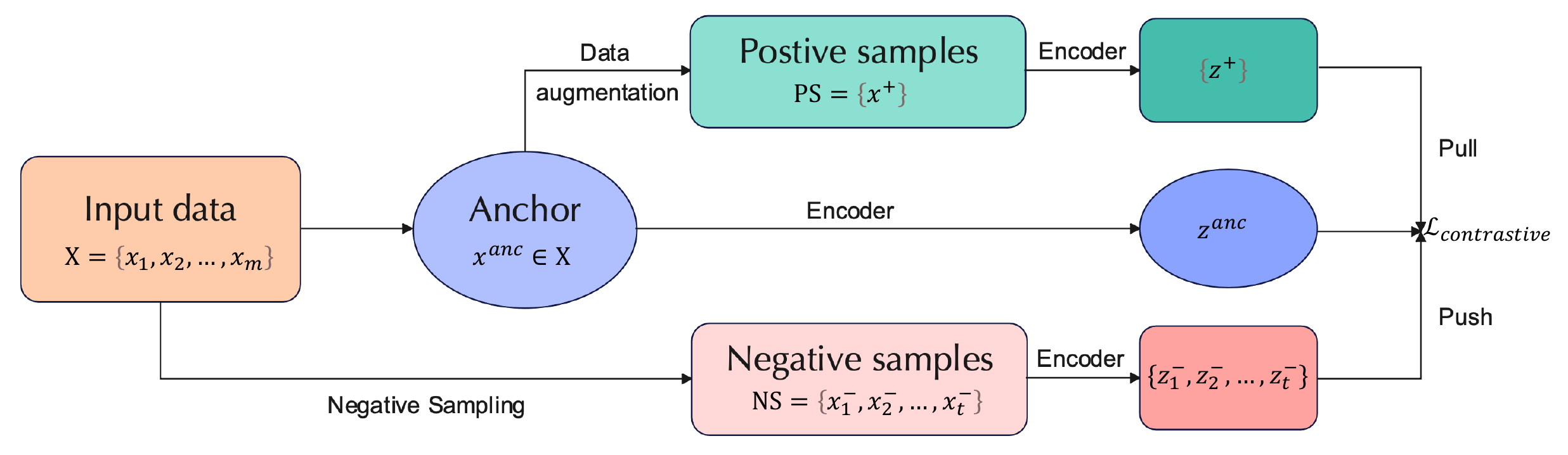}
    \caption{A simple learning framework of contrastive learning. Contrastive learning aims to learn an effective encoder, which can embed the data samples into a latent embedding space, and pull closer the embedded anchor $z^{anc}$ and the postive embeddings $\{z^{+}\}$, push away $z^{anc}$ and the negative embeddings $\{z^{-}_{1},z^{-}_{2},...,z^{-}_{t}\}$. $\mathcal{L}_{contrastive}$ is introduced to update the model.}
    \label{fig:CL}
\end{figure*}
Actor-critic strategies \cite{grondman2012survey}, \cite{haarnoja2018soft}, \cite{fan2019hybrid}, \cite{christianos2020shared}, \cite{lee2020stochastic}, \cite{zanette2021provable}, \cite{wu2021uncertainty} aim to combine the benefits of both policy search strategies and learned value functions. Here ``actor" denotes the policy $\pi$, ``critic" denotes the value function. The ``actor" learns by the feedback from the ``critic", which means actor-critic strategies can obtain effective policies by continuous learning so as to achieve high returns. Different from common policy gradient strategies which utilizes the average of several Monte Carlo returns as the baseline, actor-critic strategies can learn from full returns and TD errors. Once policy gradient strategies or value function strategies make progress, actor-critic strategies may also be improved. The detailed illstration of actor-critic strategies is presented in Figure \ref{fig:RL-AC}.
\par For these strategies, learning on small data may perform its role and demonstrate great potential for reinforcement learning. For instance, learning on small data could be adopted to value-based strategies while evaluating all policies by influencing the expected return to obtain the optimal policy ${\mathbf{\pi}}^{\ast}$ \cite{zang2020metalight}. Moreover, during the process of direct policy search for the optimal policy in policy-based strategies, learning on small data could effectively act as the auxiliary role such as perturbing the direction of policy gradient, thus influencing the final decision of policy search. Besides, in actor-critic strategies, small data learning methods can assist in the adjustment of scores from the ``critic" \cite{mitchell2021offline}. In conclusion, learning on small data could act as an important assisting role in various reinforcement learning scenarios to enhance the efficiency and robustness of models from the perspective of efficient data representation. It still awaits in-depth exploration for the integration of learning on small data and reinforcement learning.
\subsection{Contrastive Learning on Small data}
Self-supervised learning \cite{misra2020self}, \cite{gui2023survey} has obtained widespread interests thanks to its capacity to avoid the cost of annotating large-scale datasets. It mainly utilizes pretext tasks to mining the supervision information from unsupervised data. With the constructed supervision information, we may conduct the model learning and obtain efficient representations for downstream tasks. Contrastive learning \cite{chuang2020debiased}, \cite{khosla2020supervised} has become one of the dominant topics in self-supervised learning, which aims to enhance unsupervised representation by generating different contrastive views. A simple learning framework of contrastive learning is presented in Figure \ref{fig:CL}. Wu \emph{et al.} \cite{wu2021selfsupervised} put that contrastive learning usually generates multiple views for each anchor instance via data augmentation. The augmented views generated from the same anchor are positive pairs while from different anchor are negative pairs. With this, maximizing the agreement of positive pairs while minimizing the agreement of negative pairs is the ultimate goal of contrastive learning from the perspective of Mutual Information (MI), which is defined as: 
\begin{equation} \label{MI}
{\rm MI}(v_i, v_j) = {\mathbb{E}}_{p(v_i, v_j)} \Bigg[{\rm log} \frac{p(v_i, v_j)}{p(v_i)p(v_j)}\Bigg],
\end{equation}
where $v_i$ and $v_j$ denote different views,  $p(v_i, v_j)$ denotes the joint distribution of $v_i$ and $v_j$. Besides, $p(v_i)$ and $p(v_j)$ denote the marginal distributions of $v_i$ and $v_j$ respectively. Eq.~(\ref{MI}) could be estimated by InfoNCE \cite{oord2018representation}, JSD \cite{nowozin2016f}, \textit{etc}.
\par Although contrastive learning makes great efforts to learn an efficent representation, there may often exist few efficient data in various contrastive learning scenarios, thus setting restrictions on achieving the goal. Learning on small data could take full account of the cases from the perspective of efficent data representation. Futhermore, with \cite{robinson2021contrastive}, contrastive learning benefits from $hard$ negative samples. How to seek truly $hard$ negative samples for positive pairs is one of the most significant issues in contrastive learning. Learning on small data could contribute to finding the truly \emph{hard} negative samples so as to improve the performances of contrastive learning models \cite{xia2022progcl}. Besides, appropriate data augmentation strategies in contrastive learning could expand the number of trainable samples, and avoid capturing shortcut features, it could thus promote obtaining efficient positive samples and greatly improving the representation capability and generalizability of models \cite{shorten2019survey}. Learning on small data may thus be considered as a novel perspective for the choice of $hard$ negative samples and the guidance of data augmentation strategies. It may also help evaluate and design reasonable contrastive loss to constrain and reap a robust contrastive learning model. Through the above perspectives, learning on small data could be adopted to different sections of the complete pipeline and assist the contrastive learning models to harvest promising performance and great generalizability. It deserves further exploration to promote contrastive learning with learning on small data.
\subsection{Graph Representation Learning on Small data}
Graph is a common data structure for describing complicated systems such as social networks, recommendation system. Due to the great expressive power of graph, Graph Representation Learning (GRL) \cite{hamilton2020graph} has gradually attracted tremendous attention. GRL aims to build models which could learn from non-Euclidean graph data in an efficient manner. In the setting, various graph neural networks (GNNs) \cite{wu2020comprehensive} emerge as the times require, they show great potential in structured-data mining tasks, \textit{e.g.}, node classification, link prediction, or graph classification. 
\par
\begin{figure*}[ht]
    \centering
    \includegraphics[scale=0.6]{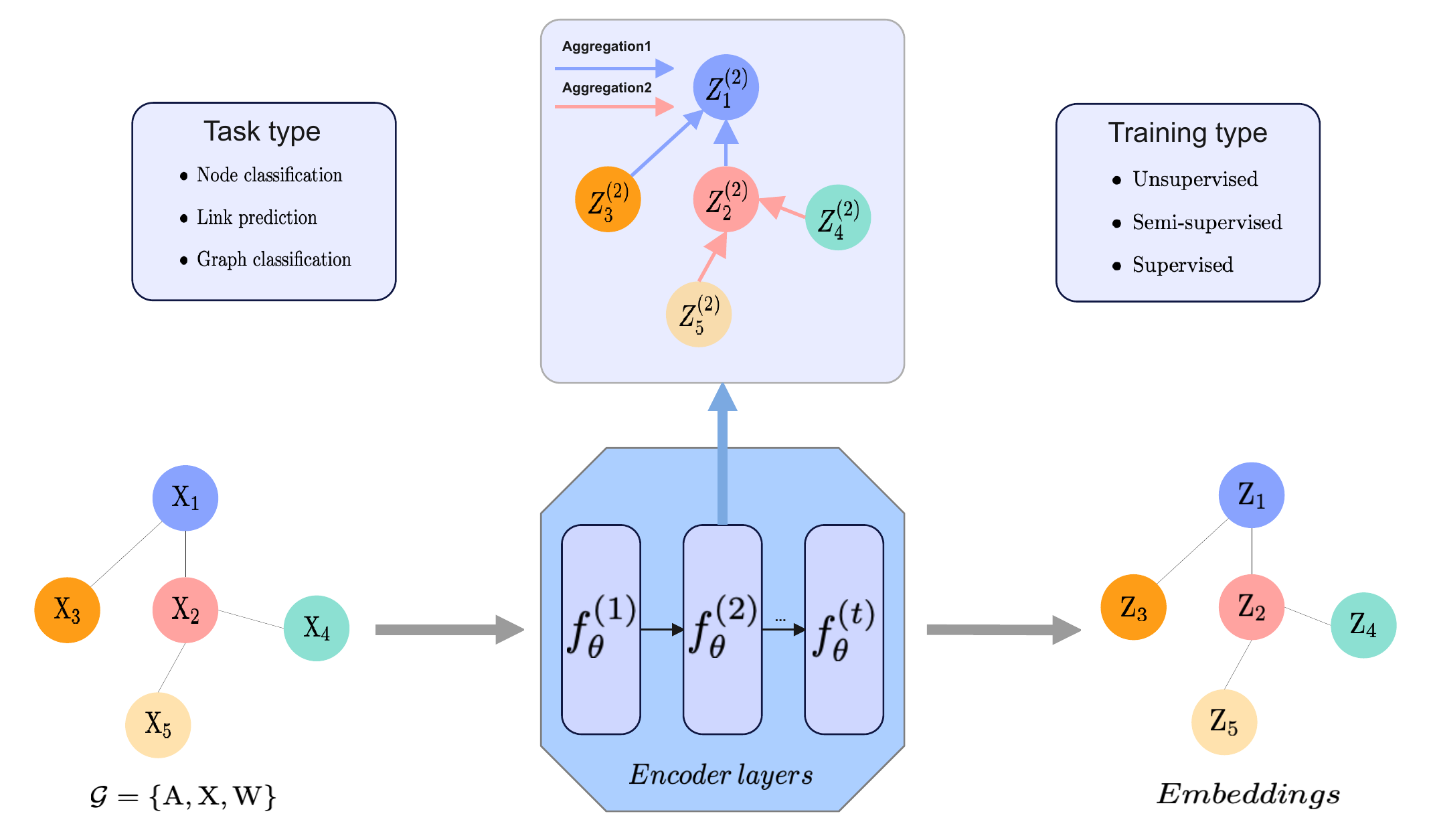}
    \caption{A general design pipeline for GNNs. For an input graph $\mathcal{G} = \{{{\rm A}, {\rm X}, {\rm W}}\}$, ${\rm A}$ denotes the adjacency matrix of $\mathcal{G}$, ${\rm X}$ denotes the feature matrix of all nodes in $\mathcal{G}$, ${\rm W}$ denotes the weight matrix composed of learnable parameters which can be dynamically adjusted during training. Mutiple encoder layers are introduced to encode $\mathcal{G}$ into the corresponding embeddings. In each encoder layer, message passing and neighborhood aggregation are conducted.} 
    \label{fig:GNNs}
\end{figure*}

 However, graph mining tasks often suffer from label sparsity issue when faced with prevailing supervised scenarios where limited efficient data or labels exist \cite{zhang2022few}. How to tackle the performance degradation and obtain a data-efficient model in this setting? Besides, label noise propagation may degrade the representation performance of GNNs and pose a threat to the robustness and generalizability of GNNs \cite{qian2023robust}. Therefore, it's of great importance to provide a label-efficient and noise-resistant GNN model in the setting. Moreover, training GNNs on large-scale graphs suffers from the $\textit{neighbor explosion}$ problem \cite{shi2023lmc}, it awaits effective subgraph-wise sampling methods to tackle it. For these issues, learning on small data may provide necessary and powerful support due to its efficient representational capability. We present a general design pipeline for GNNs in Figure \ref{fig:GNNs}. With it, let us first review the classic taxonomy of GNNs: recurrent graph neural networks (RecGNNs), convolutional graph neural networks (ConvGNNs), graph autoencoders (GAEs) \cite{wu2020comprehensive}.\\
\textbf{RecGNNs.}
RecGNNs aim to learn node representation with recurrent neural mechanisms. In detail, RecGNNs leverage the same recurrent parameters (\textit{i.e.}, the same graph recurrent layers) over nodes to obtain high-level representations \cite{wu2020comprehensive}. In this setting, let $l_{vq} = {\textbf{x}}_{\textbf{e}}(v,q)$ denote the features of edge linking node $v$ and node $q$, the hidden state of node ${v}$ at time ${t}$ is recurrently updated by
\begin{equation}
\begin{split}
{{\textbf{h}}_v^{(t)}} = \sum_{q \in N(v)} f({\textbf{x}}_v, {\textbf{x}}_{q}, l_{vq}, { {\textbf{h}}_v^{(t-1)}}),
\end{split}
\end{equation}
where ${N(v)}$ denotes the neighborhood set of node ${v}$, ${{\textbf{x}}_v}$ and ${{\textbf{x}}_q}$ denote the node features of ${v}$ and ${q}$ respectively,
$f(\cdot)$ denotes the recurrent function which maps nodes into the latent space so as to shrink the distance between the nodes. Following this way, GraphESN \cite{wu2020comprehensive}, GGNN \cite{li2016gated} and SSE \cite{dai2018learning} emerge. \\
\textbf{ConvGNNs.}
Different from RecGNNs, ConvGNNs employs different graph convolutional layers to extract high-level node representations.
Specifically, ConvGNNs approaches could be categoried into spectral-based and spatial-based \cite{liu2022deep}. The former adopts graph convolution from the perspective of graph signal processing\cite{shuman2013emerging} and the latter defines graph convolution from the perspective of information propagation \cite{schoenholz2016deep}.
\par For spectral-based ConvGNNs, let $l$ denote the layer index, ${\rm g}_{\theta} = {\Omega}_{i,j}^{(l)}$ denote the convolution filter which is a diagonal matrix composed of learnable parameters, $c_{l-1}$ denote the number of input channels, ${\rm H}^{(l-1)}$ denote the input graph signal, the hidden state of spectral-based ConvGNNs is formalized as:
\begin{equation}
\begin{split}
{\textbf{H}}^{(l)} = \sigma \Big(\sum_{i=1}^{c_{l-1}} {\textbf{U}} {\Omega^{(l)}} {\textbf{U}}^{\top} {\textbf{H}}^{(l-1)}\Big),
\end{split}
\end{equation}
where ${\textbf{U}} = [\textbf{u}_0, \textbf{u}_1,...,\textbf{u}_{n-1}] \in R^{n \times n}$ denotes the matrix composed of eigenvectors which could be obtained by decomposing the normalized Laplacian matrix. In this setting, ChebNet \cite{defferrard2017convolutional} performs approximations of filter ${\rm g}_{\theta}$ with Chebyshev polynomials, GCN \cite{kipf2017semisupervised} further simplifies the filtering with only the first-order neighborhoods. CayleyNet \cite{levie2018cayleynets} adopts Cayley polynomials to define graph convolutions. In following work, various variants such as AGCN \cite{li2018adaptive}, DGCN \cite{zhuang2018dual} emerge.
\par For spatial-based ConvGNNs, they define graph convolutions based on the spatial relations of nodes. NN4G \cite{4773279} first emerges in the setting, it performs graph convolutions by neighborhood infomation aggregation and leverages residual connections and skip connections to perserve information over each layer. Let $\Psi = \sum_{i=1}^{l-1}{\textbf{A}}{\Omega}^{(l)}{\textbf{H}}^{(l-1)}$, the hidden state is updated as: 
\begin{equation}
\begin{split}
{\textbf{H}}^{(l)} = f\Big({\textbf{W}}^{(l)} {\textbf{X}} + \Psi \Big),
\end{split}
\end{equation}
where ${\rm A}$ denotes the adjacency matrix, ${{\Omega}^{(l)}}$ denotes the matrix which consists of filter parameters, ${\rm W}^{(l)}$ denotes the weight matrix composed of learnable parameters. In this setting, DCNN \cite{atwood2016diffusion}, MPNN \cite{balcilar2021breaking}, GIN \cite{xu2019powerful}, GAT \cite{velivckovic2017graph},  and various spatial-based ConvGNNs variants emerge as the times require.  \\
\textbf{GAEs.}
GAEs hold encoder-decoder architecture on graphs, which encode nodes or graphs into a latent space and decode the corresponding information from the latent representation \cite{kipf2016variational}. Following this way, a series of GAE models burst forth such as DNGRs \cite{cao2016deep}, SDNE \cite{wang2016structural}, VGAE \cite{kipf2016variational}, ARVGA \cite{pan2018adversarially},  GraphSage \cite{hamilton2017inductive}. Take GraphSage as an example, let ${\rm dec}(\cdot)$ be a decoder which consists of a multi-layer perceptron, ${\rm z}_v$ and ${\rm z}_q$ denote two node embeddings, ${\rm Q}$ denote the number of negative samples, $D_1 = {\rm dec}({\rm z}_v, {\rm z}_q)$, $D_2 = -{\rm dec}({\rm z}_v, {\rm z}_{v_i})$. GraphSage \cite{hamilton2017inductive} puts that the negative sampling with the loss ${\mathcal{L}_{\rm Sage}}$ can preserve significant relational information between nodes:
\begin{equation}
\begin{split}
{\mathcal{L}_{\rm Sage}}({\rm z}_v)\! =\! -{\log(D_1)}  \!-\!  {\rm Q} {\rm E}_{v_i \sim D_i(v)} {{\rm \log}(D_2)},
\end{split}
\end{equation}
where node ${\rm z}_{v_i}$ denotes a distant node from ${\rm z}_v$ sampled from the negative sampling distribution $D_i(v)$, ${\mathcal{L}_{\rm Sage}}$ considers that close nodes tend to share similar representations and distant nodes to have dissimilar representations. Besides, massive work emerges from the perspective of network embedding and graph generation in the setting.\\
\par Will learning on small data truly benefits graph representation learning? The answer is: exactly. For example, learning on small data could provide feasible proposal to extract efficient small data representation in graph contrastive learning, which could promote the choice of $hard$ negative samples \cite{xia2022progcl}. Besides, in some graph mining tasks, learning on small data promotes the extraction of efficient graph data representation. Moreover, in pursuit of a robust graph representation model, learning on small data could help eliminate the untrustworthy noisy data, invoking a highly generalized model \cite{hamilton2017representation}. In the context of graph statistical characteristics, node centrality metrics \cite{rodrigues2019network} includes degree centrality, closeness centrality, eigenvector centrality, betweenness centrality, \emph{etc}. It is promising to adopt small data learning methods to explore more centrality metrics for nodes. Furthermore, in some GNNs, learning on small data may promote obtaining better neighborhood aggregation or message passing schemes, \textit{e.g.}, defining how important the message from a neighbor is to a node, and promote the message update between nodes, thus enhancing the performance of models. To conclude, it is of great potential to integrate learning on small data and graph representation learning to explore more possible collaborable scenarios.
\begin{figure}
    \centering
    \includegraphics[scale=0.32]{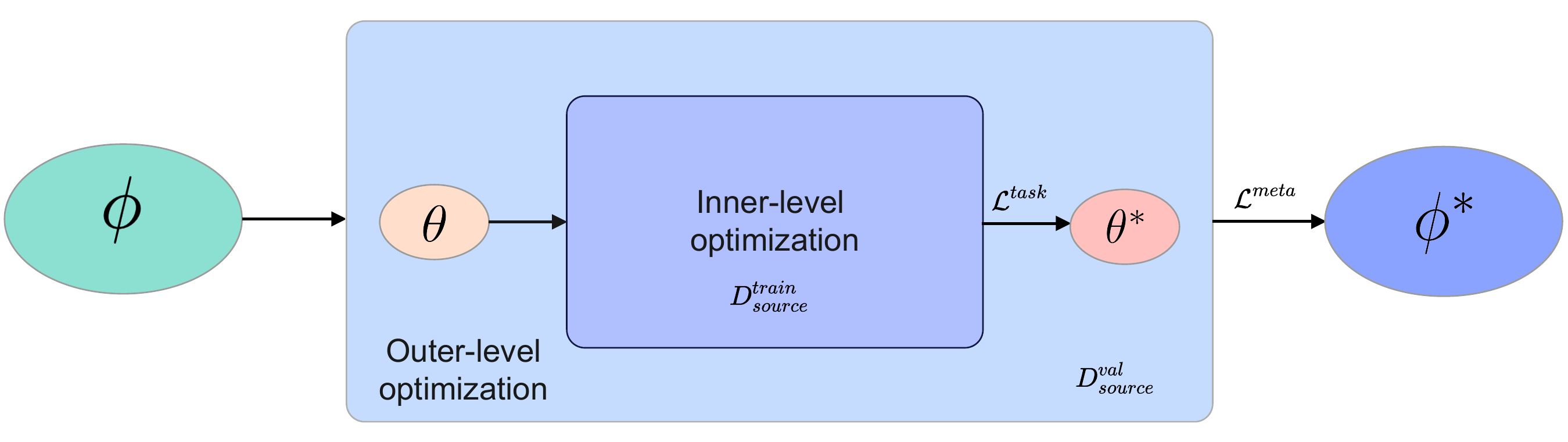}
    \caption{The bilevel optimization fashion of meta-learning. During the meta-training phrase, the inner-level optimization is to train the model on different tasks which are sampled from $D^{train}_{source}$ to obtain the optimal model with parameters ${\theta}^{\ast}$, $\mathcal{L}^{task}$ denotes the optimization objective for inner-level optimization; the outer-level optimization aims to obtain a general meta-knowledge ${\phi}^{\ast}$ which be quickly adapted to unseen tasks. $\mathcal{L}^{meta}$ denotes the optimization objective to obtain ${\phi}^{\ast}$.}
    \label{fig:meta-learning}
\end{figure}
\section{Parameter Optimization with Small Data : A Meta Fashion}
We have summarized multiple learning communities that may benefit from small data representation in Section~6. Following these settings, we below present a potential meta parameter update fashion over small data training.

Theoretically, learning parameters could be optimized by 
utilizing the well-generalized meta-knowledge across   various learning tasks, that is, teach the \emph{learning} models \emph{to learn} for unseen tasks \cite{hospedales2020metalearning}. This requires that the current parameter update policy could be effectively generalized into unseen tasks. Therefore, \emph{how to extract well-generalized meta-knowledge} \cite{tian2021consistent}  has become an critical issue in meta parameter update. To explore potential meta parameter update fashions, we integrate meta-learning from multiple fashions including bilevel, task-distribution, and feed-forward. More details are presented in Supplementary Material.
\subsection{Bilevel Fashion}
Bilevel optimization \cite{ji2021bilevel}, \cite{khanduri2021near} is a hierarchical optimization problem, which means one optimization objective contains another inner optimization objective as a constraint. The bilevel optimization scheme of meta-learning is presented in Figure \ref{fig:meta-learning}.
\subsection{Task-Distribution Fashion}
From the perspective of task-distribution \cite{vuorio2019multimodal,collins2020task}, meta parameter update considers tasks as samples of the model. Besides, this update paradigm aims to learn a common learning algorithm which can generalize across tasks. In detail, the goal is to learn a general meta-knowledge $\phi$ which can minimize the expected loss of meta-tasks.
\subsection{Feed-Forward Fashion}
With \cite{hospedales2020metalearning}, there exists different meta parameter update schemes do not adopt an explicit bilevel optimization, which synthesize models in feed-forward manners \cite{bertinetto2016learning}, \cite{denevi2018learning}. 
\section{Challenges}
In this section, we introduce multiple learning scenarios and data applications that may bring challenges for learning on small data. In practice, these settings could benefit from small data representation, but also may have potential troubles, \textit{e.g.}, efficiency degeneration, loose approximation error, and expensive computational overhead.

\subsection{Scenarios}
\subsubsection{Weak Supervision}
Weak supervision \cite{li2019towards}, \cite{zhou2018brief} often produces incomplete, inexact, and inaccurate sources of supervision to generate training features. This saves costs for the typical manual labeling in unlabeled data collection. In this scenario, how to learn an effective training model with weak supervision attracts our learning communities. 

With incomplete supervision, semi-supervised learning has shown great potential for learning from a limited amount of labeled data and a large amount of unlabeled data. In the existing efforts, pseudo-labeling and consistency regularization are the two  important  branches \cite{chen2022semi}. Specifically, the former aims to train the models with pseudo-labels whose prediction confidence goes beyond a hard threshold \cite{lee2013pseudo}, \cite{pham2021meta}, and  the latter, on the other hand, attempts to maintain the output consistency under perturbations on data \cite{berthelot2019mixmatch}, \cite{sohn2020fixmatch} or model \cite{tarvainen2017mean}, \cite{laine2016temporal}. Besides, for inexact supervision setting, multi-instance learning \cite{carbonneau2018multiple} takes the coarse-grained labels into account to promote the model training. For inaccurate supervision setting, label noise learning \cite{han2020survey} makes efforts to combat noisy labels and pursue a robust model. 
 
Despite efforts in the weakly-supervised scenario, it remains challenging to obtain effective data representations. There may be issues with missing dimensions and labels, which can result in degraded model performance. Therefore, there is an urgent need to develop data-efficient schemes that can obtain high-quality representations while preserving independent structural information for features. Learning from small data has been shown to be effective for this purpose. However, it's worth noting that weak supervision increases the risk of producing a looser approximation error for small data representations compared to general learning scenarios.

\subsubsection{Multi-label}
Multi-label learning \cite{zhou2017multi}, \cite{liu2021emerging} studies the setting that each instance is associated with multiple categories, \textit{i.e.}, map from the input $d$-dimensional instance space $\mathcal{X} = \mathcal{R}^{d}$ to $2^C$ dimensional label space $\mathcal{Y} = \{y_1, y_2, ..., y_{C}\}$ with $C$ categories. In this scenario, how to model label correlations and handle incomplete supervision remains challenging.

For multi-label learning with missing labels, it is assumed that only a subset of labels is accessible. Existing approaches can be categorized into low-rank-based methods such as \cite{xu2013speedup}, \cite{yu2014large}, and \cite{han2018multi}, as well as graph-based methods like \cite{yu2017unified} and \cite{huang2018learning}. Similarly, for semi-supervised multi-label classification, it assumes limited labeled data available and large amounts of unlabeled data given in the setting, inheriting same method categories   from typical multi-label learning  \cite{sun2019robust}  \cite{sun2019robust}.  Additionally, for partial multi-label learning, the annotators are allowed to preserve all potential candidate labels \cite{xu2020partial}.

However, even though existing approaches attempt to sufficiently exploit the label correlation and data distribution, they may invoke the potential damage to label correlation in small data setting. Learning on small data makes efforts to eliminate the damage for multi-label learning, and enables the modeling of label correlations with the power of efficient small data representation.
\subsubsection{Imbalanced Distribution}
Real-world data often presents long-tailed distributions with skewed class proportions, which poses a persisting dilemma: the imbalanced distribution may incur `label bias' issue, in which the decision boundary might be significantly altered by the majority classes \cite{yang2020rethinking}. It still requires more expressive generalization on the minority classes for the learning model.

The existing efforts for imbalanced distribution learning could be categorized into re-sampling and re-weighting \cite{cao2019learning} strategies. For re-sampling based approaches, over-sampling the minority classes \cite{byrd2019effect} and under-sampling the majority classes \cite{buda2018systematic} hold important positions. Besides, re-weighting based approaches \cite{huang2019deep} attempt to assign adaptive weights for different classes or even samples. Another line of work attempts to propose class-balanced losses \cite{cao2019learning}, \cite{cui2019class} for imbalanced classification problems.

However, effective small data representation usually requires more fairly i.i.d. sampling from each class. This may lose the original class relationship.  Only if the classification hyperplanes are structured, the approximation loss could be tight. Furthermore, learning on small data may provide data-efficient proposals for re-sampling or re-weighting strategies.
 
 
\subsection{Data}
 
\subsubsection{Large Image Data} 
Computer vision \cite{bi2022survey} attempts to mimic human vision systems and learn  from images and videos, enabling  a wide variety of applications such as image recognition, object detection, semantic segmentation, video processing thrive. With the powerful representational ability of deep models, the state-of-the-art in computer vision has been pushed forward drastically. A common setting is that scarced labeled data and large-scale unlabeled data are available.  However, a common challenge is the scarcity of labeled data and the abundance of unlabeled data. The requirement for labeled data has also been pushed to a new level due to the growing complexity of models, which may be unattainable for some applications. 

For example, in most of medical image analysis tasks, large-scale unlabeled image data are available, while labeled image data are extremely scarce due to the expensive costs of the professional facilities and considerable expertise requirement for annotation. It is essential to develop effective schemes to tackle the dilemma. Existing explorations include designing active learning algorithms for effective queries \cite{hoi2006batch, wang2020deep}, transferring knowledge from related domains \cite{ZhangMML18}, and exploiting the large-scale unlabeled data. Learning on small data could invoke the efficient extraction of data representation, thus promoting these existing efforts. With \cite{yang2022diffusion}, diffusion models have shown excellent potential in large-scale image synthesis, video generation, \emph{etc}. To speed up the sampling process, meanwhile, improving quality of the resulting data in diffusion models, learning on small data could lend a hand such as accelerating the proposals of more efficient sampling strategies. Nevertheless, it remains challenging for deep models to leverage the large-scale image data.

\subsubsection{Large Language Oracle}
Natural language processing (NLP) system \cite{hedderich2020survey} tackles with the semantic parsing, translation, speech recognition, summarization, and more, in which the large language models have yielded promising results for various NLP applications. However, the resource consumption of these models also grows with their size \cite{treviso2022efficient}. Therefore, it is urgent to develop efficient large language models that require fewer resources while still obtaining competitive results. 

Numerous existing methods make efforts for this from the perspectives of data efficiency, model design efficiency \cite{treviso2022efficient}. To improve data efficiency, the common practice is to improve data quality such as removing duplicates \cite{lee-etal-2022-deduplicating}, adversarial filtering \cite{mishra2020we} during pre-training and fine-tuning. Besides, active learning could also conduct high-quality language instance selection \cite{yuan-etal-2020-cold} with the two common criterions: uncertainty and representativeness. To improve model design efficiency and accelerate training, some work attempt to improve the attention mechanism in Transformers \cite{tay2022efficient}; Another line of efforts pursue sparse modeling for language data, which leverages sparsity for efficient training \cite{du2022glam}; In addition, some parametric models could interact with retrieval mechanisms for text generation, yielding great generalization on language data across domains \cite{li2022survey}.
With both perspectives, learning on small data can provide efficient language data representation for data quality enhancement, attention design improvement, \emph{etc}.

\subsubsection{Science Data}
Artificial intelligence (AI) holds tremendous promise in posing an impact on scientific discovery (\textit{i.e.}, AI for Science) such as addressing grand issues in structural biology \cite{rout2019principles}, accelerating drug discovery \cite{atanasov2021natural}, interacting physical insights with AI \cite{li2022learning} \emph{etc}. It remains challenging to learn from complex science data and obtain effective models.

For structural biology, AI is efficiently fueling the development of subfields such as automated genome annotation, protein binding prediction, metabolic functions prediction, \emph{etc}. However, obtaining biology data usually requires a long-time culture (\eg, cell culture), or the involving of expensive apparatuses, which bring challenges to collect large-scale labeled data. Moreover, the biomedical data is often high-dimensional and sparse \cite{2016GWAS}, much of them is even incomplete and biased \cite{vzitnik2015data}. With these challenges, training effective models for biological tasks can be rather difficult. Similarly, for drug discovery, various applications develop vigorously with the power of DNNs / GNNs in drug–protein interaction prediction, drug efcacy discovery, \emph{etc}. It awaits highly-efficient drug data representation schemes to accelerate the improvement of existing methods. Besides, motivated by the physical insight, existing physics-informed neural networks \cite{karniadakis2021physics} attempt to embed physical knowledge (\textit{e.g.}, physical laws) into DNNs to design models which could automatically hold physical invariant properties, and better model the science data. Learning on small data could promote effectively representing the complex science data and accelerating the model training, it remains in-depth exploration.

 
\section{Conclusion}
In this paper, we firstly present a formal definition for learning on small data, and then provide  theoretical  guarantees for its supervised and unsupervised analysis of model-agnostic generalization  on error and label complexity under a PAC framework.
From a geometric perspective, learning on small data could be characterized by the Euclidean and non-Euclidean geometric representation, where their geometric mean expressions are presented and analyzed with respect to a unified expression of Fréchet mean. To optimize those geometric means,  the Euclidean gradient, Riemannian gradient, and Stein gradient are investigated. Besides these technical contents, some potential learning communities that may benefit from learning on small data are also summarized. With the learning settings, a potential meta parameter update fashion is presented. Finally, 
the related advanced challenging scenarios and data applications are also presented and discussed.


%
\bibliographystyle{IEEEtran}
\bibliography{Reference}

%
\begin{IEEEbiography}
[{\includegraphics[width=1in,height=1.25in,clip,keepaspectratio]{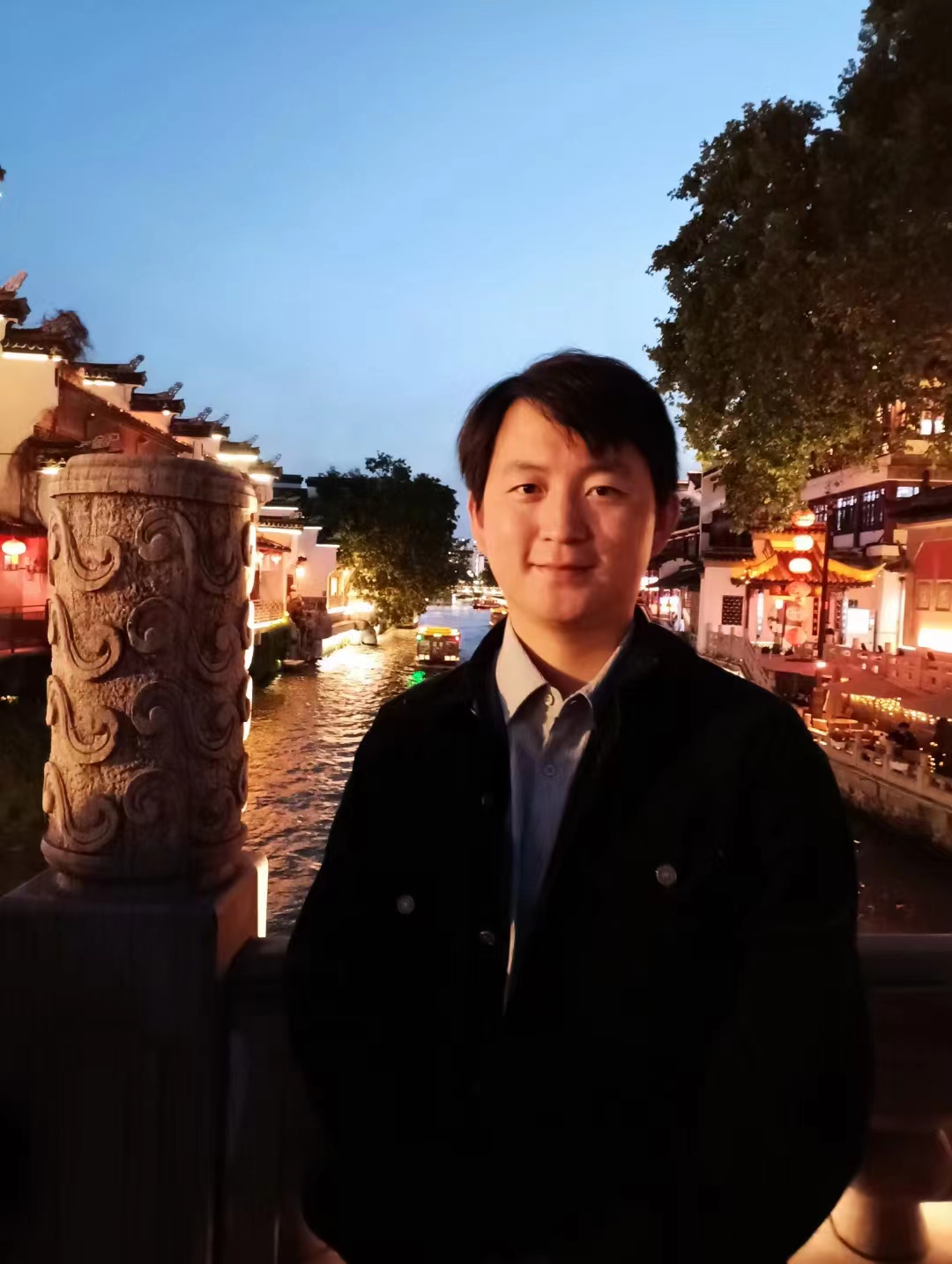}}]{Xiaofeng Cao} received his Ph.D. degree at Australian Artificial Intelligence Institute, University of Technology Sydney, Australia. He is currently an Associate
Professor at the School of Artificial Intelligence, Jilin
University, China and leading a Machine Perceptron
Research Group with more than 20 PhD and Master
students. He has published more than 20 technical
papers in top tier journals and conferences, such as
IEEE T-PAMI, IEEE TNNLS, IEEE T-CYB, ICML, CVPR,
IJCAI. His research interests include the PAC learning theory, agnostic learning algorithm, generalization analysis, and hyperbolic geometry.
\end{IEEEbiography}
\vspace*{-1.5cm}
\begin{IEEEbiography}
[{\includegraphics[width=1in,height=1.25in,clip,keepaspectratio]{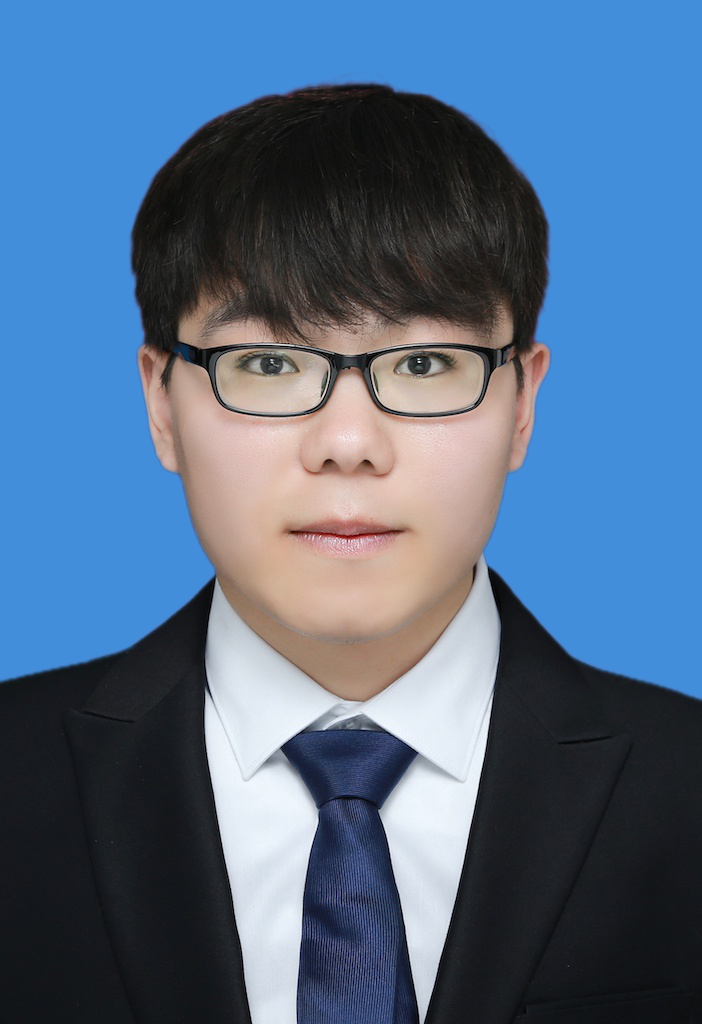}}]{Weixin Bu} is now a master student supervised by Dr.~Cao at the School of Artificial Intelligence, Jilin University. Before that, he spent almost three years from a software engineer to a leader of data team in BorderX Lab INC, where he built and optimized various data systems. He received the BSc degree in network engineering from Nanjing University of Information Science \& Technology, China, in 2018. His main research interests include contrastive learning, graph representation learning and OOD generalization.
\end{IEEEbiography}
\vspace*{-1.5cm}
\begin{IEEEbiography}
[{\includegraphics[width=1in,height=1.25in,clip,keepaspectratio]{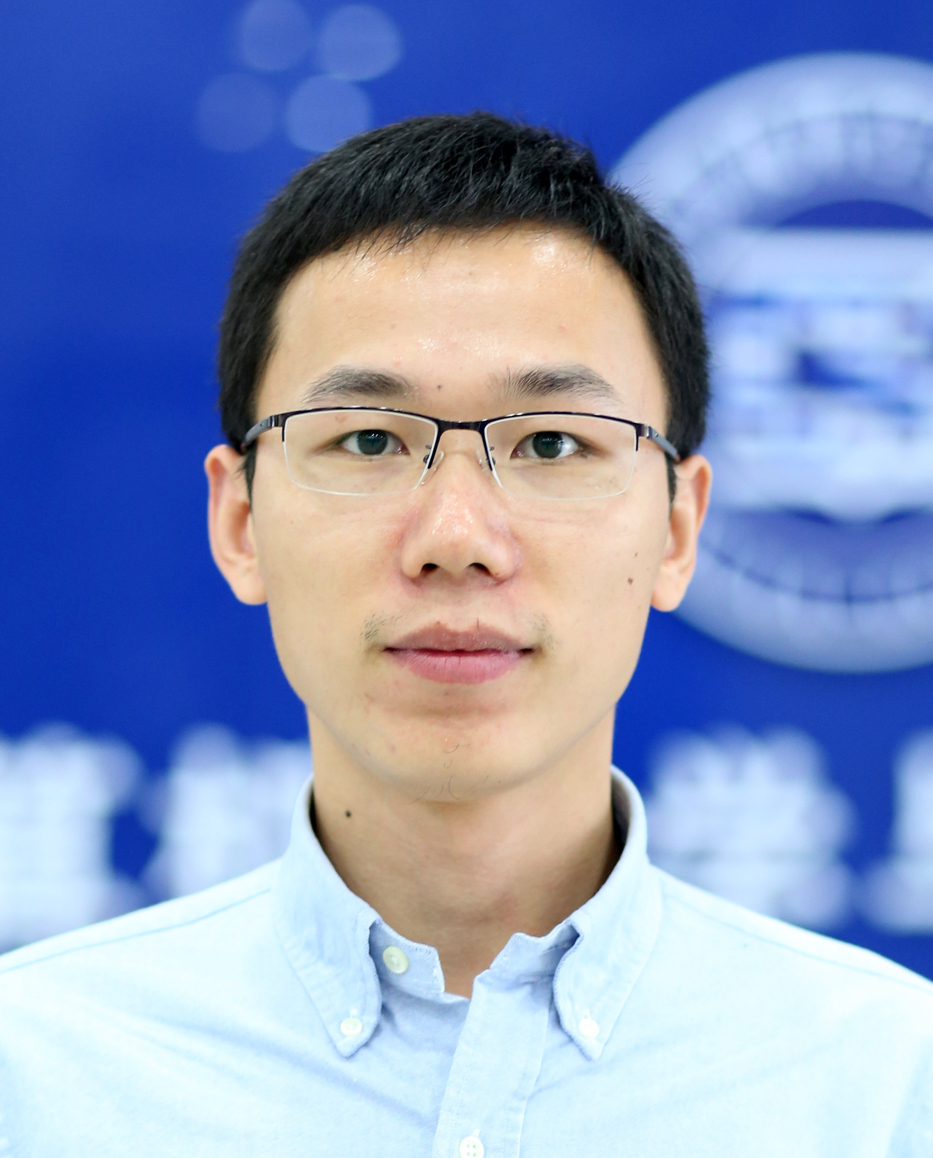}}]{Shengjun Huang}   received the BSc and PhD
degrees in computer science from Nanjing University, China, in 2008 and 2014, respectively.
He is now a professor in the College of Computer
Science and Technology at Nanjing University
of Aeronautics and Astronautics. His main research interests include machine learning and
data mining. He has been selected to the Young
Elite Scientists Sponsorship Program by CAST
in 2016, and won the China Computer Federation Outstanding Doctoral Dissertation Award in
2015, the KDD Best Poster Award in 2012, and the Microsoft
Fellowship Award in 2011. He is a Junior Associate Editor of Frontiers
of Computer Science.
\end{IEEEbiography}
\vspace*{-1.5cm}
\begin{IEEEbiography}
[{\includegraphics[width=1in,height=1.25in,clip,keepaspectratio]{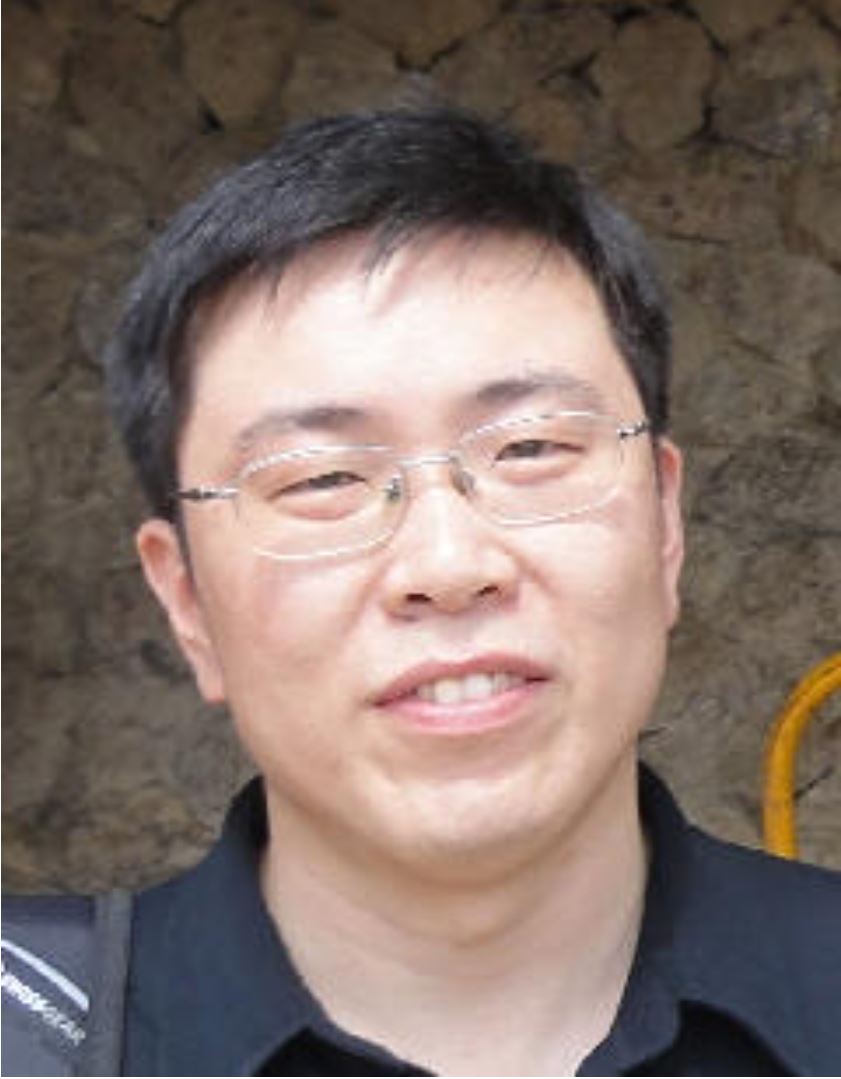}}]{Minling Zhang} (Senior Member, IEEE) received the BSc, MSc, and PhD degrees in computer science from Nanjing University, China, in 2001, 2004 and 2007, respectively. Currently, he is a professor with the School of Computer Science and Engineering, Southeast University, China. His main research interests include machine learning and data mining. 
\par In recent years, he has served as the general co-chairs of ACML’18, program co-chairs of PAKDD’19, CCF-ICAI’19, ACML’17, CCFAI’17,
PRICAI’16, senior PC member or area chair of KDD 2021, AAAI 2017-2020, IJCAI 2017-2021, ICDM 2015-2020, \textit{etc}. He is also on the editorial board of \textit{ACM Transactions on Intelligent Systems
and Technology}, \textit{Neural Networks}, \textit{Science China Information Sciences}, \textit{Frontiers of Computer Science}, \textit{etc}. He is the steering committee member of ACML and PAKDD, secretary-general of the CAAI Machine Learning Society, standing committee member of the CCF Artificial Intelligence \& Pattern Recognition Society. He is a distinguished member of CCF, CAAI, and senior member of ACM.
\end{IEEEbiography}
\begin{IEEEbiography}
[{\includegraphics[width=1in,height=1.25in,clip,keepaspectratio]{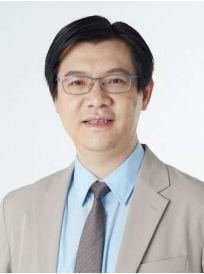}}]{Ivor W. Tsang} (Fellow, IEEE) is Professor of Artificial Intelligence, at University of Technology Sydney. He is also the Research Director of the Australian Artificial Intelligence Institute, and an   IEEE Fellow.  In 2019, his paper titled ``Towards ultrahigh dimensional feature selection for big data" received the International Consortium of Chinese Mathematicians Best Paper Award. In 2020, Prof. Tsang was recognized as the AI 2000 AAAI/IJCAI Most Influential Scholar in Australia for his outstanding contributions to the field of Artificial Intelligence between 2009 and 2019. His works on transfer learning granted him the Best Student Paper Award at International Conference on Computer Vision and Pattern Recognition 2010 and the 2014 IEEE Transactions on Multimedia Prize Paper Award. In addition, he had received the prestigious IEEE Transactions on Neural Networks Outstanding 2004 Paper Award in 2007.  
\par Prof. Tsang serves as a Senior Area Chair for Neural Information Processing Systems and Area Chair for International Conference on Machine Learning, and the Editorial Board for  Journal Machine Learning Research, Machine Learning, 
Journal of Artificial Intelligence Research, and IEEE Transactions on Pattern Analysis and Machine Intelligence. 
\end{IEEEbiography}
\vspace*{-1.5cm}
\begin{IEEEbiography}
[{\includegraphics[width=1in,height=1.25in,clip,keepaspectratio]{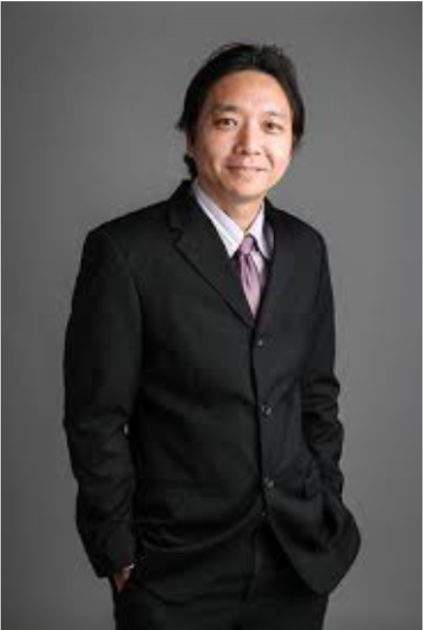}}]{Yew-Soon Ong} (Fellow, IEEE) received the Ph.D. degree in artificial intelligence in complex design from the University of Southampton, Southampton, U.K., in 2003. He is the President’s Chair Professor of Computer
Science with Nanyang Technological University (NTU), Singapore, and holds the position of Chief
Artificial Intelligence Scientist of the Agency for Science, Technology and Research, Singapore. He serves as the Co-Director of the Singtel-NTU Cognitive and Artificial Intelligence Joint Lab, NTU. His research interest is in artificial and computational intelligence.
\par Prof. Ong has received several IEEE outstanding paper awards and was listed as a Thomson Reuters highly cited researcher and among the World’s Most Influential Scientific Minds. He is the Founding Editor-in-Chief of the IEEE TRANSACTIONS ON EMERGING TOPICS IN COMPUTATIONAL
INTELLIGENCE and an Associate Editor of IEEE TRANSACTIONS ON NEURAL NETWORKS AND LEARNING SYSTEMS, IEEE TRANSACTIONS ON CYBERNETICS, IEEE TRANSACTIONS ON ARTIFICIAL INTELLIGENCE and
others.
\end{IEEEbiography}
\vspace*{-1.5cm}
\begin{IEEEbiography}
[{\includegraphics[width=1in,height=1.25in,clip,keepaspectratio]{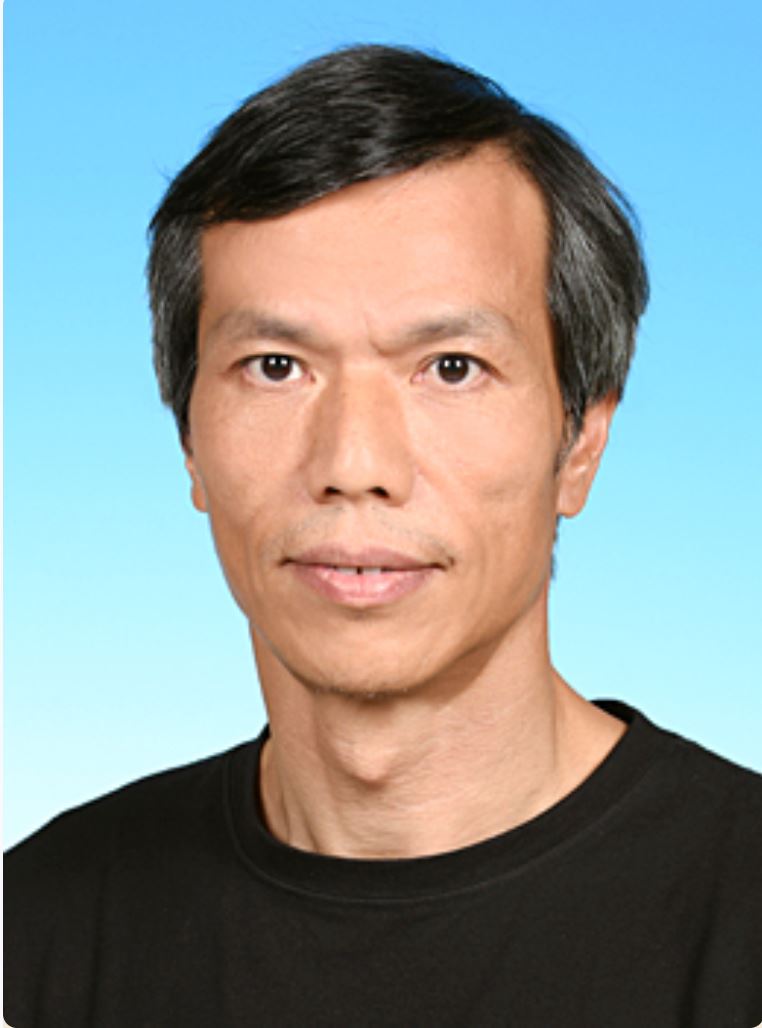}}]{James T. Kwok} (Fellow, IEEE) received the Ph.D. degree in computer science from the Hong Kong University of Science and Technology, in 1996. He is a professor with the Department of Computer Science and Engineering, Hong Kong University of Science and Technology. His research interests include machine learning, deep learning, and artificial intelligence.  
\par Prof. Kwok is serving as an Editorial Board Member
for the Machine Learning. He received the IEEE
Outstanding 2004 Paper Award and the Second Class Award in Natural
Sciences by the Ministry of Education, China, in 2008. He has served/is
serving as the senior area chair/area chair of major machine learning/AI conferences, including the Conference on Neural Information Processing System
(NeurIPS), the International Conference on Machine Learning (ICML), the
International Conference on Learning Representation (ICLR), the International
Joint Conference on Artificial Intelligence (IJCAI), the AAAI Conference
on Artificial Intelligence (AAAI), and the European Conference on Machine
Learning (ECML). He is serving as an Associate Editor for the IEEE
TRANSACTIONS ON NEURAL NETWORKS AND LEARNING SYSTEMS, the
Neural Networks, the Neurocomputing, the Artificial Intelligence Journal, and
the International Journal of Data Science and Analytics.
\end{IEEEbiography}
\clearpage
\section*{Theory Part of Supplementary Material}
\subsection*{Part A: Proof of Theorem~\ref{Unsupervised-Fashion}}
\begin{proof} 
IWAL denotes a set of observations for its weighted sampling: $\mathcal{F}_t=\{(x_1,y_1,Q_1), (x_2,y_2,Q_2),..., (x_t,y_t,Q_t)\}$.  {The key step to prove Theorem~\ref{Supervised-Fashion} is to observe the  martingale difference sequence for any pair $f$ and $g$ in the t-time hypothesis class $\mathcal{H}_t$, that is, $\xi_t=\frac{Q_t}{p_t}\Big(\ell(f(x_t),y_t)-\ell(g(x_t),y_t)\Big)-\Big(R(h)-R(g)\Big)$, where $f, g \in \mathcal{H}_t$.}  {By adopting Lemma~3 of \cite{kakade2008generalization}, with $\tau>3$ and $\delta>0$, we firstly know
\begin{equation}
\begin{split}
&{\rm var}[\xi_t|\mathcal{F}_{t-1}]\\
&\leq \mathbb{E}_{x_t}\Big[ \frac{Q_t^2}{p_t^2}\Big(\ell(f(x_t),y_t)\\ &-\ell(g(x_t),y_t)\Big)\!-\!\Big(R(h)-R(g)\Big)^2 |\mathcal{F}_{t-1}\Big]\\
& \leq \mathbb{E}_{x_t}\Big[ \frac{Q_t^2p_t^2}{p_t^2}  |\mathcal{F}_{t-1}\Big]\\
& = \mathbb{E}_{x_t}\Big[ p_t |\mathcal{F}_{t-1}\Big],\\
\end{split}
\end{equation}
and then there exists
\begin{equation}
\begin{split}
&|\sum_{t=1}^T\xi_t| \\
&\leq   \max_{\mathcal{H}_i, i=1,2,...,k} \Bigg\{ 2 \sqrt{\sum_{t=1}^{\tau}\mathbb{E}_{x_t}[p_t|\mathcal{F}]_{t-1}}, 6\sqrt{{\log}\Big(\frac{8{\rm log}\tau }{\delta}\Big)} \Bigg\} \\ 
&\times \sqrt{{\rm log}\Big(\frac{8{\rm log}\tau}{\delta}\Big)},\\
\end{split}
\label{xi_obsevation}
\end{equation}
where $\mathbb{E}_{x_t}$ denotes the expectation over the operation on $x_t$. With Proposition~2 of \cite{cesa2008improved}, we have
\begin{equation}
\begin{split}
&\sum_{t=1}^{\tau}\mathbb{E}_{x_t}[p_t|\mathcal{F}_{t-1}]\\
&\leq \Big(\sum_{t=1}^\tau p_t \Big)+36 {\rm log}\Big(\frac{(3+\sum_{t=1}^\tau)\tau^2}{\delta}\Big)\\
&+2\sqrt{\Big(\sum_{t=1}^\tau p_t\Big){\rm log}\Big(\frac{(3+\sum_{t=1}^\tau)\tau^2}{\delta}\Big)}\\
& \leq  \Bigg( \sum_{t=1}^\tau p_t + 6 {\rm log}\Big(\frac{(3+\sum_{t=1}^\tau)\tau^2}{\delta}\Big) \Bigg).   \\
\end{split}
\label{F_obsevation}
\end{equation}}
Then, introducing Eq.~(\ref{F_obsevation}) to  Eq.~(\ref{xi_obsevation}), with a probability at least $1-\delta$, we have

\begin{equation}
\begin{split}
& |\sum_{t=1}^T\xi_t|\\
& \leq  \max_{\mathcal{H}_i, i=1,2,...,k} \Bigg\{\frac{2}{\tau}  \Bigg[\sqrt{\sum_{t=1}^{\tau}p_t}+6\sqrt{{\rm log}\Big[\frac{2(3+\tau)\tau^2}{\delta}\Big] } \Bigg]\\ 
&\times \sqrt{{\rm log}\Big[\frac{16\tau^2|\mathcal{H}_i|^2 {\rm log}\tau}{\delta}\Big]}\Bigg\}.
\end{split}
\end{equation}
{For any $\mathcal{B}_i$, the final hypothesis converges into $h_{\tau}$, which usually holds a tighter disagreement to the optimal $h^*$. We thus have $R(h_\tau)-R(h^*) \leq |\sum_{t=1}^\tau\xi_t|$. Therefore, the error disagreement bound of Theorem~\ref{Supervised-Fashion} holds.} We next prove the label complexity bound of MHEAL. Following Theorem~\ref{Supervised-Fashion}, there exists
\begin{equation}
\begin{split}
& \mathbb{E}_{x_t}[p_t|\mathcal{F}_{t-1}]\\
& \leq 4\theta_{\rm IWAL}K_\ell  \times \Bigg(R^*+\sqrt{(\frac{2}{t-1}){\rm log}(2t(t-1)|)  \frac{|\mathcal{H}|^2  }{\delta})}\Bigg), \\
\end{split}
\end{equation}
where  $R^*$ denotes $R(h^*)$. Let  
$\sqrt{(\frac{2}{t-1}){\rm log}(2t(t-1)|)  \frac{|\mathcal{H}|^2  }{\delta})} \propto O\Big({\rm log}(\frac{\tau|\mathcal{H}|}{\delta})\Big)$, with the proof of Lemma~4 of \cite{cortes2020region}, Eq.~(\ref{F_obsevation}) can thus be approximated  as
\begin{equation}
\begin{split}
& \mathbb{E}_{x_t}[p_t|\mathcal{F}_{t-1}] \\
& \leq 4\theta_{\rm IWAL}  K_\ell\Bigg(\tau R^*\!+\!O\Big(R(h^*)\tau {\rm log}(\frac{t|\mathcal{H}|}{\delta})   \Big )  \!+\! O\Big({\rm log}^3(\frac{t|\mathcal{H}|}{\delta})\Big) \Bigg).
\end{split}
\end{equation}
For any cluster $\mathcal{B}_i$, by adopting the proof of Theorem~2 of \cite{cortes2020region}, we know
\begin{equation}
\begin{split}
& \mathbb{E}_{x_t}[p_t|\mathcal{F}_{t-1}]\\  
&\leq   8 K_\ell\Bigg\{\Big[\sum_{j=1}^{N_i} \theta_{\rm MHEAL} R_j^*\tau p_j\Big] 
 +\sum_{j=1}^{N_i} O\Bigg(\sqrt{R_j^*\tau p_j{\rm log}\Big[\frac{\tau|\mathcal{H}_i|N_i}{\delta} \Big]}\Bigg)\!\\
  &+\!O\Bigg(N_i {\rm log}^3\Big(\frac{\tau|\mathcal{H}_i|N_i}{\delta}\Big)\Bigg)\Bigg\}. \\
\end{split}
\end{equation}
Since $\mathcal{Q}=k\times \max_{\mathcal{H}_i}\mathbb{E}_{x_t}[p_t|\mathcal{F}_{t-1}], {\rm s.t.}, i=1,2,...,k$, our analysis of Theorem~\ref{Unsupervised-Fashion} on $\mathcal{Q}$ holds.
\end{proof}

\subsection*{Part B: Proof of Theorem~\ref{thm:kernel mean}}
\begin{proof}
Let $\{\sqrt{\lambda_i}\phi_i(x)\}$ be an orthogonal base of $\mathcal{H}$, the kernel function can be written as: 
\begin{equation}
    K(x,y)=\sum_{i}^{\infty}\lambda_i\phi_i(x)\phi_i(y).
\end{equation}
For any vector $k_x=K(x,\cdot)\in\mathcal{H}$, it can be expressed as a sum of orthogonal bases, \textit{i.e.},  $K(x,\cdot)=\sum_{i}^{\infty}\sqrt{\lambda_i}\phi_i(x)\sqrt{\lambda_i}\phi_i(\cdot)$.
May wish to assume the kernel mean $\mu_{\mathcal{H}}=\sum_{i}^{\infty}\alpha_i\sqrt{\lambda_i}\phi_i(\cdot)$, for any $k_x\in\mathcal{H}$, there exists
\begin{equation}\label{equ:new dh}
    d_{\mathcal{H}}(k_x,\mu_{\mathcal{H}})=\sqrt{\sum_{i}^{\infty}(\sqrt{\lambda_i}\phi_i(x)-\alpha_i)^2}.
\end{equation}
Substituting Eq.~(\ref{equ:new dh}) into Eq.~(\ref{equ:Kernel Fréchet mean}), the kernel mean is equivalent to minimizing the following problem:
\begin{equation}\label{equ:new kernel mean}
           \min_{\mu \in \mathcal{H}} \int \sum_{i}^{\infty}(\sqrt{\lambda_i}\phi_i(x)-\alpha_i)^2 \dif \mathbb{P}(x). 
\end{equation}
The solution of Eq.~(\ref{equ:new kernel mean}) is $\alpha_i=\int\sqrt{\lambda_i} \phi_i(x)\dif \mathbb{P}(x)$, that is, $\mu_{\mathcal{H}}=\sum_{i}^{\infty}\int\sqrt{\lambda_i} \phi_i(x)\dif \mathbb{P}(x)\sqrt{\lambda_i}\phi_i(\cdot)$. To simplify $\mu_{\mathcal{H}}$, we next analyze the interchange of limit and integral. Let $f_t(x)=\sum_i^{t}\sqrt{\lambda_i}\phi_i(x)$, and then $\{f_t(x)\}$ is a Cauchy sequence in $\mathcal{H}$. Since $\mathcal{H}$ denotes the complete metric space, there exists $\lim_{t \to \infty}  f_t(x) =f(x):=\sum_i^{\infty}\sqrt{\lambda_i}\phi_i(x)\in \mathcal{H}$, \textit{i.e.}, 
\begin{equation}
    \lim_{t \to \infty}\| f_t-f \|_{\mathcal{H}}=0.
\end{equation}
The above equation shows that $f_t(x)$ converges to $f(x)$ in norm $\|\cdot\|_{\mathcal{H}}$. For RKHS, the evaluation functional $\delta_x:\delta_x f \mapsto f(x)$ is a continuous linear functional, which means that for all $x\in\mathcal{X}$:
\begin{equation}
\begin{split}
    \lim_{t \to \infty} |f_t(x)-f(x)|  &\leq \lim_{t \to \infty} |\delta_x f_t - \delta_x f| \\
    &\leq \lim_{t \to \infty} \| \delta_x \| \| f_t -f \|_{\mathcal{H}} =0.\\
\end{split}
\end{equation}
The above inequality shows that convergence in norm implies pointwise convergence in RKHS. We now consider the difference between the integral of $f_t(x)$ and the integral of $f(x)$. According to the linearity and monotonicity of the Bochner integral, there exists 
\begin{equation}
    \left|\int f_t(x)\dif\mathbb{P}(x)\!-\!\int f(x)\dif\mathbb{P}(x)\right|\leq\int \left|f_t(x)\!-\!f(x)\right|\dif\mathbb{P}(x).
\end{equation}
By the reverse of Fatou's lemma related to $|f_t(x)-f(x)|$, we have the following inequality:
\begin{equation}
\begin{split}
   & \limsup_{t \to \infty} \int |f_t(x)-f(x)|\dif\mathbb{P}(x)\\ 
    &\leq \int \limsup_{t \to \infty} |f_t(x)-f(x)| \dif\mathbb{P}(x)
      =0,
    \end{split}
\end{equation}
which implies that the limit exists and 
\begin{equation}\label{equ:limit of ft}
   \begin{split}
  &  \lim_{t \to \infty} \left|\int f_t(x)\dif\mathbb{P}(x) -\int f(x)\dif\mathbb{P}(x) \right|\\
    &\leq \lim_{t \to \infty} \int |f_t(x)-f(x)|\dif\mathbb{P}(x) =0.\\
       \end{split}
\end{equation}
The above inequality shows that limit and integral is interchangeable with respect to kernel mean $\mu_{\mathcal{H}}$. Based on the uniqueness of limits for $\mathcal{H}$, the kernel mean is unique with the closed-form solution: 
\begin{equation}
\begin{split}
    \mu_{\mathcal{H}}&=\sum_{i}^{\infty}\int\sqrt{\lambda_i} \phi_i(x)\dif \mathbb{P}(x)\sqrt{\lambda_i}\phi_i(\cdot)\\
    &=\lim_{t \to \infty}\sum_{i}^{t}\int\sqrt{\lambda_i} \phi_i(x)\dif \mathbb{P}(x)\sqrt{\lambda_i}\phi_i(\cdot)\\
    &=\lim_{t \to \infty}\int\sum_{i}^{t}\sqrt{\lambda_i} \phi_i(x)\dif \mathbb{P}(x)\sqrt{\lambda_i}\phi_i(\cdot)\\
    &=\int\lim_{t \to \infty}\sum_{i}^{t}\sqrt{\lambda_i} \phi_i(x)\dif \mathbb{P}(x)\sqrt{\lambda_i}\phi_i(\cdot)\\
    &=\int\sum_{i}^{\infty}\sqrt{\lambda_i} \phi_i(x)\dif \mathbb{P}(x)\sqrt{\lambda_i}\phi_i(\cdot)\\
    &=\int K(x,\cdot) \dif \mathbb{P}(x).
\end{split}
\end{equation}
\end{proof}

\subsection*{Part C:  Stein methods and  SVGD algorithm}

\textbf{Stein’s Identity.} 
Given a smooth density  $p(x)$ observed on $\mathcal{X} \subseteq \mathbb{R}^{n}$, let $A_p$ be the Stein operator, $\varphi(x) = {[\varphi_1(x), ..., \varphi_n(x)]}^{\top}$ be the  smooth vector function. For sufficiently regular $\varphi$, Stein’s Identity is defined as:
\begin{equation}\label{eq:SI}
\begin{split}
\mathbbm{E}_{x\sim p}[A_p \varphi(x)] = 0,
\end{split}
\end{equation}
where
\begin{equation}
\begin{split}
A_p \varphi(x) = {\varphi(x)}\nabla_x {\rm log} p(x)^{\top} + \nabla_x \varphi(x).
\end{split}
\end{equation}
Here $x \sim p(x)$ denotes the continuous random variable or parameter sampled from $\mathcal{X}$, $\nabla_x \varphi(x) $ denotes the score function of $\varphi(x)$, $Q=A_p \varphi(x)$ denotes the Stein operator $A_p$ acting on function $\varphi$. \\
\textbf{Kernelized Stein Discrepancy.} 
Stein Discrepancy is a discrepancy measure which can maximize violation of Stein’s Identity and can be leveraged to define how different two smooth densities $p$ and $q$ are:
\begin{equation}\label{eq:SD}
{{\rm SD}(q, p)} = {\max \limits_{\varphi \in \mathcal{F}} \Bigg\{\Big[\mathbbm{E}_{x\sim q}\Big({\rm trace}(Q)\Big)\Big]^2 \Bigg\}}, 
\end{equation}
where $\mathcal{F}$ denotes a set of smooth functions with bounded Lipschitz norms, it determines the discriminative capability of Stein Discrepancy. However, with \cite{liu2019stein}, \cite{gorham2015measuring}, it is computationally intractable to resolve the functional optimization problem on $\mathcal{F}$. To bypass this and provide a close-form solution, Liu \emph{et al.} \cite{liu2016kernelized} introduce Kernelized Stein Discrepancy (KSD) which maximizes $\varphi$ in the unit ball of a reproducing kernel Hilbert space (RKHS) ${\mathcal{H}}$. The definition of KSD is thus presented:
\begin{equation}
\begin{split}\label{eq:KSD}
{{\rm KSD}(q, p)} = &{\max \limits_{\varphi \in {\mathcal{H}}}} \Bigg\{ \Big[\mathbbm{E}_{x\sim q}\Big({\rm trace}(Q)\Big)\Big]^2 \Bigg\} \\
 &{\rm s.t.} \quad {\|\varphi\|_{\mathcal{H}} \leqslant 1}.
\end{split}
\end{equation}
With \cite{chwialkowski2016kernel}, \cite{liu2016kernelized}, given a positive definite kernel $k(x, x^{’}): \mathcal{X} \times \mathcal{X} \to \mathbb{R}$, the optimal solution $\varphi(x)$ of Eq.~(\ref{eq:KSD}) is given as:
\begin{equation}\label{eq:optimal-solution}
\begin{split}
\varphi(x) = {\varphi_{q, p}^*(x) / {\|{\varphi^*_{q, p}}\|}_{\mathcal{H}}},
\end{split}
\end{equation}
where $\varphi_{q, p}^*(\cdot) = \mathbbm{E}_{x\sim q}[A_p k(x, \cdot)]$ indicates the optimal direction for gradient descent.
\par With notions of SD and KSD, Liu \emph{et al.} rethink the goal of variational inference which is defined in Eq.~(\ref{eq:vi-kl}), they consider the distribution set $\Omega$ can be obtained by smooth transforms from a tractable reference distribution $q_0(x)$ where $\Omega$ denotes the set of distributions of random variables which takes the form $r = T(x)$ with density:
\begin{equation}\label{eq:density}
\begin{split}
q_{[T]}(r) = q(\mathcal{R}) \cdot |\det(\nabla_r \mathcal{R})|,
\end{split}
\end{equation}
where $T: \mathcal{X} \to \mathcal{X}$ denotes a smooth transform,  $\mathcal{R}=T^{-1}(r)$ denotes the inverse map of $T(r)$ and $\nabla_r \mathcal{R}$ denotes the Jacobian matrix of $\mathcal{R}$. With the density, there should exist some restrictions for $T$ to ensure the variational optimization in Eq.~(\ref{eq:vi-kl}) feasible. For instance, $T$ must be a one-to-one transform, its corresponding Jacobian matrix should not be computationally intractable. Also, with \cite{rezende2015variational}, it is hard to screen out the optimal parameters for $T$.
\par Therefore, to bypass the above restrictions and minimize the KL divergence in Eq.~(\ref{eq:vi-kl}), an incremental transform ${T(x) = x + \varepsilon \varphi(x)}$ is introduced, where $\varphi(x)$ denotes the smooth function controlling the perturbation direction and $\varepsilon$ denotes the perturbation magnitude. With the transform $T(x)$ and the density defined in Eq.~(\ref{eq:density}), we have ${q_{[T^{-1}]}(x) = q(T(x))\cdot |\det(\nabla_x T(x))|}$ and ${{\rm KL}(q_{[T]} || p) = {\rm KL}(q || p_{[T^{-1}]})}$. We thus have
\begin{equation}
\begin{split}
\nabla_{\varepsilon} {\rm KL}(q || p_{[T^{-1}]}) = \nabla_{\varepsilon} \Big[\int q(x) \log \frac{q(x)}{p_{[T^{-1}]}(x)}\dif x{\Big]}.
\end{split}
\end{equation}
By converting the formula of ${{\rm KL}(q || p_{[T^{-1}]})}$ to the corresponding mathematical expection form, ${\nabla_{\varepsilon} {\rm KL}(q || p_{[T^{-1}]})}$ becomes ${-\mathbbm{E}_{x\sim q} {\Big[}{\nabla_{\varepsilon}\log\Big(}p(T(x)) |{\det(\nabla_x T(x))}|{\Big)}{\Big]}}$. Deriving this formula further, we have
\begin{equation}
\begin{split}
\nabla_{\varepsilon} {\rm KL}(q || p_{[T^{-1}]}) = -\mathbbm{E}_{x\sim q} \Big[\mathcal{C} + \mathcal{D}{\Big]},
\end{split}
\end{equation}
where 
\begin{equation}
\begin{split}
&\mathcal{C}=(\nabla_{T(x)}\log p(T(x)))^{\top}{\nabla_{\varepsilon} T(x)}, \\
&\mathcal{D}={(\nabla_{x} T(x))^{-1} \nabla_{\varepsilon} \nabla_{x} T(x)}.
\end{split}
\end{equation}
Besides, if ${\varepsilon = 0}$, there exists ${T(x)=x}$, ${\nabla_{\varepsilon}T(x)=\varphi(x)}$, ${\nabla_{x}T(x)=I}$, and ${\nabla_{\varepsilon} \nabla_{x} T(x)=\nabla_{x} \varphi(x)}$. With these knowledge, we present the following theorem.
\begin{theorem}\label{Stein-KL}
Let $T(x) = x + \varepsilon \varphi(x)$, $q_{[T]}(r)$ be the density $r = T(x)$, then
\begin{equation}
\begin{split}
\nabla_{\varepsilon}{\rm KL}(q_{[T]} \| p)|_{\varepsilon=0} =
-\mathbbm{E}_{x\sim q}[{\rm trace}(Q)],
\end{split}
\end{equation}
where $\nabla_{\varepsilon}{\rm KL}(q_{[T]} || p)$ denotes the directional derivative of ${\rm KL}(q_{[T]} \|p)$ when $\varepsilon$ tends to be infinitesimal (\textit{i.e.}, $\varepsilon \to 0$).
\end{theorem}
Relating the definition of KSD in Eq.~(\ref{eq:KSD}), Eq.~(\ref{eq:optimal-solution}), and Theorem \ref{Stein-KL}, the optimal perturbation direction for Eq.~(\ref{eq:optimal-solution}) can be identified as  $\varphi^*(\cdot)_{q, p}$. We thus present Lemma \ref{Stein-varphi}.
\begin{lemma}\label{Stein-varphi}
Given the conditions in Theorem \ref{Stein-KL} and consider all the perturbation directions $\varphi$ in the ball $\mathcal{B} = \{\varphi \in \mathcal{H}^n : \|\varphi\|^2_{\mathcal{H}} \leqslant {\rm KSD}(q, p) \}$ 
of $\mathcal{H}^n$ (\textit{i.e.}, RKHS), the direction of steepest descent that maximizes the negative gradient in Eq.~(\ref{eq:optimal-solution}) is $\varphi^*_{q, p}$:
\begin{equation}\label{eq:varphi}
\begin{split}
\varphi^*_{q, p} = \mathbbm{E}_{x\sim q}[k(x, \cdot)\nabla_x{\rm log} p(x) + \nabla_x k(x, \cdot)],
\end{split}
\end{equation}
for which we have $\nabla_{\varepsilon}{\rm KL}(q_{[T]} || p)|_{\varepsilon=0}$ = $-{{\rm KSD}(q, p)}$.
\end{lemma}

\subsection*{Part D: Parameter Optimization with Small Data}
\subsection*{D.1 Bilevel Fashion}
From the perspective of bilevel optimization \cite{rajeswaran2019meta}, \cite{wang2020global}, the meta-training phrase is formalized as:
\begin{equation}
\begin{split}
&\phi^{\ast} = \argmin \limits_{\phi} \sum_{i=1}^{M}{\mathcal{L}^{meta}{\Big(}{\theta^{\ast}}^{(i)}(\phi)}, \phi, {D_{source}^{val}}^{(i)}\Big) \\
&{\rm s.t.} \quad {{{\theta^{\ast}}^{(i)}(\phi)} = \argmin \limits_
{\phi} {\mathcal{L}^{task}\Big(\theta, \phi, {D_{source}^{train}}^{(i)}\Big)}},
\end{split}
\end{equation}
where $\mathcal{L}^{meta}$ and $\mathcal{L}^{task}$ denote the outer and the inner loss objectives, respectively. The inner-level optimization is with $\phi$ which is defined in outer-level optimization as condition, but $\phi$ cannot be changed during the inner-level optimization phrase; the outer-level optimization utilize ${{\theta^{\ast}}^{(i)}(\phi)}$ obtained from the inner-level optimization to optimize the meta-knowledge $\phi$. 
\subsection*{D.2 Task-Distribution Fashion}
Let $q(\mathcal{T})$ denote the distribution of tasks, $D$ denote the dataset for meta-tasks,  meta-task optimization (meta-learning) can be formalized as:
\begin{equation}
\begin{split}
\min \limits_{\phi} \underset{\mathcal{T} \sim q(\mathcal{T})}{\mathbb{E}} \mathcal{L}(D; \phi),
\end{split}
\end{equation}
where $\mathcal{L}(D; \phi)$ denotes the loss function to measure the performance of  the learning model.
To address the above optimization problem, it is assumed that we can obtain a set of source tasks sampled from $q(\mathcal{T})$.
With \cite{hospedales2020metalearning},  the above optimization consists of two phases: \textbf{meta-training} phrase and \textbf{meta-testing} phrase.
Let $\mathcal{D}_{source} = \{(D_{source}^{train}, D_{source}^{val})^{(i)}\}_{i=1}^{M}$ denote the set of $M$ source tasks in meta-training phase, where $D_{source}^{train}$ and $D_{source}^{val}$ denote the training and validation data respectively for source tasks, the serial number $i$ indicates each task; $\mathcal{D}_{target} = \{(D_{target}^{train}, D_{target}^{test})^{(i)}\}_{i=1}^N$ denote the set of $N$ target tasks in meta-testing phase, where $D_{target}^{train}$ and $D_{target}^{test}$ denote the training and testing data for target tasks,  respectively. On this setting, the meta-training phrase is to learn the optimal meta-knowledge $\phi^{\ast}$ and maximize the log likelihood by sampling different source tasks from $\mathcal{D}_{source}$, it is thus formalized as:
\begin{equation}
\begin{split}
\max \limits_{\phi} \log {p(\phi|\mathcal{D}_{source})}.
\end{split}
\end{equation}
By solving the maximum problem, we obtain a well-generalized meta-knowledge $\phi^{\ast}$, which is utilized to facilitate the model learning on unseen target tasks. The meta-testing phase aims to obtain a robust model on the training data of each unseen target task sampled from $\mathcal{D}_{target}$ with the help of $\phi^{\ast}$, which can be formalized as:
\begin{equation}
\begin{split}
\max \limits_{\theta} \log {p\Big(\theta|\phi^{\ast}, {{D_{target}^{train}}\Big)}}.
\end{split}
\end{equation}
We can thus obtain a model with parameters ${\theta}^{\ast}$ by solving the above maximum problem and evaluate its performance by conduct target tasks which are sampled from $D_{target}^{test}$.

\subsection*{D.3 Feed-Forward Fashion}
Let $\gamma = \mathbf{x}^{\top} \mathbf{e}_{\phi}\left(\mathcal{D}^{train}\right)$, a simple example for meta-training linear regression objective which optimizes over a distribution of meta-training tasks from feed-forward perspective is defined as:
\begin{equation}
\min _{\phi} \underset{\mathcal{T} \sim q(\mathcal{T})}{\mathbb{E}} \sum_{(\mathbf{x}, y) \in \mathcal{D}^{val}}\left[\left(\gamma -y\right)^{2}\right],
\end{equation}
where the training set $\mathcal{D}^{train}$ is embedded into the vector $\mathbf{e}_{\phi}$ which defines the linear regression weights, thus making prediction of samples $\mathbf{x}$ from the validation set $\mathcal{D}^{val}$.

It is noteworthy that meta-learning could perform well on optimizing small data training. That is, we could provide a potential parameter optimization over small data training with the power of meta-learning. In bilevel fashion, during the inner-level optimization phrase, the parameters could be optimized over small data training tasks; and the meta-knowledge could be efficiently updated during the outer-level optimization phrase. In task-distribution fashion, meta-learning could efficiently guide the model update in meta-training phrase under small data scenarios, and generalize well across unseen task distributions in meta-testing phrase. In feed-forward fashion, meta-learning could promote the parameter update over training task distributions in feed-forward manner under small data setting. Based on above fashions, the well-generalized meta-knowledge could be efficiently extracted. It's promising to introduce meta parameter optimization over small data training, and awaits futher exploration.
\end{document}


%
\title{A Survey of Learning on Small Data
   %
   }
%
%

\author{Xiaofeng~Cao,  
        Weixin~Bu, Shengjun~Huang,
        Yingpeng Tang, Yaming~Guo, \\
        Yi~Chang, \IEEEmembership{Senior Member IEEE},  
        and Ivor W. Tsang, \IEEEmembership{Fellow IEEE}  
\IEEEcompsocitemizethanks{\IEEEcompsocthanksitem \emph{X. Cao, W. Bu, Y. Guo and Y. Chang are with  the School of Artificial Intelligence, Jilin University, Changchun, 130012, China.  E-mail: xiaofeng.cao.uts@gmail.com, \{buwx21,guoym21\}@mails.jlu.edu.cn,  yichang@jlu.edu.cn}}\protect 

\IEEEcompsocitemizethanks{\IEEEcompsocthanksitem \emph{S. Huang and Y. Tang   are with the MIIT Key Laboratory of Pattern Analysis and Machine Intelligence, College of Computer Science and Technology, Nanjing University of Aeronautics and Astronautics, Nanjing 211106, China. 
E-mail: \{huangsj,tangyp\}@nuaa.edu.cn.     }}\protect

\IEEEcompsocitemizethanks{\IEEEcompsocthanksitem \emph{I. W. Tsang is with the  
Australian Artificial Intelligence Institute,  University of Technology Sydney, NSW 2008, Australia, and  the Centre for Frontier AI Research (CFAR), Agency for Science, Technology and Research (A$^*$STAR). 
E-mail: ivor.tsang@uts.edu.au.   }}\protect

\IEEEcompsocitemizethanks{\IEEEcompsocthanksitem \emph{Part of this work was finished when Dr. Cao was a research assistant at AAII of the University of Technology  Sydney.  }}\protect }

%
%

\markboth{Journal of \LaTeX\ Class Files,~Vol.~14, No.~8, August~2015}%
{Shell \MakeLowercase{\textit{et al.}}: Bare Demo of IEEEtran.cls for Computer Society Journals}
%


\IEEEtitleabstractindextext{%
\begin{abstract}\justifying  
Learning on big data brings success for artificial intelligence (AI), but the annotation and training costs are expensive. In future, learning on small data is one of the ultimate purposes of  AI, which   requires machines to recognize objectives and scenarios relying on small data as  humans.  A series of machine learning models is going on this way such as active learning, few-shot learning,  deep clustering. However, there are few theoretical guarantees for their generalization performance. Moreover, most of their settings are passive, that is, the label   distribution is explicitly controlled by one specified sampling scenario. 
This survey  follows the agnostic active sampling under a PAC (Probably Approximately Correct) framework    to analyze the generalization error and label complexity of  learning on small data using a supervised and unsupervised fashion. With these theoretical analyses, we categorize the small data learning models  from two geometric perspectives: the Euclidean  and non-Euclidean (hyperbolic) mean  representation, where  their   optimization solutions  are also presented and discussed. Later, some potential  learning scenarios that may benefit from small data  learning are then summarized, and   their potential learning scenarios are also analyzed.  Finally,   some challenging applications such as computer vision, natural language processing  that 
may benefit from learning on small data  are also surveyed.
\end{abstract}

\begin{IEEEkeywords}
Big data, artificial intelligence, small data, active learning,  PAC framework,  theoretical guarantee, hyperbolic geometry.
\end{IEEEkeywords}}

\maketitle

\IEEEdisplaynontitleabstractindextext

\IEEEpeerreviewmaketitle

\setcounter{section}{6}
\section{Challenging Learning Scenarios}

In this section, we introduce some challenging but practical learning settings. They become even troublesome in the small data regime, however, new opportunities also arise, which can broaden the applications of the methods for small data.

\subsection{Deep Learning Scenario}

Deep learning is one of the recent trending topics. It significantly pushes forward the state-of-the-art of various tasks \cite{kelleher2019deep}. However, deep models are usually data-hungry due to their massive parameters, which lead  to unfriendly tuning for applications. To tackle with this challenging problem, many works are proposed to reduce the amount of data for model training. Thus, we summary some related techniques and provide possible future directions for deep learning with small data.

To fully exploit the limited labeled examples, data augmentation \cite{shorten2019survey} is one of the most important techniques to enhance the performance without extra data requirements. It transforms the examples without changing the semantics to help the model learning. With the recent advances of Generative Adversarial Network (GAN) \cite{AldausariSMM23}, some studies also propose to generate more examples to augment the training set.
Another method is to exploit the unlabeled data. One can assign pseudo-labels to the unlabeled data, which can be achieved by, \eg, co-training \cite{QiaoSZWY18}, semi-supervised learning \cite{JiangMMH14}. Alternatively, unsupervised representation learning methods, \eg, contrastive learning \cite{ChenK0H20}, can also be employed to enhance the feature extraction. It constructs a pretext task on the unlabeled data for representation learning, then adapt to the downstream tasks. To show the effectiveness of this technique, Chen \textit{et al.} \cite{ChenKSNH20} report that they use only 10\% of the labels to fine-tune the pre-trained network, and achieve a comparable performance with the model trained with all labels from scratch for 90 epochs in ImageNet ILSVRC-2012 dataset.
Active learning \cite{RenXCHLGCW22} and transfer learning have also been widely applied in training deep models. The former selectively queries informative data for labeling, while the latter introduces external knowledge to help the model training (\eg, by domain adaptation \cite{WangD18}). Both of them are well validated to reduce the label requirement of training an effective deep model.

Possible future directions may include the interpretable deep learning methods \cite{chen2019looks}, so that we can incorporate prior knowledge to help the learning on small data. Besides, effective model selection methods with small data may also improve the performances by choosing suitable network architectures.

\subsection{Weakly-supervised Scenario}

Most of machine learning studies assume that the data can be accurately annotated with the ground-truth labels. However, in many real learning scenarios, the supervision can not meet this assumption due to the lack of perfect annotators, which leads to the noisy and inexact labels \cite{cohen2018acceleration}. The former case includes mislabeled training data, the latter case includes redundant labels, insufficient labels and coarse labels. Such learning scenarios become more tough in the small data regime. To help getting over this obstacle, we review some related methods and provide possible directions for the research on small data under this challenging setting.

\textbf{Noisy label.} To tackle with the noisy supervision in small data, possible solutions include noise level control and robust learning methods. The former can be implemented by, for example, active learning to query the easy but informative examples \cite{TangH19} or query from the high quality annotators \cite{HuangCMZ17} in order to reduce the noise level with limited budget. Another related technique for the low cost regime may be the active label cleaning \cite{bernhardt2022active}, which tries to actively rectify the most valuable mislabeled data. We note that identifying the informative noise data will be more challenging with only limited examples.
For the latter case, some related explorations, \textit{i.e.}, noisy few-shot learning, are conducted to learn robust representation with small data. For example, Gao \textit{et al.} \cite{GaoH0S19} propose a hybrid attention technique to solve the noisy few-shot relation classification task. They apply the attention modules to the prototypical networks to emphasize the crucial instances and features, so that the influence of the noise can be alleviated.  {Liang \textit{et al.} \cite{05494} propose a feature aggregation method adopted in a  transformer architecture to capture those noises by re-weighting  images.} The experimental results show that their proposed method can achieve better performance in the noisy setting compared to the state-of-the-art few-shot learning approaches. 

For the possible future directions, we believe the transfer learning and meta learning may be effective to improve the robustness of the model to the noise based on small data by introducing extra knowledge. Besides, exploiting the unlabeled data (\eg, semi-supervised learning, contrastive learning) may also provide useful information for learning on limited noisy labels.

\textbf{Inexact label.} For the inexact labels under small data regime, there are some pioneer works that try to utilize the unlabeled data to improve the representation learning. For example, Xie and Huang \cite{xie2020semi} tackle with the semi-supervised partial multi-label learning, where each labeled data are annotated with all relevant labels and some redundant labels. They exploit the unlabeled data to learn a low-dimensional embeddings and achieve better performances. Other possible techniques include the graph representation learning, which is well-suited to the inexact supervision setting due to its great potential in label completion \cite{ZhangZZ0XH20}.  Transfer learning is also a commonly used approach to complement the absent supervision from a related fully-supervised source domain, \eg, Cao \textit{et al.} \cite{CaoDZC0W21} propose a category transfer framework to help the object detection model training on the images with only class label annotation. It can handle both overlapping and non-overlapping category transfer, thus has wide applications. 

Look into the future, we also believe that the unsupervised learning techniques, \eg, contrastive learning, clustering, may help the model training on inexact supervised small data. Alternatively, active learning with inexact label querying may also be a promising direction. Because in some cases, the model may not need all information of a fully supervised example, thus querying the essential partial supervision can be a more cost-effective choice.

\subsection{Multi-label Scenario}

Multi-label learning \cite{zhang2013review} studies the setting that each instance is associated with multiple categories, \textit{i.e.}, map the data into $2^C$ dimensional label space with $C$ categories. Annotating multiple categories accurately can be difficult and expensive. Thus, the demand of learning on small data arises naturally to reduce the application threshold of multi-label learning. Next, we introduce some related explorations and possible future directions for this challenging setting.

Multi-label learning with limited supervision has been studied extensively. Some related approaches include semi-supervised multi-label learning \cite{chen2008semi}, partial multi-label learning \cite{XieH18}, active multi-label learning \cite{huang2015multi}, multi-label learning with missing label \cite{SunZZ10} and noisy label \cite{hu2019weakly}, \emph{etc}. Recently, the studies of few-shot multi-label learning \cite{AlfassyKASHFGB19} show the feasibility of training multi-label classifier on small data with the help of excess knowledge. These techniques sufficiently exploit the label correlation and data distribution. For example, the low rank regularization \cite{jing2015semi} is a commonly used constraint to incorporate the prior knowledge of class correlation to reduce the training data. Graph-based methods exploit the data topological information to tackle with the, \eg, missing label \cite{yu2014large}, semi-supervised data \cite{WangLHH0HC20}. Active learning methods have also been exploited to deal with the low labeling budget, some of them perform instance-level query and evaluate the uncertainty of the data based on different criteria \cite{LiG13,ChakrabortyBP11}, while the others query the instance-label pair for fine-grained supervision \cite{WuGSZCL17}.

Future directions may include the meta multi-label leaning. Because the label correlation prior can be crucial to the small data setting, which may be induced from historical learning experiences by the recent trending technique, \textit{i.e.}, meta-learning. More effective active querying methods for multi-label learning are also expected to be developed.

\section{Challenging Applications}

In many real applications, the data gathering  or annotation is extremely expensive, which leads to the limited training data and brings challenges to the model learning. This phenomenon brings opportunities for small data techniques. This section summarizes some of the major application areas for small data, which include computer vision, natural language processing and other topics.

 \subsection{Computer Vision}
Computer vision \cite{shapiro2001computer} deals with the image recognition, object detection, semantic segmentation, video processing, \emph{etc}. It has a wide range of applications in our daily life. Thanks to the powerful feature extraction ability of deep models, the state-of-the-art in computer vision has been pushed forward significantly. However, the requirement of labeled data has also been promoted to a new level due to the growing model complexity, which may be unreachable to some applications. 
For example, in most of medical image analysis tasks, the labeled examples are highly scarce because of the involving of expensive facilities and considerable expertise requirement for annotation. Especially for the rare diseases, both of the data and reliable annotators are limited. It is thus necessary to develop small data techniques to tackle with this scenario. Existing explorations include designing active learning algorithms for effective queries \cite{hoi2006batch,wang2020deep}, transfer knowledge from related domains \cite{ZhangMML18}, exploiting the unlabeled data, \emph{etc}. Nevertheless, learning on small data for medical images is still a challenging problem.
Remote sensing image analysis \cite{richards1999remote} is another typical task which is lack of large amount of labeled data. It is not only because of the privacy of the data, but also the wealthy number of objects in each single image. Some studies try to tackle with this problem by proposing weakly-supervised learning methods to relax the requirement of accurate annotations \cite{MaggioriTCA17}. Besides, novel data augmentation methods are also exploited to help the model training with limited examples \cite{LongGXL17}. Active learning is also a popular technique in the remote sensing image. The uncertainty and diversity are two commonly used criteria for remote sensing data selection \cite{demir2010batch}.

Other important applications in computer vision may also include video processing \cite{TianCGYSY20}, 3D image analysis \cite{konyushkova2015introducing}, \emph{etc}.

 \subsection{Natural Language Processing}

Natural language processing (NLP) system tackles with the semantic parsing, translation, speech recognition, summarization, \emph{etc}. for the human language. It is one of the core techniques to the man-machine interaction. However, a large numbers of labeled examples are required to train the leading NLP models, \eg, Transformer \cite{DevlinCLT19}, which can be prohibitive to some applications.
For example, in the text summarization task, whose goal is to extract the key words, outline, or abstract for a long document, the annotators need to read the whole article to provide the accurate summarization. This process is usually laborious and tedious. As a result, a large scale of labeled data is usually not available in many cases. Further, due to the high dimensional feature and label spaces, learning with small data can be very challenging. Some explorations are conducted to tackle with this problem. Karn \textit{et al.} \cite{abs-2103-05131} propose a few-shot learning method for text summarization, which exploits the synthetic data to pretrain the model. Data augmentation \cite{fabbri2020improving} has been studied to improve the model performance for few-shot text summarization learning. Bayesian active learning \cite{gidiotis2021bayesian} has also been exploited to reduce the annotation cost. More effective techniques are expected to be explored in the future to tackle with this task.
Another important application is the question-answering and dialogue system. It includes speech-recognition, semantic parsing, dialogue generation, and other functions, which lead to the heavy burden of data labeling. Also, due to the difficulty of the learning task, small data can usually not result in a high performance. Several active learning methods \cite{sugiura2011situated} are proposed to alleviate this problem. While transfer learning \cite{ShalyminovLEL19} and reinforcement learning \cite{EshghiSL17} have also been explored for reducing the training data in dialogue system. However, there still exists vast space for development on the application of question-answering and dialogue system with small data.

Beyond the above topics, speech transcription \cite{VashisthaSA17}, voice wake-up \cite{ChenBMWSX21}, machine translation \cite{FosterVUMKFJSWB18}, \emph{etc}., can also be important applications for small data learning.


 \subsection{Recommender System}

Recommender system \cite{zhang2019deep} is one of the most important applications in our daily life. It recommends possible interested contents to the users based on their historical behaviors (\eg, user-item rating, transactions) and has been widely applied to the e-commerce, music, video sharing, social media platforms, \emph{etc}. One of the major problems of the recommender system is the cold start problem, \textit{i.e.}, there is few or no historical data for a new user, and the data is usually private and expensive to acquire. It is thus induce the demand of developing small data techniques for the recommender system. However, the targets in this application are usually plentiful and sparse, and the feature dimension is also high, which makes the small data learning beset with difficulties.
One solution to deal with this problem is introducing the extra knowledge, \eg, social media \cite{CamachoS18}, related source domain \cite{WangAMHNYK19}, Linked Open Data repositories \cite{NatarajanVNG20}. Besides, some studies assume that a small questionnaire can be queried from the user to alleviate the cold start problem \cite{SunLLZLZ13}. A related approach may be the active learning, which tries to selection the most essential user-item entries for querying \cite{CaiTCHWH19}. Nevertheless, this problem is still open for further explorations.

 \subsection{Time Series Data}

Time series data \cite{brillinger2001time} is very common in our life (\eg, the temperature records, stock market trends, monitoring), It consists a sequence of successive data points. However, many time series data confront with lack of sufficient labels, incomplete sequence problems \cite{farha2019ms}, and the annotation for each data   is laborious.
Existing studies employ active learning \cite{shin2021coherence} to selectively label the most informative data   to improve the model with limited budget; or exploit the unlabeled data by assigning pseudo-labels for model training, which can be implemented by label propagation \cite{ma2020sf} and model responses \cite{moltisanti2019action}. Further studies of learning on small data for time series data are expected to conducted.

 \subsection{Biology}

In recent years, machine learning has been widely applied to help the modern biological research \cite{CAMACHO20181581}. Typical applications include automated genome annotation, protein binding prediction, metabolic functions prediction, \emph{etc}. However, obtaining biology data usually requires a long-time culture (\eg, cell culture), or the involving of expensive apparatuses, which bring challenges to collect large scale labeled data. Moreover, the biomedical data is often high-dimensional and sparse \cite{2016GWAS}, much of them is even incomplete and biased \cite{vzitnik2015data}. Because of these challenges, training effective models for biological tasks can be rather difficult, especially when the labeled data is limited.
Some explorations employ active learning to reduce the label acquisitions \cite{begum2021application}; transfer learning has also been exploited to reconstruct the incomplete data \cite{mignone2020exploiting}; graph neural network and matrix factorization techniques are applied to the disease-gene association identification task \cite{han2019gcn}. More effective approaches to deal with the small data need to be further studied.

\section{Conclusion}
In this paper, we firstly present a formal definition for learning on small data, and then provide  theoretical guarantees for its supervised and unsupervised    generalization analysis on error and label complexity    under a PAC framework.
From a geometric perspective, learning on small data can be characterized by  the Euclidean and non-Euclidean geometric representation, where their geometric mean representations are  presented and analyzed with respect to a unified expression of Fréchet mean. To optimize those geometric means,  the Euclidean gradient, Riemannian gradient, and Stein gradient are investigated. Besides these technical contents, some potential future directions of learning on small data are also summarized, and  their related advanced  challenging   scenarios and applications  are also presented and discussed.



%
 \bibliographystyle{IEEEtran}
\bibliography{Reference,typ}

%
\vspace{-1.6cm}
\begin{IEEEbiography}
[{\includegraphics[width=1in,height=1.25in,clip,keepaspectratio]{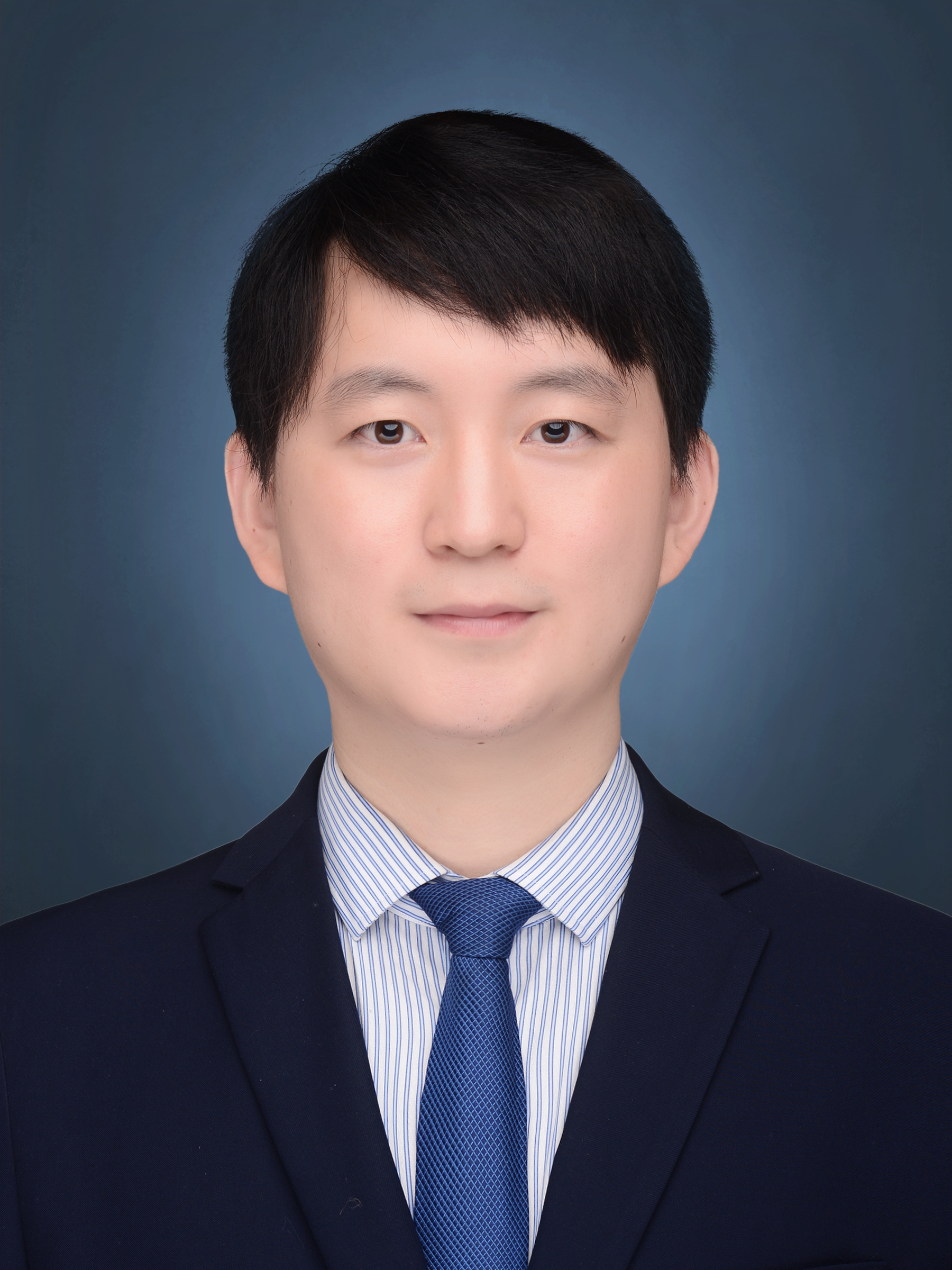}}]{Xiaofeng Cao} received his Ph.D. degree at Australian
Artificial Intelligence Institute, University of Technology
Sydney, Australia. He is currently an Associate
Professor at the School of Artificial Intelligence, Jilin
University, China and leading a Machine Perceptron
Research Group with more than 15 PhD and Master
students. He has published more than 10 technical
papers in top tier journals and conferences, such as
IEEE T-PAMI, IEEE TNNLS, IEEE T-CYB, CVPR,
IJCAI. 
His research interests include the PAC learning theory, agnostic learning algorithm,
generalization analysis, and hyperbolic geometry.

\end{IEEEbiography}

\vspace{-1.6cm}
\begin{IEEEbiography}
[{\includegraphics[width=1in,height=1.25in,clip,keepaspectratio]{WeixinBu}}]{Weixin Bu}  is now a master student supervised by Dr. Cao at the School of Artificial Intelligence, Jilin University. Before that, he spent almost three years from a software engineer to a leader of data team in BorderX Lab INC, where he built and optimized various data systems. He received the BSc degree in network engineering from Nanjing University of Information Science \& Technology, China, in 2018. His main research interests include small data learning and graph representation learning.
\end{IEEEbiography}

\vspace{-1.6cm}

\begin{IEEEbiography}
[{\includegraphics[width=1in,height=1.25in,clip,keepaspectratio]{huangsj}}]{Sheng-Jun Huang }   received the BSc and PhD
degrees in computer science from Nanjing University, China, in 2008 and 2014, respectively.
He is now a professor in the College of Computer
Science and Technology at Nanjing University
of Aeronautics and Astronautics. His main research interests include machine learning and
data mining. He has been selected to the Young
Elite Scientists Sponsorship Program by CAST
in 2016, and won the China Computer Federation Outstanding Doctoral Dissertation Award in
2015, the KDD Best Poster Award at the in 2012, and the Microsoft
Fellowship Award in 2011. He is a Junior Associate Editor of Frontiers
of Computer Science.
\end{IEEEbiography}

\vspace{-1.6cm}

\begin{IEEEbiography}[{\includegraphics[width=1in,height=1.25in,clip,keepaspectratio]{typ.jpg}}]{Ying-Peng Tang}
	received the BSc degree from the Nanjing University of Aeronautics and Astronautics, China, in 2020. He is currently pursuing the Ph.D. degree in computer science and technology with the Nanjing University of Aeronautics and Astronautics, Nanjing, China. His current research interests include active learning and semi-supervised learning. He has been awarded for China National Scholarship in 2019,  and the Excellent Master thesis in Jiangsu Province in 2021.
\end{IEEEbiography}

\vspace{-1.6cm}
\begin{IEEEbiography}
[{\includegraphics[width=1in,height=1.25in,clip,keepaspectratio]{YamingGuo}}]{Yaming Guo}  is currently pursuing the  Ph.D degree supervised by Dr. Cao at the School of Artificial Intelligence, Jilin University. He received the BSc degree in applied mathematics from Inner Mongolia University, China, in 2021. He has won the China National Scholarship in 2021. His main research interests include active learning and federated learning.
\end{IEEEbiography}

\begin{IEEEbiography}
[{\includegraphics[width=1in,height=1.25in,clip,keepaspectratio]{YiChang}}]{Yi Chang}  received the
Ph.D. degree in computer science from the
University of Southern California, Los Angeles, CA,
USA, in 2016.
He is currently the Dean of the School of
Artificial Intelligence, Jilin University, Changchun,
China. He has published more than 100 research
papers in premium conferences or journals. He has
broad research interests on information retrieval,
data mining, machine learning, and natural language
processing.

Prof. Chang is an Associate Editor of the IEEE Transactions on Knowledge and Data Engineering.
\end{IEEEbiography}

\begin{IEEEbiography}
[{\includegraphics[width=1in,height=1.25in,clip,keepaspectratio]{Ivor}}]{Ivor W. Tsang}  is Professor of Artificial Intelligence, at University of Technology Sydney. He is also the Research Director of the Australian Artificial Intelligence Institute, and an   IEEE Fellow.  In 2019, his paper titled ``Towards ultrahigh dimensional feature selection for big data" received the International Consortium of Chinese Mathematicians Best Paper Award. In 2020, Prof Tsang was recognized as the AI 2000 AAAI/IJCAI Most Influential Scholar in Australia for his outstanding contributions to the field of Artificial Intelligence between 2009 and 2019. His works on transfer learning granted him the Best Student Paper Award at International Conference on Computer Vision and Pattern Recognition 2010 and the 2014 IEEE Transactions on Multimedia Prize Paper Award. In addition, he had received the prestigious IEEE Transactions on Neural Networks Outstanding 2004 Paper Award in 2007.  
\par Prof. Tsang serves as a Senior Area Chair for Neural Information Processing Systems and Area Chair for International Conference on Machine Learning, and the Editorial Board for  Journal Machine Learning Research, Machine Learning, 
Journal of Artificial Intelligence Research, and IEEE Transactions on Pattern Analysis and Machine Intelligence. 
\end{IEEEbiography}